%% file: iclr2027_musebench.tex
\documentclass{article}
\usepackage{iclr2027_conference,times}
\iclrfinalcopy
\usepackage[utf8]{inputenc}
\usepackage[T1]{fontenc}
\usepackage{hyperref}
\usepackage{titletoc}
\hypersetup{
  hidelinks,
  pdftitle={MuseBench: Benchmarking Intent-Level Audiovisual Arts Understanding in MLLMs},
  pdfauthor={Yuxuan Fan, Gyusik Seo, Jing Hao, Jaemin Cho, Mohit Bansal, Jaehong Yoon}
}
\usepackage{url}
\usepackage{booktabs}
\usepackage{amsfonts}
\usepackage{nicefrac}
\usepackage{microtype}
\usepackage[table]{xcolor}
\usepackage{graphicx}
\usepackage{subcaption}
\usepackage{wrapfig}
\usepackage{multirow}
\usepackage{enumitem}

\usepackage[ruled,vlined]{algorithm2e}
\SetKwFor{ForEach}{for each}{do}{endfor}
\usepackage{amsmath}
\usepackage{tabularx}
\usepackage{makecell}
\usepackage{ragged2e}
\usepackage{pifont}
\usepackage{xspace}
\usepackage{cleveref}
\usepackage{tcolorbox}
\tcbuselibrary{skins}
\usepackage{longtable}
\newtcolorbox{findingbox}{enhanced,colback=gray!12,colframe=gray!55,boxrule=0.6pt,arc=4pt,left=8pt,right=8pt,top=5pt,bottom=5pt}
\definecolor{myblue}{RGB}{189,215,238}
\definecolor{mygray}{RGB}{240,240,240}
\definecolor{mymultiem}{RGB}{235,244,251}
\definecolor{cherryred}{RGB}{220,20,60}
\newcommand{\benchname}{\textsc{MuseBench}\xspace}
\newcommand{\cmark}{\ding{51}}
\newcommand{\xmark}{\ding{55}}

\Crefname{section}{Section}{Sections}
\Crefname{table}{Table}{Tables}
\crefname{section}{Sec.}{Secs.}
\crefname{table}{Tab.}{Tabs.}
\crefname{figure}{Fig.}{Figs.}
\crefname{appendix}{Sec.}{Secs.}

\title{MuseBench: Benchmarking Intent-Level Audiovisual Arts Understanding in MLLMs}

\author{%
    \bfseries Yuxuan Fan\textsuperscript{1}\quad
    Gyusik Seo\textsuperscript{1}\quad
    Jing Hao\textsuperscript{2}\quad
    Jaemin Cho\textsuperscript{3\,4}\quad
    Mohit Bansal\textsuperscript{5}\quad
    Jaehong Yoon\textsuperscript{1\dag}\\[0.1in]
    \normalfont\textsuperscript{1}NTU Singapore\quad
    \normalfont\textsuperscript{2}The University of Hong Kong\\
    \normalfont\textsuperscript{3}Johns Hopkins University\quad
    \normalfont\textsuperscript{4}AI2\quad
    \normalfont\textsuperscript{5}UNC-Chapel Hill
}
\begin{document}

\maketitle
\lhead{Preprint}
\renewcommand{\thefootnote}{\fnsymbol{footnote}}
\footnotetext[2]{Corresponding author}
\renewcommand{\thefootnote}{\arabic{footnote}}
\input{sections/00.abs_iclr.tex}
\input{sections/01_intro}

\input{sections/02_related}
\input{sections/03_bench}
\input{sections/04_exp}
\input{sections/05_con}

\subsection*{AI use statement}
AI models were used both in benchmark construction and to assist manuscript preparation.
As described in \cref{sec:collection_annotation}, GPT-5.4-mini assisted with search-keyword generation and source-relevance assessment, Whisper-Large-v3 transcribed the source videos, and Keye-VL-1.5 generated clip captions.
Language models generated candidate questions, answers, and distractors from the construction materials (\cref{sec:appendix:phase_c,sec:appendix:phase_d}).
The construction pipeline incorporated iterative human review and prompt revision, and every retained item was manually verified before model assessment (\cref{sec:hitl_loop}).
LLMs also assisted with writing and revising code.

For manuscript preparation, LLMs assisted with drafting, revising, and polishing the text, including improving clarity and organization.
The authors retain responsibility for the research design, experimental results, interpretation, citations, and final content of the manuscript.

\bibliography{musebench}

\newpage
\appendix
\startcontents[appendix]
\section*{Appendix Contents}
\begingroup
\small
\printcontents[appendix]{}{1}{\setcounter{tocdepth}{2}}
\endgroup
\newpage
\input{sections/99_appendix}
\stopcontents[appendix]

\end{document}

%% file: sections/00.abs_iclr.tex
\begin{abstract}
Audiovisual arts encompass diverse creative disciplines, including cinema, visual arts, stage performance, and game design, where artistic meaning arises from deliberate combinations of visual, auditory, and narrative elements (e.g., fear amplified through claustrophobic framing, or grief conveyed through silence and lingering close-ups). 
Artistic understanding extends beyond recognizing what is depicted to reasoning about why it is expressed through particular creative choices.
Despite progress in multimodal large language models (MLLMs), reasoning about artistic intent remains underexplored, as existing benchmarks largely measure perceptual recognition.
To address this gap, we introduce \benchname, a benchmark for evaluating intent-level artistic understanding in MLLMs.
It comprises 4,016 questions derived from video essays spanning cinematic arts, static visual arts, stage performing arts, and game arts.
To capture the open-ended nature of artistic analysis at scale, the benchmark combines single-select and variable-option multi-select questions. 
All questions are generated and refined through a four-phase iterative pipeline combining shortcut filtering, adversarial distractors, and expert validation.
We evaluate 34 MLLM versions using either video and audio transcripts (V+T) or video, native audio, and text (V+A+T). The highest observed accuracy is 64.62\% (V+A+T), trailing the 87.18\% human expert baseline. Further analysis reveals that native-audio models score lowest on game arts, models often recover partial interpretations but struggle to identify complete correct option sets on multi-select items, and adaptive frame-selection models trail the strongest video-specific model. These results identify artistic knowledge and complete interpretation as challenges for audiovisual models.
\end{abstract}
\begin{center}
Project Page: \url{https://musebench.github.io}
\end{center}

%% file: sections/01_intro.tex
\section{Introduction}
\label{sec:intro}
\begin{figure}[t]
\centering
\vspace{-0.1in}
\includegraphics[width=\textwidth]{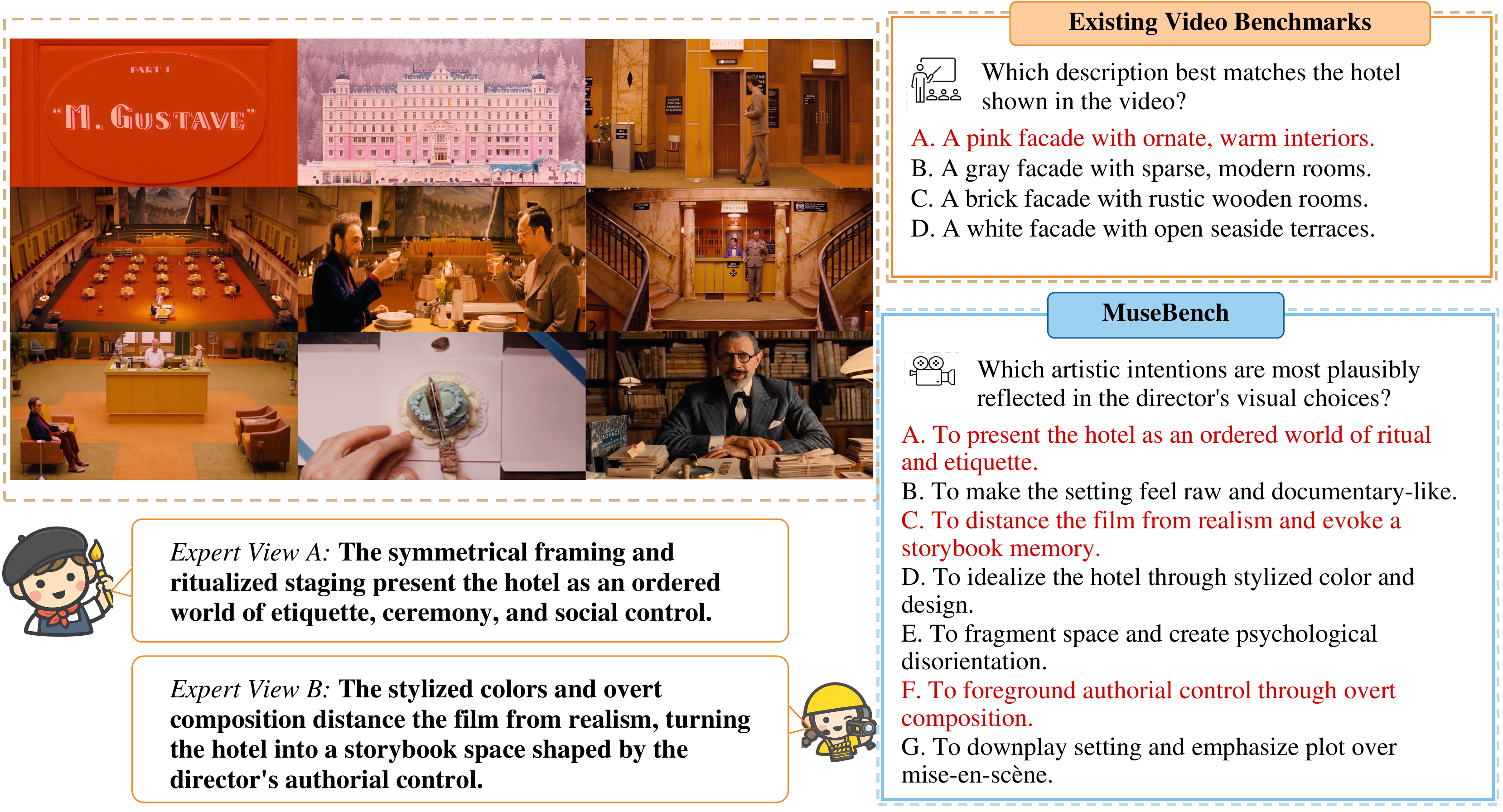}
\vspace{-0.2in}
\caption{\textbf{Overview of \benchname.} Cinematic frames on the left ground two contrasting questions on the right. Existing video benchmarks (orange) test surface-level recall with a single correct answer, while \benchname (blue) probes the artistic intent behind directorial choices and allows multiple defensible answers, highlighted in red. The bottom-left dialogue shows how viewers can reach different yet well-supported interpretations of the same scene. This plurality in artistic understanding motivates our multi-select design, in which multiple options may be simultaneously correct.
}
\label{fig:intro}
\vspace{-0.2in}
\end{figure}

What does it mean to understand art? 
It is not merely recognizing what is shown, but interpreting why it is expressed in a particular way.
The audiovisual arts~\citep{krupskyy2021determinants, lokki1998realtime, kot2021problems}, spanning cinema, visual arts, stage performance, and interactive media, provide a uniquely demanding setting for exposing this distinction.
An artistic work is a deliberately designed expressive system in which creators orchestrate camera movement, composition, editing pace, lighting, blocking, and visual style to convey emotion, theme, and aesthetic intent~\citep{carvalho2015audiovisual, slawek2023perspective}. Understanding such artistic expression requires reasoning about why a technique was chosen, how a visual arrangement serves creative intention, and what deeper artistic meaning emerges from the interplay of form and content~\citep{barber2012understanding, naphade2002extracting}.
For example, as illustrated in \cref{fig:intro}, asking why a director pairs symmetric framing with warm lighting, or punctuates a scene with prolonged silence, requires linking visual form to emotional intent rather than naming on-screen objects.
This demands a level of comprehension that goes well beyond factual recognition or surface-level description: models must grasp not only what appears on screen but also the creator's underlying intent and the cultural conventions that inform it.

Although multimodal large language models (MLLMs)~\citep{Song2023MovieChatFD, zohar2025apollo, Yang2024VCAVC, he2026video, Wang2025VideoRTSRR, tao2025moss, song2025videonsa} have rapidly improved on standard perception and reasoning tasks, their capacity for deeper artistic understanding remains underexplored. As illustrated in \cref{fig:intro}, existing video understanding benchmarks~\citep{xu2017video, yu2019activitynet, Tan2025ALLVBAL,Song2025VideoMMLUAM} primarily evaluate what is happening in a scene with a single correct option, rather than whether models can infer the intent behind creative decisions, such as why a director relies on symmetric composition, warm palettes, and ritualized blocking, or interpret their artistic significance.
However, constructing a rigorous benchmark for intent-level artistic understanding is challenging on three intertwined fronts. \textbf{(i) Expert-knowledge scarcity:} Professional artistic analysis is inherently sparse, and authoring intent-level questions at benchmark scale is prohibitively expensive, well beyond what crowdsourcing can reliably supply. \textbf{(ii) Multiple valid interpretations:} Many analytical questions are constrained but non-unique and admit several defensible perspectives, so the dominant fixed four-option single-choice format collapses this plurality onto a single answer and reduces to pattern matching. \textbf{(iii) Reliable assessment of interpretation:} Even with high-quality questions, evaluation itself is a measurement problem. Naive accuracy on fixed-option items is not comparable across questions with different option counts, conflates successful guessing with genuine interpretation, and may fail to capture partial credit for the set-valued analytical judgments that artistic reasoning naturally produces. Addressing these challenges requires rethinking data sourcing, question format, and evaluation protocol in concert.

We tackle these challenges through three coordinated design choices, each directly aligned with the corresponding issues outlined above.
\textbf{(i) Constructing expert-supervised data from video essays.} We use video essays~\citep{bresland2010origin, lavik2012video} sourced primarily from YouTube, Bilibili, and TikTok. These analytical videos pair professional commentary with on-screen demonstrations, providing temporally aligned expertise and visual evidence. We develop a four-phase construction pipeline with iterative in-context updating (\cref{sec:collection_annotation}) to transform over 10{,}000 video essays into benchmark questions that require visual understanding rather than transcript-based shortcuts. Within this pipeline, every distractor is crafted by an adversarial step that combines seven strategies spanning technical misread, over-simplification, concept swap, cause-effect inversion, scope error, temporal confusion, and partial truth, so that all options appear equally plausible to a reader without access to the clip and shortcut-driven guessing is suppressed. \textbf{(ii) Representing plurality through mixed formats.} We move beyond the fixed four-option paradigm (\cref{sec:scope_design}) and interleave single-select questions, which probe whether a model can identify the most precise interpretation, with multi-select questions that probe whether it can enumerate the full set of valid analytical dimensions, while the per-question option count varies between four and eight so that the answer space reflects the open-ended structure of artistic reasoning rather than a uniform template. \textbf{(iii) Principled evaluation protocol.} For reliable assessment, we introduce a new scoring protocol designed for this heterogeneous setting (\cref{sec:metrics}). We report overall accuracy (ACC) as the fraction of questions answered fully correctly. Our primary format-specific metrics are Chance-Adjusted Accuracy (CAA) for single-select questions, with expected score $0$ under uniform guessing, and exact match (EM) for multi-select questions, requiring the complete correct option set. Precision, recall, and F1 diagnose incorrect selections, missed answers, and partial agreement. Empirically (\cref{sec:exp}), models often recover part of the correct set but struggle with exact recovery: both omitted answers and excess selections limit multi-select performance (\cref{tab:app_evaluation_results}).

Across 34 evaluated MLLM versions, the highest observed overall accuracy is $64.62\%$ (using video, native audio, and text), leaving a 22.56-point gap to the $87.18\%$ expert baseline. We report results separately for transcript-based (V+T) and native-audio (V+A+T) input settings. Our analysis shows that native-audio models score lowest on game arts, whereas some V+T models achieve their highest category CAA there. Furthermore, models often recover partial interpretations but struggle to identify complete correct option sets on multi-select questions, and benefit little from adaptive key-frame selection. These findings motivate richer artistic supervision and evaluation of both partial and complete interpretations.

In summary, this paper makes three key contributions:
\begin{itemize}
    \item \textbf{Benchmark for Audiovisual Arts Understanding.} We introduce \benchname, a new benchmark for audiovisual arts expertise, covering four art categories and 11 sub-domains, with questions over a variable number of options to capture interpretive plurality.
    \item \textbf{Scalable Expert-Knowledge Pipeline.} We develop a four-phase construction pipeline under iterative in-context updating that leverages video essays and vision-language models to generate visually grounded, intent-level questions at scale, addressing the fundamental challenge of acquiring expert knowledge for creative domains.
    \item \textbf{Principled Evaluation Protocol and Analysis.} We report overall ACC, single-select CAA, and multi-select EM, with P/R/F1 diagnosing partial agreement. Our evaluation covers 34 MLLM versions. Separate V+T and V+A+T comparisons characterize performance under different inputs. Multi-select analysis reveals that partial agreement can coexist with poor recovery of the complete correct option set.
\end{itemize}

%% file: sections/02_related.tex
\section{Related Work}
\label{sec:related_work}
\textbf{Multimodal Large Language Models for Video Understanding.}
Multimodal Large Language Models (MLLMs) have advanced rapidly in video understanding, with efficient processing of many frames as a central challenge. One line of work pursues efficient encoding via sparse token memory~\citep{Song2023MovieChatFD}, visual summarization tokens~\citep{Shu2024VideoXLEV}, or native sparse attention for long contexts~\citep{song2025videonsa}. A parallel line casts video understanding as agentic retrieval, using tree search~\citep{Yang2024VCAVC}, interleaved reasoning with temporal grounding~\citep{yang2025longvt}, or multi-agent coordination~\citep{chen2025lvagent}. Complementary training-side advances include process rewards for temporal alignment~\citep{tao2025moss} and empirical studies of sampling and scaling~\citep{zohar2025apollo}.
Most video-understanding evaluations focus on everyday activities, open-domain QA, or academic lectures. ShotBench~\citep{liu2025shotbench} extends evaluation to expert cinematic understanding; \benchname examines how creative choices serve expressive or functional purposes across four artistic domains.

\noindent \textbf{Benchmarks for Video Understanding.}
Video understanding benchmarks have progressed from short-clip QA~\citep{xu2017video,yu2019activitynet} to story-level and temporal-reasoning frameworks~\citep{huang2020movienet,li2024mvbench}. Recent work expands along multi-modal breadth~\citep{fu2025video}, long-video scale~\citep{wang2025lvbench,Tan2025ALLVBAL}, and domain knowledge on expert lectures and STEM reasoning~\citep{Hu2025VideoMMMUEK,Song2025VideoMMLUAM}. Audio-visual perception has been explored in parallel, with AV-Odyssey Bench~\citep{gong2024av} probing fine-grained contrasts such as pitch and loudness, and our work extends this inquiry from low-level perception toward the interpretation of artistic intent. Despite this growing breadth, existing benchmarks largely center on general activities, factual comprehension, or academic STEM knowledge, 
leaving expertise in cinematographic technique, compositional principles, and performance craft underexplored.
ShotBench evaluates cinematic language across eight dimensions using film images and clips, with four-option single-choice questions. \benchname connects techniques to their effects and functions across cinematic, static visual, stage performing, and game arts, using both single-select and multi-select questions. For example, recognizing a camera angle and explaining how that angle conveys authority test related but distinct capabilities. Appendix~\ref{sec:appendix:comparison} compares their scope, construction, and answer formats.

%% file: sections/03_bench.tex
\section{MuseBench Construction}
\label{sec:design_construction}

This section details the design and construction of \benchname. We first motivate video essays as an expert-narrated knowledge source for probing audiovisual analytical understanding (\cref{sec:video_essay}), and then introduce our hierarchical capability taxonomy and two complementary question formats
(\cref{sec:scope_design}). Building on this design, we describe a four-phase construction pipeline that transforms raw video essays into candidate question-answer pairs (\cref{sec:collection_annotation}), followed by an iterative quality review
(\cref{sec:hitl_loop}). Finally, we introduce the specific evaluation metrics in \benchname (\cref{sec:metrics}).

\begin{figure}[t]
\centering
\vspace{-0.1in}
\includegraphics[width=1.0\textwidth]{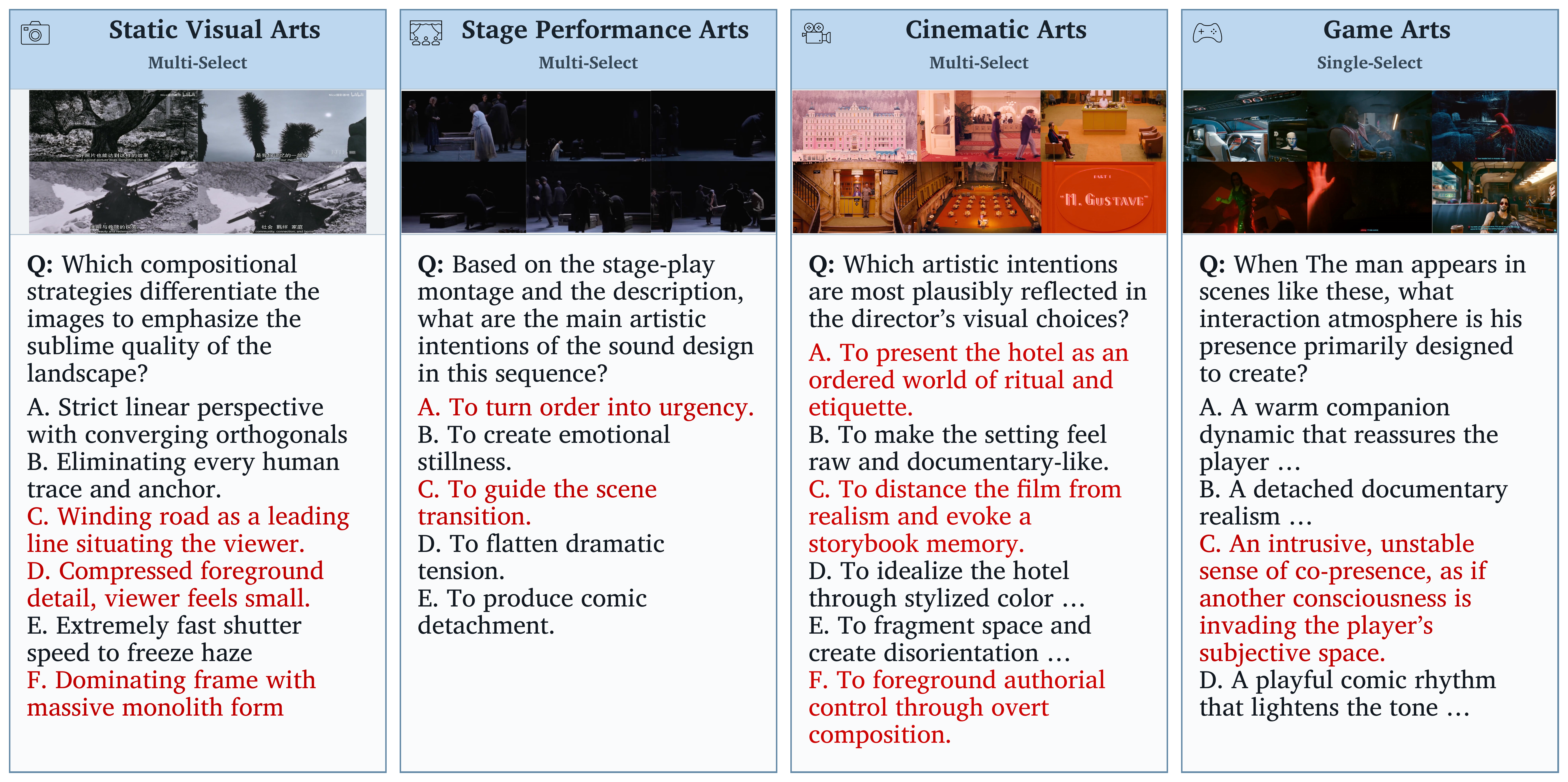}
\vspace{-0.17in}
\caption{Representative examples from four \benchname categories.}
\label{fig:category_examples}
\vspace{-0.1in}
\end{figure}

\subsection{Video Essays as a Knowledge Source}
\label{sec:video_essay}

A \textit{video essay} is an analytical audiovisual format in which critics, educators, or practitioners examine artistic works through temporally aligned expert commentary and supporting visual or auditory evidence.
Video essays are particularly well suited to our setting due to three key properties: (i)~\emph{expert-narration density}, creators explain not only what a technique is but why it produces a particular effect; (ii)~\emph{narration-to-evidence alignment}, spoken analysis directly references on-screen evidence; 
and (iii)~\emph{creative-arts coverage} across domains under-represented in existing video benchmarks~\citep{Song2025VideoMMLUAM}, such as cinematography, fine art, photography, stage performance, and game art. Together, these properties enable us to derive intent-level questions about \emph{why} a creative choice was made, rather than only \emph{what} happens in a scene.

\subsection{Evaluation Scope and Question Design}
\label{sec:scope_design}
Inspired by~\citep{jarvinen2002gran}, we establish a hierarchical
capability taxonomy that drives both data collection and reporting to ensure comprehensive coverage across the audiovisual arts. At the top level, we identify four art categories (\textbf{Cinematic Arts}, \textbf{Static Visual Arts}, \textbf{Stage Performing Arts}, \textbf{Game Arts}) together with 11 sub-domains (see~\cref{sec:appendix:taxonomy}), informed by the canonical organization of creative disciplines and the artistic topics most actively discussed by expert video essayists. 
\cref{fig:category_examples} shows representative pairs from all four art categories.
See \cref{sec:appendix:add_examples} for more examples.

Within this taxonomy, \benchname combines two complementary question formats to capture the open-ended nature of artistic analysis. \textbf{Single-select} questions present a variable number of options (4 to 8) with exactly one correct answer and probe discrete recognition under a known-answer contract. 
\textbf{Multi-select} questions embed 2 to 4 correct answers among the options and probe set-valued analytical judgment, where the existence of multiple valid perspectives is itself part of the signal.
Single-select isolates whether a model can discriminate the right interpretation under certainty. Multi-select tests whether it can enumerate the full set of valid interpretations without over-claiming. In both formats, the evaluation instruction indicates whether the question is single-select or multi-select, but does not reveal the exact number of correct answers. 

\subsection{Data Collection and Question-Answer Annotation}
\label{sec:collection_annotation}

\begin{figure}[t]
    \centering
    \includegraphics[width=1.0\textwidth]{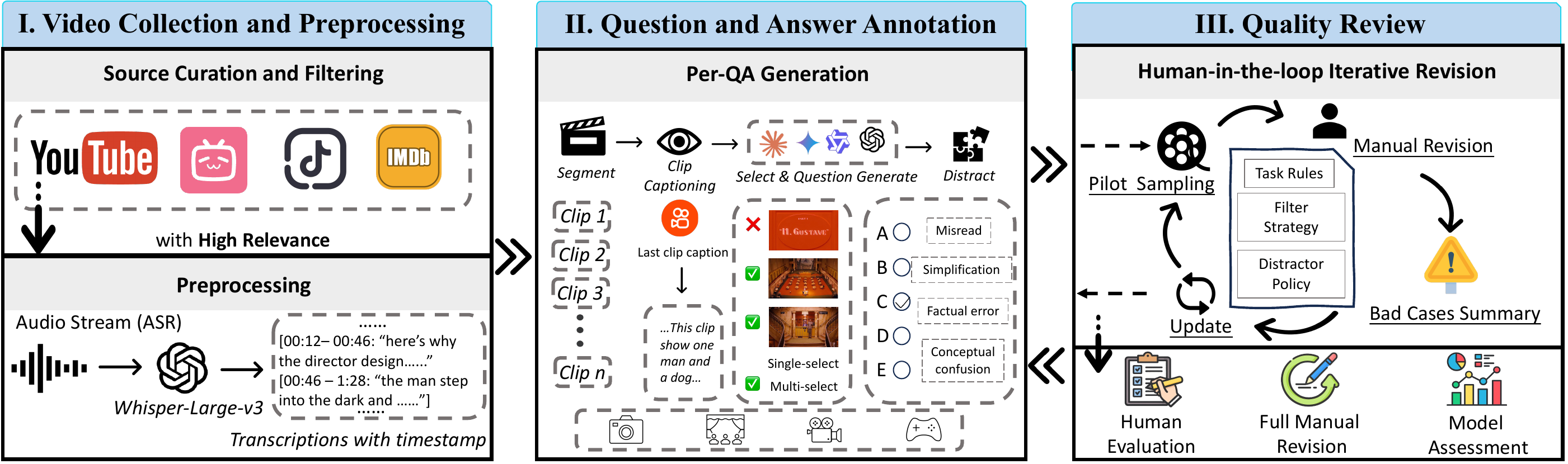}
    \vspace{-0.2in}
    \caption{\textbf{Construction pipeline of \benchname.} Panel~I curates video essays primarily from YouTube, Bilibili, and TikTok, applies relevance filtering against the audiovisual-arts taxonomy, and separates each retained video into two synchronized outputs with distinct roles, narrator transcripts for question construction and narrator-removed 10-second audiovisual clips for model evaluation. Panel~II generates candidate questions through four per-video phases, \emph{Segment}, \emph{Clip Captioning}, \emph{Select \& Question Generate}, and \emph{Distract}. Panel~III closes a human-in-the-loop revision cycle (\emph{Pilot Generation}, \emph{Manual Revision}, \emph{Bad Cases Summary}, \emph{Update}) that feeds a \emph{Full Regeneration} arrow back to Panel~II; each category exits once the stopping criterion is met.}
    \label{fig:pipeline}
    \vspace{-0.1in}
\end{figure}

\noindent \textbf{Video collection.}
Guided by the taxonomy above, we collect video essays primarily from YouTube, Bilibili, and TikTok that cover a broad range of expert commentary on the four art categories. We use GPT-5.4-mini~\citep{openai2026gpt54mini} to generate category-aware search keywords and assess candidate relevance, retaining only videos with substantial audiovisual-arts analysis.
Each retained video is then transcribed with Whisper-Large-v3~\citep{radford2023robust} to produce timestamped expert commentary aligned with the source video, supporting question generation and revision.

\noindent \textbf{Question-Answer Annotation.}
As illustrated in Panel~II of Figure~\ref{fig:pipeline},
candidate QA pairs
are generated through four successive phases shown in Figure~\ref{fig:pipeline}.
\\
\ding{182}~\textbf{Segment} (Figure~\ref{fig:pipeline}, Panel~II top): following~\citep{wang2025videoitg}, each video is partitioned into 10-second intervals to establish a uniform temporal granularity for subsequent analysis.

\ding{183}~\textbf{Clip Captioning} (Figure~\ref{fig:pipeline}, Panel~II second row): Keye-VL-1.5~\citep{yang2025kwai} samples each 10-second segment at 1\,fps and produces a single fine-grained caption per segment, conditioned on the temporally aligned narrator transcript. The captions cover visual attributes such as color, composition, motion and scene context. They are used solely as construction resources for downstream question generation and review, and are never exposed to models under evaluation.

\ding{184}~\textbf{Select \& Question Generate} (Figure~\ref{fig:pipeline}, Panel~II third row): the clip-level captions and full narrator transcripts are provided as inputs for generating 3 to 5 candidate questions per video in single-select and multi-select formats. For each candidate item, relevant evidence clips are first identified, after which the question prompt and correct answer are generated conditioned on those clips. 
The process follows two constraints:
(i)~the question must remain answerable solely from the narrator-removed evidence clips, 
 and (ii)~the correct answer is formulated prior to any distractor to mitigate stylistic or lexical saliency bias toward the correct option.

\ding{185}~\textbf{Distract} (Figure~\ref{fig:pipeline}, Panel~II bottom):
we generate plausible distractors using seven strategies: \emph{technical misread}, \emph{over-simplification}, \emph{concept swap}, \emph{cause-effect inversion}, \emph{scope error}, \emph{temporal confusion}, and \emph{partial truth} (\cref{sec:appendix:phase_d}). These strategies introduce errors in terminology, interpretation, causality, scope, or timing while keeping the alternatives plausible. Each item draws from multiple strategies, and every option is required to appear equally plausible to a reader who has not seen the clip. Single-select pairs receive 3--7 distractors (4--8 options total); multi-select pairs mix 2--4 correct options with distractors. We further forbid proper nouns, prohibit near-identical phrasing across distractors, and randomly shuffle option positions. Full details in \cref{sec:appendix:phase_a,sec:appendix:phase_b,sec:appendix:phase_c,sec:appendix:phase_d}.

\textbf{Copyright and release.} \benchname is intended for noncommercial research. Data collection followed applicable platform policies and copyright requirements. We excluded works whose terms prohibit redistribution and do not treat public availability as permission to redistribute. 
We plan to release QA pairs, annotations, evaluation code, and source pointers. Third-party media will be included only when redistribution is explicitly permitted by the applicable license or rights holder.
Each item records provenance, source information, evidence timestamps, and available licensing or attribution metadata, and is reviewed before release.
Items without redistribution permission will remain pointer-only.
Users obtain source media through the source host or other authorized channels and remain subject to the corresponding terms. We maintain a takedown procedure for creators and rightsholders and version the benchmark when items become unavailable or are removed. Full release and rights-review procedures are provided in Appendix~\ref{sec:appendix:release_protocol}.

\subsection{Quality Review}
\label{sec:hitl_loop}

All 4,016 final evaluation clips undergo manual inspection for embedded text, assisted by OCR. Narration subtitles and explanatory or answer-revealing overlays are removed or masked, while dialogue subtitles, game interfaces, and artwork-intrinsic text are retained. Appendix~\ref{sec:appendix:text_audit} reports the audit and paired masking controls.

\begin{wrapfigure}{r}{0.55\textwidth}
    \centering
    \vspace{-0.2in}
    \includegraphics[width=\linewidth]{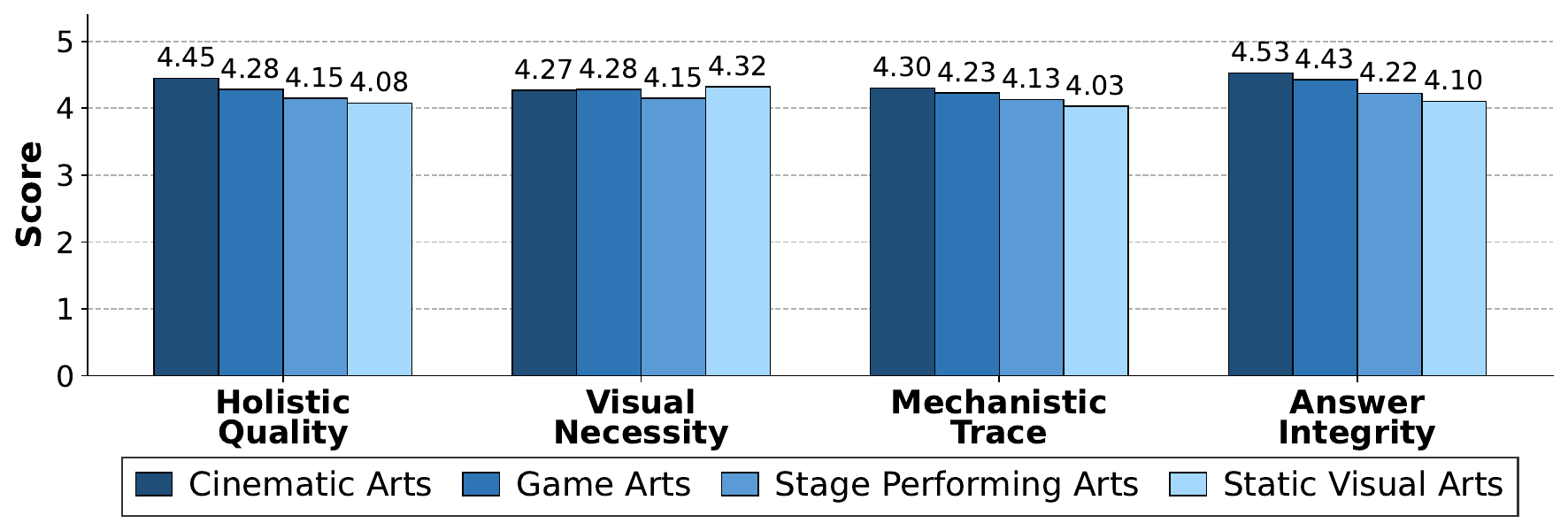}
    \vspace{-0.25in}
    \caption{Expert quality ratings on 90 sampled items across four dimensions (0--5 Likert scale). Four domain experts participate, with two matched experts independently rating each category. Per-category averages range from 4.03 to 4.53 (STD: 0.41).}
    \label{fig:human_eval_likert}
    \vspace{-0.1in}
\end{wrapfigure}

To ensure quality, we run the construction phases through an iterative review loop, where the in-context prompt is updated each round with new exclusion rules and domain-specific constraints.
 The loop is applied independently to each of the four art categories, with each round proceeding in four steps.
\ding{182}~\textbf{Pilot Generation}: a batch of candidate QA pairs is generated under the current prompt.
\ding{183}~\textbf{Manual Revision}: domain-expert reviewers assign binary pass/fail tags to each sampled QA pair under a shared failure taxonomy covering narrator-dependent answerability, ambiguous stems, weak or factually incorrect distractors, and misaligned clip references. Narrator-dependent cases are those answerable only from the expert transcript, not from the narrator-removed evaluation clips.
\ding{184}~\textbf{Bad Cases Summary}: we consolidate failure tags into a list of newly observed failure types for the round.
\ding{185}~\textbf{Update}: we rewrite the prompt with additional exclusion rules targeting new failures and then trigger a full regeneration. 

During review of an early draft containing $5{,}000$ QA pairs, around $9\%$ were flagged for severe errors in the CRITICAL or HIGH tiers. Across all severity levels, the audit identified eight tagged failure modes grouped into four tiers. The most severe tier covers hard schema violations such as labels pointing to nonexistent options, inline option lists in the stem disagreeing with the canonical options array, and multi-select pairs with an empty answer set. The lower tiers cover option-quality issues such as duplicated option texts, near-identical option prefixes, multi-select pairs that degenerate to single-select, and \texttt{correct\_answer} fields that paraphrase rather than reproduce the option string. Flagged severe cases were replaced from the QA pool; lower-severity cases were addressed through prompt refinement and programmatic alignment. Prompt rewriting also targeted seven systemic issues, including low option discriminability, overly academic wording, imprecise distractors, and inconsistent enforcement of the visual-evidence requirement.
Full details in \cref{sec:appendix:prompt_evolution,sec:appendix:failure_taxonomy}.
These construction-stage counts are distinct from the final-set option audit: among 400 stratified random items, 11 require wording refinements and three require equivalent-option corrections (Appendix~\ref{sec:appendix:option_audit}).

After generation, we manually verify every retained item before model assessment. To further validate the final benchmark, four domain experts are invited to independently rate a set of 90 samples along four quality dimensions on a 0--5 Likert scale (two assigned to Static Visual Arts and Game Arts, the other two to Stage Performing Arts and Cinematic Arts); the resulting per-category averages exceed 4.0 across every dimension with an average Inter-Annotator Agreement of Gwet AC2~\citep{james2026counting, popplewell2019appropriate} $= 0.855$, indicating near-perfect consistency across raters, as summarized in \cref{fig:human_eval_likert}. Rater backgrounds, assignments, and the scoring rubric are detailed in \cref{sec:appendix:human_eval_details}.

Quality-rating AC2 measures agreement on the four rating dimensions. A separate, overlapping 320-item answer study yields 86.88\% single-answer agreement and 68.75\% multi-answer exact-set agreement; expert adjudication confirms 303 answer sets and updates 17 (Appendix~\ref{sec:appendix:answer_agreement}).

\subsection{Evaluation Metrics}
\label{sec:metrics}

Metrics are computed over the valid questions included in each evaluation. Overall ACC uses all valid scored items; CAA and EM use the corresponding single-select and multi-select subsets. Category scores use the valid items within that category.

Single-select questions contain $K_i \in \{4,...,8\}$ options, so uniform-random accuracy varies as $1/K_i$. Raw accuracy therefore mixes model performance with item-specific guessing probability. We report chance-adjusted accuracy (CAA), which subtracts the per-item chance baseline and rescales the score:
\begin{equation}
    \mathrm{CAA}_i = \frac{a_i - 1/K_i}{1 - 1/K_i},
    \qquad
    \mathrm{CAA} = \frac{1}{N_{\mathrm{single}}}\sum_{i=1}^{N_{\mathrm{single}}} \mathrm{CAA}_i,
    \label{eq:caa}
\end{equation}
where $a_i \in \{0,1\}$ indicates correctness and $N_{\mathrm{single}}$ is the number of valid single-select questions included in the evaluation. CAA is normalized to 0 under uniform guessing and 1 for a correct response; negative aggregate scores indicate worse-than-chance performance. This enables fair comparison across subdomains with different option-count distributions. For multi-select questions, exact-match accuracy (EM) requires the predicted option set to match the ground truth exactly, measuring complete coverage of valid answers without selecting distractors. CAA and EM are the primary format-specific metrics. Overall ACC summarizes fully correct answers across both formats, with equal weight per question and no chance adjustment. Precision, recall, and F1 diagnose incorrect selections, missed answers, and partial agreement, respectively. Complete definitions and supplementary metrics are in \cref{sec:appendix:eval_metrics_details}.

Human responses use the same item-level scoring: $a_i^{\mathrm{human}}=\mathbf{1}[h_i=y_i]$, where $h_i$ is the assigned expert's answer and $y_i$ the reference answer. CAA is computed per question before averaging, using its actual $K_i$. Appendix~\ref{sec:appendix:human_caa} gives the explicit aggregation.

\subsection{Benchmark Statistics}
\benchname contains $4{,}016$ expert-validated questions across four art categories and 11 sub-domains, drawn from over $10{,}000$ candidate video essays. Each retained video contributes $3$--$5$ questions with $4$--$8$ options each (\Cref{tab:dataset_stats}). Distributions by sub-domain, source platform, language, and cultural or regional provenance are provided in \cref{sec:appendix:composition}.

%% file: sections/04_exp.tex
\section{Experiments}
\label{sec:exp}
\subsection{Experimental Setup}
\label{sec:exp_setup}
We evaluate 34 MLLM versions zero-shot, counting each version once across input settings. The models include proprietary, open source general-purpose, and video-specific systems, including five with dynamic key frame selection. Narrator-removed clips are sampled at 1\,fps within each model's frame limit. Prompts identify the question as single-select or multi-select without revealing the number of correct options.

We compare two input settings. V+T provides visual inputs and an audio transcript together with the task instruction, question, and options. V+A+T provides native audio in place of the transcript. Expert narration and construction captions are withheld from both settings. Tables~\ref{tab:evaluation_results} and~\ref{tab:audio_results} report the two leaderboards. The benchmark contains 4,016 questions. Scores are computed over the valid items included in each model and input setting's evaluation, with format-specific and category-specific denominators. Table~\ref{tab:paired_audio_results} reports the evaluated sample sizes for its input-setting comparisons. Appendix~\ref{sec:appendix:input_routes} details model access and input processing.

\subsection{Main Results}
\label{sec:main_results}

\Cref{tab:evaluation_results,tab:audio_results} report zero-shot results; \cref{sec:appendix:add_exp} provides full per-category breakdowns.
\input{tables/main_results_table}

Appendix~\ref{sec:appendix:input_routes} provides the input-setting comparisons and their category-level analysis (Figure~\ref{fig:input_setting_comparison}).

\noindent \textbf{Finding 1. MLLMs trail expert performance on audiovisual arts reasoning.} Across the evaluation, Gemini-3.8-Flash achieves the highest overall ACC in both input settings: 59.11\% with V+T (Table~\ref{tab:evaluation_results}) and 64.62\% with V+A+T (Table~\ref{tab:audio_results}). The latter falls 22.56 percentage points short of the 87.18\% expert baseline. General-purpose models match or surpass video-specialized models, suggesting that video processing alone is insufficient for grounded artistic inference.

\begin{figure}[t]
\centering
\includegraphics[width=0.85\textwidth]{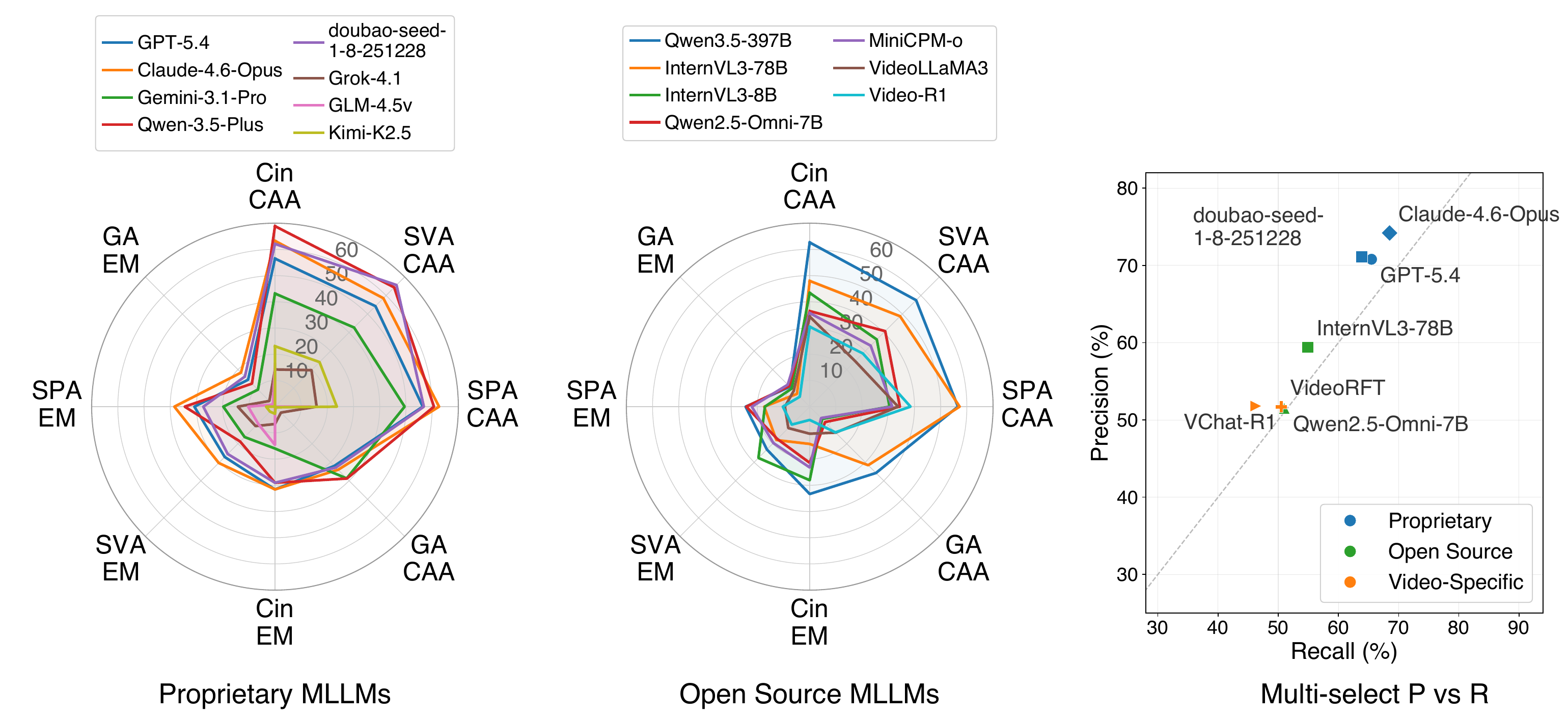}
\caption{Original evaluations: per-category Single CAA (upper axes) and Multi EM (lower axes) for proprietary (left) and open-source (middle) models; multi-select precision versus recall (right). Cin/SVA/SPA/GA denote Cinematic/Static Visual/Stage Performing/Game Arts.}
\label{fig:performance_summary}
\vspace{-0.1in}
\end{figure}

\noindent \textbf{Finding 2. Category strengths depend on the input setting.} With native audio, Game Arts has the lowest CAA and EM for every model in Table~\ref{tab:audio_results}. Gemini-3.8-Flash leads Multi EM in all four categories, while Qwen3.8-Omni-Flash leads Cinematic Single CAA. With transcript-based V+T, Gemini-3.8-Flash, MiMo-V2.6-Flash, and Muse-Ember instead achieve their highest category CAA on Game Arts (Table~\ref{tab:evaluation_results}). Native audio improves overall performance, but the category profiles show that its benefit depends on both the model and the artistic domain. Figure~\ref{fig:performance_summary} summarizes category profiles and multi-select precision and recall for the original evaluations.

\input{tables/audio_results_table}

\subsection{In-Depth Analysis}
\label{sec:in_depth_analysis}

\noindent \textbf{Finding 3. Adaptive frame-selection models trail the strongest video model.} Five models adaptively select frames: AKS~\citep{tang2025adaptive}, Q-Frame~\citep{zhang2025q}, LongVT~\citep{yang2025longvt}, Video-CCAM~\citep{fei2024video}, and TimeChat~\citep{ren2024timechat}. They reach 14.42--20.52 ACC, compared with 27.19 for VideoLLaMA3, a gap of 6.67--12.77 points. This comparison identifies artistic interpretation as a challenge for these adaptive frame-selection models.

\noindent \textbf{Finding 4. Partial agreement does not guarantee complete interpretation.} On multi-select questions, precision and recall expose distinct error modes (Table~\ref{tab:app_evaluation_results}). Claude-4.6-Opus has higher precision than recall across all categories, indicating missed valid interpretations. LongVU has higher recall than precision, pointing to excess selections. For example, on Game Arts, LongVU reaches 91.65\% recall and 67.64\% F1 but only 1.72\% EM; Claude-4.6-Opus has lower recall (55.78\%) and F1 (59.59\%) yet substantially higher EM (18.36\%) (Tables~\ref{tab:evaluation_results} and~\ref{tab:app_evaluation_results}). Broad coverage can yield high F1 scores despite failing to isolate the exact correct option set. Together, precision, recall, and EM distinguish between omitted interpretations and over-selection.

Answer format also changes comparisons within a family. InternVL3-78B surpasses InternVL3-8B in overall ACC (37.80 versus 33.07) and single-select CAA (47.03 versus 29.49), but falls behind in multi-select EM (13.53 versus 20.30). Its advantage in selecting one answer does not extend to recovering the complete answer set.

\noindent \textbf{Finding 5. Native audio improves accuracy across question formats.} MiniCPM-o-4.5~\citep{yu2026minicpm} and Gemma-4-E4B~\citep{team2024gemma} achieve higher ACC, Single CAA, and Multi EM with native audio than with transcripts (Appendix Table~\ref{tab:modality_ablation}).

\newpage
\begin{wrapfigure}{r}{0.5\textwidth}
\centering
\vspace{-2em}
\includegraphics[width=\linewidth]{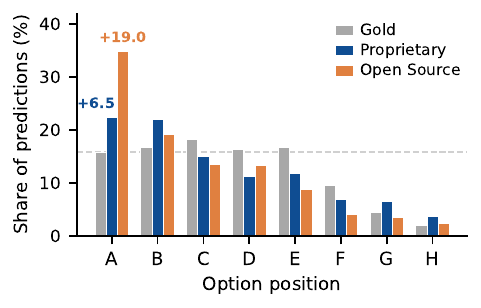}
\vspace{-2.5em}
\caption{Option position bias in the 28-model analysis ($n{=}1{,}407$ single-select questions with at least five choices).}
\label{fig:position_bias}
\vspace{-1.5em}
\end{wrapfigure}
\noindent \textbf{Finding 6. Open-source MLLMs exhibit a pronounced first-position bias.}
In a position-bias analysis of 28 models on the 1{,}407 single-select questions with five or more choices (\Cref{fig:position_bias}), the gold labels are approximately uniformly distributed between positions A through E at 15.9--18.2\% each, yet the model predictions focus heavily on the early positions. Position A alone accounts for 30.9\% of all model predictions, roughly double its 15.9\% gold share, while positions E through H collectively receive only 22.2\% of predictions against a 33.0\% gold share.
This front-loading is markedly stronger among open source MLLMs, whose 34.9\% A-prediction rate exceeds the gold share by 19.0 points; proprietary models, by contrast, predict A at 22.4\%, only 6.5 points above the gold baseline.
The prediction distributions show first-position over-selection relative to the benchmark's randomized gold labels, with a larger excess among open source models.
Additional evidence controls and answer-validation results are reported in Appendix~\ref{sec:evidence_validation}.

%% file: tables/main_results_table.tex
\begin{table*}[t]
\centering
\caption{Zero-shot V+T results on \benchname (\%). T includes the task, question, options, and audio transcript. Single-select uses chance-adjusted accuracy (CAA), multi-select uses exact match (EM), and overall ACC is the percentage of questions answered exactly. CAA bootstrap standard error is about 1.3 points. \textbf{Bold}/\underline{underlined}: best/second-best model scores per column, excluding random and human baselines. Scores use each evaluation's valid items, with denominators specific to the question format and category; Table~\ref{tab:paired_audio_results} gives sample sizes for its comparisons. Native-audio results appear in Table~\ref{tab:audio_results}.}
\label{tab:evaluation_results}
\vspace{-0.1in}
\resizebox{\textwidth}{!}{%
\begin{tabular}{@{}l|c|c>{\columncolor{mymultiem}}c|c>{\columncolor{mymultiem}}c|c>{\columncolor{mymultiem}}c|c>{\columncolor{mymultiem}}c|c>{\columncolor{mymultiem}[\tabcolsep][0pt]}c@{}}
\toprule
\multirow{3}{*}{\textbf{Model}}
  & \multirow{3}{*}{\textbf{Overall}}
  & \multicolumn{2}{c|}{\textbf{Overall}}
  & \multicolumn{2}{c|}{\textbf{Cinematic Arts}}
  & \multicolumn{2}{c|}{\textbf{Static Visual Arts}}
  & \multicolumn{2}{c|}{\textbf{Stage Perf. Arts}}
  & \multicolumn{2}{c}{\textbf{Game Arts}} \\
\cmidrule(lr){3-4} \cmidrule(lr){5-6} \cmidrule(lr){7-8} \cmidrule(lr){9-10} \cmidrule(lr){11-12}
  &
  & \textbf{Single} & \textbf{Multi}
  & \textbf{Single} & \textbf{Multi}
  & \textbf{Single} & \textbf{Multi}
  & \textbf{Single} & \textbf{Multi}
  & \textbf{Single} & \textbf{Multi} \\
  & \textbf{ACC}
  & \textbf{CAA} & \textbf{EM}
  & \textbf{CAA} & \textbf{EM}
  & \textbf{CAA} & \textbf{EM}
  & \textbf{CAA} & \textbf{EM}
  & \textbf{CAA} & \textbf{EM} \\
\midrule
Random & 13.55 & 0.02 & 6.04 & 0.04 & 5.85 & -0.04 & 4.47 & 0.02 & 6.34 & 0.06 & 6.37 \\
Human Expert & 87.18 & 90.98 & 78.00 & 98.74 & 86.42 & 90.13 & 76.18 & 89.42 & 70.55 & 86.15 & 78.83 \\
\midrule
gemini-3.8-flash~\citep{team2023gemini} & \textbf{59.11} & \textbf{63.50} & \textbf{46.93} & 60.12 & \textbf{48.64} & 62.14 & \textbf{46.93} & \textbf{65.57} & \underline{43.42} & \textbf{65.94} & \textbf{47.80} \\
qwen3.8-omni-flash~\citep{team2026qwen3} & \underline{54.82} & \underline{60.54} & \underline{40.31} & 58.20 & \underline{41.10} & 57.78 & \underline{38.07} & 61.55 & \textbf{44.03} & \underline{64.99} & \underline{38.46} \\
Claude-4.6-Opus~\citep{anthropic2026opus46} & 48.28 & 55.13 & 28.91 & \underline{63.26} & 31.58 & 58.51 & 30.38 & \underline{62.65} & 38.42 & 34.07 & 18.36 \\
Qwen-3.5-Plus~\citep{bai2025qwen3} & 47.26 & 58.52 & 23.21 & \textbf{68.88} & 29.04 & \underline{64.36} & 18.82 & 60.69 & 34.47 & 38.80 & 12.43 \\
doubao-seed-1-8-251228~\citep{seed2025seed18} & 46.12 & 55.00 & 24.22 & 62.10 & 29.04 & \textbf{65.63} & 25.54 & 56.84 & 27.37 & 32.86 & 16.25 \\
grok-4.3~\citep{xai2026grok43} & 45.99 & 51.18 & 27.26 & 55.31 & 28.29 & 53.31 & 34.13 & 62.23 & 32.11 & 31.29 & 17.78 \\
muse-ember~\citep{meta2026museember} & 45.28 & 46.66 & 32.76 & 42.15 & 34.56 & 41.53 & 30.13 & 47.48 & 34.21 & 56.35 & 31.79 \\
Qwen3.5-397B-A17B~\citep{bai2025qwen3} & 44.77 & 53.42 & 22.71 & 62.65 & 33.33 & 57.45 & 23.12 & 56.60 & 24.47 & 35.79 & 10.71 \\
GPT-5.4~\citep{singh2025openai} & 44.57 & 50.28 & 25.50 & 56.50 & 31.58 & 54.24 & 27.15 & 56.43 & 30.79 & 32.00 & 14.53 \\
mimo-v2.6-flash~\citep{qiao2026mimo} & 43.40 & 47.39 & 27.87 & 42.93 & 28.29 & 41.33 & 27.20 & 49.95 & 27.37 & 56.10 & 28.30 \\
doubao-seed-2-0-lite-260428~\citep{seed2026seed2} & 41.86 & 39.82 & 29.26 & 40.88 & 30.62 & 38.67 & 36.53 & 45.36 & 33.42 & 33.55 & 19.69 \\
InternVL3-78B~\citep{chen2024expanding} & 37.80 & 47.03 & 13.53 & 47.98 & 14.23 & 48.79 & 18.01 & 57.25 & 17.37 & 31.59 & 6.88 \\
InternVL3-8B~\citep{chen2024expanding} & 33.07 & 29.49 & 20.30 & 43.41 & 28.07 & 36.23 & 27.69 & 30.48 & 17.11 & 6.66 & 9.75 \\
VideoLLaMA3~\citep{zhang2025videollama} & 27.19 & 26.82 & 9.90 & 34.37 & 10.33 & 24.55 & 11.56 & 33.76 & 9.47 & 14.02 & 8.60 \\
Video-R1~\citep{feng2025video} & 26.72 & 28.41 & 7.21 & 30.50 & 5.07 & 28.70 & 9.68 & 38.49 & 10.26 & 13.87 & 5.35 \\
VideoRFT~\citep{wang2025videorft} & 26.12 & 26.17 & 8.17 & 26.50 & 7.99 & 30.14 & 10.75 & 36.04 & 8.68 & 9.01 & 6.12 \\
VideoChat-R1~\citep{li2025videochat} & 26.07 & 26.49 & 7.77 & 35.87 & 5.26 & 29.05 & 12.10 & 33.04 & 6.05 & 6.46 & 8.41 \\
Video-XL-2~\citep{Shu2024VideoXLEV} & 24.18 & 29.91 & 0.11 & 28.63 & 0.00 & 29.29 & 0.54 & 40.42 & 0.00 & 19.17 & 0.00 \\
Grok-4.1~\citep{grok_4} & 20.54 & 13.71 & 8.00 & 14.19 & 6.63 & 19.70 & 10.48 & 15.90 & 14.21 & 3.20 & 3.06 \\
LongVT~\citep{yang2025longvt} & 20.52 & 17.14 & 4.64 & 17.99 & 2.73 & 21.61 & 9.14 & 21.92 & 6.32 & 5.02 & 2.10 \\
LLaVA-OneVision-7B~\citep{li2024llava} & 20.42 & 21.24 & 0.50 & 22.01 & 0.00 & 25.77 & 1.88 & 25.09 & 0.00 & 10.24 & 0.38 \\
Kimi-K2.5~\citep{team2026kimi} & 19.92 & 18.33 & 2.07 & 23.06 & 2.73 & 24.05 & 2.69 & 23.62 & 3.42 & 0.35 & 0.00 \\
AKS~\citep{tang2025adaptive} & 19.30 & 18.99 & 0.00 & 17.61 & 0.00 & 21.46 & 0.00 & 28.01 & 0.00 & 6.34 & 0.00 \\
Q-Frame~\citep{zhang2025q} & 18.75 & 9.65 & 8.05 & 13.81 & 9.55 & 13.83 & 9.68 & 3.30 & 9.21 & 8.19 & 4.59 \\
VideoChat2~\citep{li2024mvbench} & 17.78 & 15.27 & 0.34 & 17.20 & 0.19 & 14.35 & 0.00 & 18.67 & 0.79 & 10.45 & 0.38 \\
Video-CCAM~\citep{fei2024video} & 17.53 & 15.10 & 0.00 & 23.38 & 0.00 & 18.31 & 0.00 & 16.92 & 0.00 & 1.05 & 0.00 \\
GLM-4.5v~\citep{hong2025glm} & 17.13 & 5.43 & 8.61 & 16.17 & 14.62 & -2.97 & 9.14 & 13.60 & 10.26 & -4.34 & 1.15 \\
LongVU~\citep{shen2024longvu} & 14.87 & 8.21 & 1.01 & 14.40 & 0.19 & 6.50 & 1.34 & 5.46 & 0.79 & 7.75 & 1.72 \\
TimeChat~\citep{ren2024timechat} & 14.42 & 7.79 & 0.34 & 12.27 & 0.58 & 9.66 & 0.00 & 4.61 & 0.26 & 5.04 & 0.38 \\
\bottomrule
\end{tabular}%
}
\vspace{-0.1in}
\end{table*}

%% file: tables/audio_results_table.tex
\begin{table*}[t]
\centering
\caption{Native-audio V+A+T results on \benchname (\%); metrics follow Table~\ref{tab:evaluation_results}. Models receive video, native audio, and text without transcripts. \textbf{Bold}/\underline{underlined}: best/second-best scores per column. Scores use each evaluation's valid items, with denominators specific to the question format and category; Table~\ref{tab:paired_audio_results} gives sample sizes for its comparisons.}
\label{tab:audio_results}
\vspace{-0.1in}
\resizebox{\textwidth}{!}{%
\begin{tabular}{@{}l|c|c>{\columncolor{mymultiem}}c|c>{\columncolor{mymultiem}}c|c>{\columncolor{mymultiem}}c|c>{\columncolor{mymultiem}}c|c>{\columncolor{mymultiem}[\tabcolsep][0pt]}c@{}}
\toprule
\multirow{3}{*}{\textbf{Model}}
  & \multirow{3}{*}{\textbf{Overall}}
  & \multicolumn{2}{c|}{\textbf{Overall}}
  & \multicolumn{2}{c|}{\textbf{Cinematic Arts}}
  & \multicolumn{2}{c|}{\textbf{Static Visual Arts}}
  & \multicolumn{2}{c|}{\textbf{Stage Perf. Arts}}
  & \multicolumn{2}{c}{\textbf{Game Arts}} \\
\cmidrule(lr){3-4} \cmidrule(lr){5-6} \cmidrule(lr){7-8} \cmidrule(lr){9-10} \cmidrule(lr){11-12}
  &
  & \textbf{Single} & \textbf{Multi}
  & \textbf{Single} & \textbf{Multi}
  & \textbf{Single} & \textbf{Multi}
  & \textbf{Single} & \textbf{Multi}
  & \textbf{Single} & \textbf{Multi} \\
  & \textbf{ACC}
  & \textbf{CAA} & \textbf{EM}
  & \textbf{CAA} & \textbf{EM}
  & \textbf{CAA} & \textbf{EM}
  & \textbf{CAA} & \textbf{EM}
  & \textbf{CAA} & \textbf{EM} \\
\midrule
gemini-3.8-flash~\citep{team2023gemini} & \textbf{64.62} & \textbf{68.57} & \textbf{51.90} & \underline{67.57} & \textbf{57.95} & \textbf{66.06} & \textbf{57.60} & \textbf{77.03} & \textbf{61.58} & \textbf{62.44} & \textbf{34.80} \\
qwen3.8-omni-flash~\citep{team2026qwen3} & \underline{61.62} & \underline{67.71} & \underline{45.93} & \textbf{69.97} & \underline{52.84} & \underline{64.66} & \underline{50.40} & \underline{76.78} & \underline{53.05} & \underline{58.47} & \underline{30.77} \\
mimo-v2.6-flash~\citep{qiao2026mimo} & 48.73 & 55.12 & 29.54 & 58.10 & 31.20 & 55.60 & 35.47 & 64.67 & 33.68 & 40.18 & 20.65 \\
muse-ember~\citep{meta2026museember} & 48.61 & 48.94 & 35.40 & 51.93 & 38.37 & 48.09 & 41.07 & 57.60 & 46.05 & 36.68 & 20.65 \\
doubao-seed-2-0-lite-260428~\citep{seed2026seed2} & 47.61 & 48.60 & 33.50 & 48.76 & 39.53 & 48.53 & 38.13 & 55.98 & 36.84 & 39.63 & 21.80 \\
Gemini-3.1-pro-preview~\citep{team2023gemini} & 36.90 & 43.77 & 14.88 & 43.16 & 15.98 & 42.70 & 16.40 & 49.50 & 19.74 & 38.72 & 9.18 \\
Qwen2.5-Omni-7B~\citep{xu2025qwen25omni} & 32.69 & 30.71 & 18.18 & 36.47 & 21.44 & 40.76 & 17.47 & 34.43 & 24.21 & 8.31 & 11.09 \\
MiniCPM-o-4.5~\citep{yu2026minicpm} & 31.35 & 27.05 & 18.90 & 35.72 & 23.20 & 32.92 & 19.62 & 31.49 & 22.11 & 6.14 & 11.85 \\
Gemma-4-E4B~\citep{team2024gemma} & 27.61 & 28.67 & 9.06 & 39.40 & 11.89 & 32.55 & 13.17 & 31.44 & 8.68 & 10.30 & 3.63 \\
VideoLLaMA2.1-AV~\citep{cheng2024videollama} & 20.34 & 20.07 & 1.17 & 31.35 & 1.95 & 18.07 & 1.61 & 25.34 & 1.32 & 5.40 & 0.00 \\
\bottomrule
\end{tabular}%
}
\vspace{-0.1in}
\end{table*}

%% file: sections/05_con.tex
\section{Conclusion}
\label{sec:Conclusion_limitations}
We introduce \benchname, a benchmark with $4{,}016$ expert-validated questions across four art categories, constructed from video essays through a four-phase pipeline with human review. Across 34 evaluated MLLM versions, Gemini-3.8-Flash achieves the highest Overall ACC: 59.11\% with V+T and 64.62\% with V+A+T. This best performance remains 22.56 percentage points below the 87.18\% expert baseline. Analysis of the V+T and V+A+T input settings reveals distinct domain profiles: native-audio models consistently score lowest on game arts. Furthermore, models struggle to identify complete multi-select option sets, and the five adaptive frame-selection models score below VideoLLaMA3. These findings motivate models that connect audiovisual evidence to artistic knowledge and recover complete sets of supported interpretations.

%% file: sections/99_appendix.tex
\section{Pointer-First Release Plan, Provenance, and Rights}
\label{sec:appendix:release_protocol}

\noindent \textbf{Pointer-first public release.}
We plan to release \benchname and the associated evaluation code on Hugging Face under a \emph{pointer-first} protocol for noncommercial research use. The package will contain question--answer pairs, benchmark annotations, source pointers (source URLs or platform video IDs), evidence timestamps, and item-level provenance and attribution records. Third-party content, including videos, clips, frames, audio, transcripts, thumbnails, platform API responses, or cached copies, will be included only when redistribution is explicitly permitted by the applicable license or rightsholder; otherwise, the corresponding item will remain pointer-only. The intermediate pool of more than 10{,}000 candidate videos will not be released.

\noindent \textbf{Source eligibility and review.}
During source curation, we excluded sources whose stated terms were incompatible with our collection or release protocol, and we did not treat public availability as permission to redistribute. Each item is reviewed before release, taking into account available license information, creator permissions, and attribution requirements for the source video and, where identifiable, the underlying works it discusses.

\noindent \textbf{Access to source media.}
Users obtain source media through the source host or other official or authorized channels, such as the platform's own interface or approved APIs, under the corresponding terms of service and licensing conditions.

\noindent \textbf{Provenance and attribution.}
Each released item records its source pointer, platform, creator or channel, evidence timestamps, available license or attribution metadata, and, when identifiable, the underlying film, game, performance, artwork, or other work. Rights in the linked media and underlying works remain with their respective rightsholders.

\noindent \textbf{Copyright and platform compliance.}
Data collection and benchmark construction were conducted with consideration of applicable platform policies and copyright requirements. The benchmark is intended for noncommercial research. The public benchmark is pointer-first; third-party media are included only when redistribution is explicitly permitted.

\noindent \textbf{Correction and takedown.}
The release documentation will describe how creators and rightsholders can request correction or removal of an item. When a request is accepted, we will remove the corresponding pointer and annotations, record the change in the release notes, and not provide a cached substitute.

\noindent \textbf{Versioning and continued availability.}
Because linked media may be edited, made private, or removed by their hosts, the set of evaluable items may change over time. We will maintain versioned releases with a changelog recording added, corrected, removed, and unavailable items. 
Items that become unavailable will be flagged in subsequent versions. Items removed at a rightsholder’s request will no longer be distributed in any hosted release, while the changelog will preserve the record that the item was removed.
We recommend reporting results together with the benchmark version and the number of items evaluated, so that results remain comparable across versions.

\section{Limitation and Future Work}
\label{sec:appendix:limitations}
\benchname evaluates four art categories using video essays as the primary data source. While these essays provide rich expert commentary, they may not capture the full diversity of artistic expression present in raw creative works, and their availability varies across disciplines and languages. Future work should expand to additional art forms (e.g., music and architecture), incorporate diverse multilingual sources, develop open-ended evaluations beyond multiple-choice formats, and investigate whether targeted fine-tuning on arts-specific data can close the performance gap between models and human experts.

\section{Benchmark Details}
\label{sec:appendix:benchmark_details}

\subsection{Comparison with Existing Benchmarks}
\label{sec:appendix:comparison}

\input{tables/comparison_table}

\Cref{tab:comparison} situates \benchname against representative video-understanding benchmarks. Existing datasets typically focus on short clips of everyday activities or open-domain factual QA, often without coupling audio and subtitles, enforcing domain expertise, or controlling for visual dependency. Recent expert-oriented benchmarks like Video-MMMU~\citep{Hu2025VideoMMMUEK} introduce multi-level difficulty for lecture videos but target STEM knowledge transfer rather than artistic interpretation, and do not require joint audio-subtitle reasoning.

\benchname examines artistic interpretation across four categories using audiovisual evidence and domain knowledge. These properties stem directly from our construction pipeline (\cref{sec:design_construction}): the hierarchical taxonomy spanning four categories and 11 sub-domains organizes the required artistic knowledge, while pairing narrator-removed clips with timestamped expert commentary (\cref{sec:collection_annotation}) supports questions about audiovisual evidence. Together, these design choices establish a targeted evaluation for expert-level analytical understanding of the audiovisual arts.

\input{sections/revision_comparison}

\subsection{Category and Sub-domain Definitions}
\label{sec:appendix:taxonomy}

\benchname organizes 4{,}016 questions under a hierarchical capability taxonomy: four top-level art categories, each partitioned into fine-grained sub-domains for a total of 11 sub-domains. The taxonomy is grounded in the canonical organization of creative disciplines and the topics most actively analyzed by expert video essayists.

\paragraph{Cinematic Arts.}
This category covers the analytical study of moving-image works, including narrative film, television drama, documentary, and animation. Items probe how filmmakers translate dramatic intent into image and sound through four sub-domains.
\begin{itemize}[leftmargin=*, itemsep=2pt, topsep=2pt]
    \item \textbf{Cinematography} -- shot composition, camera movement, framing, lensing, depth of field, lighting, and exposure choices that shape how a scene is seen.
    \item \textbf{Mise-en-sc\`ene} -- on-screen staging including blocking, props, set design, costume, and spatial composition that organize the world inside the frame.
    \item \textbf{Editing and Pacing} -- shot-to-shot construction, including cuts, transitions, montage, parallel action, and the rhythm imposed by edit length.
    \item \textbf{Sound} -- diegetic and non-diegetic audio, score, ambient sound, silence, and sound bridges that anchor or extend the visual narrative.
\end{itemize}

\paragraph{Static Visual Arts.}
This category covers analyses of still images and tangible artifacts, where temporal structure is replaced by compositional and material reasoning.
Material reasoning includes how an artist's working process shapes the visible form of a work. In Figure~\ref{fig:appendix_example_vis_04}, variation among the depicted plant forms connects exploratory material selection to the resulting artwork. The intended visual evidence is that variation, which the question links to the artist's working method.
\begin{itemize}[leftmargin=*, itemsep=2pt, topsep=2pt]
    \item \textbf{Fine Art} -- painting, drawing, printmaking, and sculpture, focusing on composition, color theory, perspective, brushwork, chiaroscuro, and material technique.
    \item \textbf{Photography} -- documentary, fine-art, and commercial photography, focusing on exposure, aperture, lensing, framing, and the photographer's documentary intent.
\end{itemize}

\paragraph{Stage Performing Arts.}
This category covers live performance forms in which meaning is constructed through performer presence, staged space, and time-bound audience reception.
\begin{itemize}[leftmargin=*, itemsep=2pt, topsep=2pt]
    \item \textbf{Performance} -- acting, vocal delivery, physical gesture, body language, and the live communicative work of the performer, including drama, dance, and stand-up.
    \item \textbf{Stage Design} -- the spatial language of the stage, including set design, scenery, props, platforms, and the blocking that organizes performers within them.
    \item \textbf{Theatrical Lighting} -- the artistic use of stage lighting, including spot, follow spot, side and back lighting, color washes, and warm/cool contrasts that direct attention and shape mood.
\end{itemize}

\paragraph{Game Arts.}
This category covers the audiovisual craft of video games, where artistic choices co-exist with interactive systems and player agency.
\begin{itemize}[leftmargin=*, itemsep=2pt, topsep=2pt]
    \item \textbf{CG} -- the rendered visual language of games, including art style, real-time and pre-rendered cinematics, shading and lighting models, and post-processing aesthetics.
    \item \textbf{Interactive Visuals} -- visual elements bound to player interaction, including level design, environmental storytelling, on-screen guidance and HUD, navigation cues, and gameplay-conditioned camera and animation.
\end{itemize}

\subsection{Dataset Statistics}
\label{sec:statistics}

\Cref{tab:dataset_stats} consolidates the statistics that characterize \benchname, covering global scope, source data, question format.

\begin{table}[!htbp]
\centering
\caption{Comprehensive statistics of \benchname, covering global properties and construction parameters.}
\label{tab:dataset_stats}
\small
\begin{tabular}{@{}ll@{}}
\toprule
\textbf{Property} & \textbf{Value} \\
\midrule
Total questions                              & 4{,}016 \\
Art categories                               & 4 (cinematic, static visual, stage performing, game) \\
Sub-domains                                  & 11 \\
Candidate source video essays                & $>$10{,}000 \\
Sampling rate during captioning              & 1\,fps \\
Questions per video                          & 3 to 5 \\
Options per question                         & 4 to 8 \\
\bottomrule
\end{tabular}

\end{table}

\subsection{Composition by Sub-domain and Source}
\label{sec:appendix:composition}

The distribution below covers all 4{,}016 retained questions. We report source language separately from the cultural or regional provenance of the work discussed in a video essay, since narration language does not necessarily identify that work's origin.

\begin{table}[htbp]
\centering
\caption{Question counts by art category and sub-domain.}
\label{tab:subdomain_distribution}
\footnotesize
\setlength{\tabcolsep}{3pt}
\begin{tabular}{@{}llrrrr@{}}
\toprule
\textbf{Category} & \textbf{Sub-domain} & \textbf{Single} & \textbf{Multi} & \textbf{Total} & \textbf{Share (\%)} \\
\midrule
Cinematic Arts & Cinematography & 204 & 219 & 423 & 10.53 \\
Cinematic Arts & Mise-en-sc\`ene & 85 & 83 & 168 & 4.18 \\
Cinematic Arts & Editing and Pacing & 93 & 97 & 190 & 4.73 \\
Cinematic Arts & Sound & 102 & 117 & 219 & 5.45 \\
Static Visual Arts & Fine Art & 363 & 213 & 576 & 14.34 \\
Static Visual Arts & Photography & 262 & 162 & 424 & 10.56 \\
Stage Performing Arts & Performance & 315 & 197 & 512 & 12.75 \\
Stage Performing Arts & Stage Design & 151 & 97 & 248 & 6.18 \\
Stage Performing Arts & Theatrical Lighting & 142 & 86 & 228 & 5.68 \\
Game Arts & CG & 195 & 211 & 406 & 10.11 \\
Game Arts & Interactive Visuals & 310 & 312 & 622 & 15.49 \\
\midrule
Total & 11 sub-domains & 2{,}222 & 1{,}794 & 4{,}016 & 100.00 \\
\bottomrule
\end{tabular}
\end{table}

\begin{table}[htbp]
\centering
\caption{Distribution of retained questions by source platform, narration language, cultural or regional provenance, and genre. Each dimension totals 4{,}016 questions; percentages are rounded.}
\label{tab:source_distribution}
\small
\begin{tabular}{@{}llrr@{}}
\toprule
\textbf{Dimension} & \textbf{Group} & \textbf{Questions} & \textbf{Share (\%)} \\
\midrule
Platform & YouTube & 2{,}073 & 51.62 \\
 & Bilibili & 1{,}696 & 42.23 \\
 & TikTok & 156 & 3.88 \\
 & IMDb & 91 & 2.27 \\
\midrule
Source language & English & 2{,}519 & 62.72 \\
 & Chinese & 1{,}497 & 37.28 \\
\midrule
Culture/region & North America & 1{,}196 & 29.78 \\
 & Global/Mixed & 1{,}028 & 25.60 \\
 & Europe & 924 & 23.01 \\
 & East Asia & 791 & 19.70 \\
 & Others & 77 & 1.92 \\
\midrule
Genre & Mixed & 885 & 22.04 \\
 & Action/Adventure & 344 & 8.57 \\
 & Role-Playing & 279 & 6.95 \\
 & Portrait/Figure & 271 & 6.75 \\
 & Drama & 236 & 5.88 \\
 & Variety/Circus/Magic & 218 & 5.43 \\
 & Landscape/Cityscape & 200 & 4.98 \\
 & Comedy & 187 & 4.66 \\
 & Musical & 187 & 4.66 \\
 & Puzzle/Platformer & 185 & 4.61 \\
 & Others & 1{,}024 & 25.50 \\
\bottomrule
\end{tabular}
\end{table}

\subsection{Benchmark Vocabulary Overview}
\label{sec:appendix:wordcloud}

\begin{figure}[!htbp]
    \centering
    \includegraphics[width=0.90\textwidth,pagebox=cropbox]{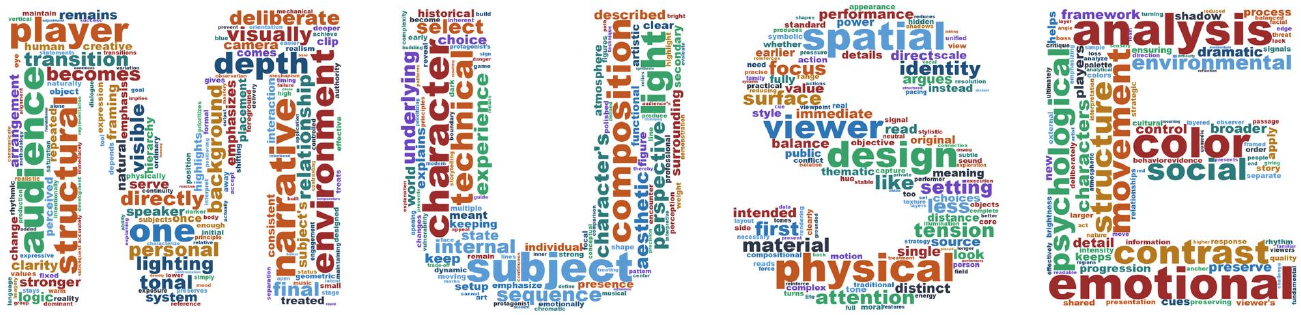}
    \caption{Vocabulary of \benchname. Word size reflects token frequency in questions, options, and core intents after removing function words and generic analytical fillers.}
    \label{fig:wordcloud}
\end{figure}

We visualize the dominant terms of \benchname in \Cref{fig:wordcloud} after removing function words and generic analytical fillers, exposing the artistic vocabulary that drives stems, options, and core intents.

\section{Construction Details}
\label{sec:appendix:construction_protocol}
This appendix expands the construction pipeline summarized in \Cref{sec:collection_annotation} and the human-in-the-loop quality review described in \Cref{sec:hitl_loop}. The subsections that follow track the per-video phases of the main text in order: source curation feeds the corpus; transcription and category inference attach the metadata each video needs; \emph{Phase A} segments the video into 10\,s clips; \emph{Phase B} captions every clip; \emph{Phase C} first selects evidence ranges and then generates 3 to 5 candidate questions per video; \emph{Phase D} synthesizes distractors. The two final subsections document the iterative review loop that mutates the in-context prompt across rounds and the final-stage human evaluation that validates the released benchmark.

\subsection{Source Curation}
\label{sec:appendix:source_curation}

This subsection details the source-discovery stage. We aim to collect a corpus of long-form video essays where a domain expert verbally analyzes an artistic artifact while relevant visual content appears on screen. Source discovery proceeds in four LLM-controlled stages: (1) A category-aware keyword generator produces an initial keyword list grounded in the category's controlled vocabulary. (2) These keywords query public web video searches to collect candidate \texttt{(video\_id, title, channel, description, view\_count)} records. (3) A relevance judge inspects each candidate's metadata, emitting a binary verdict and a confidence score; only candidates with \texttt{is\_relevant=true} and confidence $\ge 0.55$ are admitted. (4) When an active keyword exhausts its top results without yielding new admitted videos, a variant generator extends the keyword list, continuing until the per-category quota is met. Finally, a human-vetting pass removes residual failures undetectable by the metadata filter. The four LLM stages are each governed by an explicit system and user prompt, reproduced in \Cref{fig:source_keyword_prompt,fig:source_relevance_prompt,fig:source_variant_prompt,fig:source_human_vet}. A summary of the deterministic fallback keyword lists, with per-block sizes and 3--5 representative entries per block, is reported in \Cref{tab:source_seeds_summary}.

\paragraph{Keyword generation prompt.}
\Cref{fig:source_keyword_prompt} reproduces the system and user prompt used at stage~1, instantiated for Cinematic Arts. The same envelope is reused for the other three categories with the focus list and example terms swapped to the matching controlled vocabulary. Output is constrained to a single JSON object \texttt{\{"keywords":[\,...\,]\}} and the decoding temperature is fixed at $0.2$.

\input{figures/source_keyword_prompt}

\paragraph{Relevance judgment prompt.}
For every candidate returned by the search step, the relevance judge sees only public metadata plus the active keyword list and emits a JSON verdict. Of the four categories, Stage Performing is the broadest, since it admits stand-up specials, sketch comedy, dance theater, and traditional opera forms in addition to musical theater and opera. Its prompt, reproduced in \Cref{fig:source_relevance_prompt}, is therefore the most explicit about positive and negative cases. The Cinematic, Static Visual, and Game variants share the same envelope and additionally hard-exclude DIY tutorials, software walkthroughs, and gameplay-only content respectively.

\input{figures/source_relevance_prompt}

\paragraph{Variant expansion prompt.}
When a keyword exhausts its top results without producing new admitted videos, the keyword generator is invoked again under the prompt in \Cref{fig:source_variant_prompt}, conditioned on the most recent ten keywords already issued. The variant generator is asked to stay inside the category's focal vocabulary (the \texttt{variant\_focus} string).

\input{figures/source_variant_prompt}

\paragraph{Human-vetting prompt.}
After the relevance judge admits a candidate, a final pass removes the three residual failure modes that metadata filtering cannot detect. The instruction shown to reviewers (and used as a verbatim system prompt when the same pass is delegated to a stronger LLM as a sanity check) is reproduced in \Cref{fig:source_human_vet}.

\input{figures/source_human_vet_prompt}

\paragraph{Keyword list summary.}
\Cref{tab:source_seeds_summary} reports the per-block size of the deterministic-fallback keyword list together with 3--5 representative entries per block. The fallback covers 273 keywords across the four categories: Cinematic Arts and Static Visual Arts use a flat 10-keyword controlled vocabulary each; Game Arts is split into a core controlled vocabulary, case-anchored studies, studio/director vocabulary, and a general-purpose expansion block; Stage Performing Arts is the largest list, organized into thirteen thematic blocks that span comedy, musical theater, opera, drama, dance, variety, world theater, and general performance analysis. The complete word-for-word lists are released with the benchmark code release.

\input{tables/source_seeds_summary}

\subsection{Transcription}
\label{sec:appendix:transcription}

Every retained video is transcribed with Whisper-Large-v3~\citep{radford2023robust} into a JSON record that anchors the rest of the pipeline. \Cref{fig:transcript_json_example} shows the schema. The full free-text transcript is consumed by the QA generation prompt of Phase~C; the timestamped \texttt{segments} entries are routed to the clip-level captioning prompt of Phase~B so that each $10$\,s window can see the narration that overlaps it; the \texttt{metadata} block lets the segmenter enforce the duration, resolution, and channel-count limits documented in \Cref{tab:dataset_stats}.

\input{figures/transcript_json_example}

\subsection{Phase A: Clip Segmentation}
\label{sec:appendix:phase_a}

Following~\citep{wang2025videoitg}, each retained video is partitioned into non-overlapping ten-second clips that establish a uniform temporal granularity for downstream captioning, question generation, and evaluation. The maximum video duration is capped at $1{,}800$\,s ($30$ minutes) so that a single source contributes a bounded number of clips and a bounded number of QA pairs. Each clip carries a stable index that is reused as a stable handle by the captioning and clip-matching prompts and by the per-question \texttt{relevant\_clips} field.

\subsection{Phase B: Clip Captioning}
\label{sec:appendix:phase_b}

Each ten-second clip is captioned by Keye-VL-1.5~\citep{yang2025kwai} sampling at one frame per second, conditioned on the temporally aligned narrator transcript. The caption covers visual attributes such as color, composition, motion, and scene context. Captions serve only as a construction artifact for downstream question generation and review; they are never exposed to evaluated models. The prompt envelope is reproduced in \Cref{fig:clip_description_prompt}; subsequent clips additionally see the running chronological narrative of previous clips so that descriptions remain locally coherent. The system prompt is concatenated at runtime with one of four category-specific guidance blocks (Cinematic, Static Visual, Stage Performing, or Game) that biases the caption toward category-relevant evidence.

\input{figures/clip_description_prompt}

\subsection{Phase C: Evidence Selection and Question Generation}
\label{sec:appendix:phase_c}

Given the chronological clip captions and the full narrator transcript, the generator first identifies a single contiguous \texttt{relevant\_clips} range for each candidate item. It then generates the question and correct answer conditioned on the selected evidence, producing 3 to 5 candidate pairs per video. The target proportion of \texttt{multi\_select} questions is 44.67\% (2 to 4 independent correct answer points) and the rest are \texttt{single\_select} (one correct answer). The prompt envelope is reproduced in \Cref{fig:qa_generation_prompt}. As with captioning, the system prompt is augmented with one of four category-specific question-design guidance blocks at runtime. The evidence-selection step follows the contiguous-range constraint documented in \Cref{fig:quality_review_matrix} (failure 4).

\input{figures/qa_generation_prompt}

\subsection{Phase D: Distractor Generation}
\label{sec:appendix:phase_d}

The distractor generator completes each question's answer set with plausible alternatives drawn from seven strategies: \emph{technical misread}, \emph{over-simplification}, \emph{concept swap}, \emph{cause-effect inversion}, \emph{scope error}, \emph{temporal confusion}, and \emph{partial truth}. Single-select questions receive 3 to 7 distractors; multi-select questions combine 2 to 4 correct answers with distractors, with 4 to 8 options in total. Each distractor uses a different strategy. The prompt is reproduced in \Cref{fig:distractor_generation_prompt}. The category-specific \texttt{scope\_description} field shares the controlled vocabulary used by the source-curation prompts in \Cref{sec:appendix:source_curation}.

\input{figures/distractor_generation_prompt}

\subsection{Quality Review Loop}
\label{sec:appendix:prompt_evolution}

This subsection documents the iterative review loop introduced in \Cref{sec:hitl_loop}. The loop is applied independently to each art category. In every round, a pilot batch of QA pairs is generated under the current in-context prompt, domain-expert reviewers tag each sampled item under the four failure dimensions of \Cref{sec:hitl_loop}, the new failure types of the round are consolidated, and the prompt is rewritten with additional rules. Rules take two forms. \emph{Hard red lines} are short-tagged content constraints written into the QA-Generation prompt that gate the stem and options at generation time. \emph{Category filters} are short-tagged content constraints written into the source-curation human-vetting pass that gate the source clips before they enter Phase~A.

\Cref{fig:quality_review_matrix} walks through one representative bad case per main-text failure dimension and shows the prompt-level rule added in response together with the regenerated form of the same item. The four rows of the figure correspond exactly to the four failure dimensions named in \Cref{sec:hitl_loop} (narrator-dependent answerability; ambiguous stems; weak or factually incorrect distractors; misaligned clip references), so a reader who has just read \Cref{sec:hitl_loop} can scan top to bottom and see the loop in action for each named dimension.

\input{figures/quality_review_matrix}

\paragraph{Game Arts trajectory across three rounds.}
A category-specific instance of the loop is reproduced in \Cref{fig:prompt_evolution_game}, which traces a single Game Arts item through the three prompt revisions that retired the round-by-round failure types. Round~1 used a generic instruction that asked the model to ``analyze the design intent'' of an arbitrary clip; pilot review surfaced two recurring failures, namely stems written in an abstract academic register that no longer commit to a specific design choice, and pairs in which the generation pipeline failed to settle on a unique key. Round~2 tightened the prompt with two hard requirements: every question must name a specific game in the stem, and every stem must point to a concrete on-screen design choice. This eliminated the abstract-register failure but introduced a second-order shortcut, since the named title is sufficient for a model with strong text priors to retrieve the answer from training-set memory without watching the clip. Round~3 added two anti-shortcut rules in response: \emph{R1} forbids proper nouns (game titles, character names, locations) anywhere in the stem, options, or correct answer; \emph{R2} requires each stem to identify a visible or audible element that locates the question in the clip, while leaving the relevant spatial, temporal, or expressive relationships for the solver to infer. After Round~3 no new failure type was raised in three consecutive rounds, and Game Arts exited the refinement loop.

\begin{figure}[!htbp]
\centering
\begin{minipage}[t]{0.32\textwidth}
\begin{tcolorbox}[colback=red!4!white, colframe=red!50!black, title=\textbf{Round 1: initial generation}, fontupper=\scriptsize, boxrule=0.4pt]
\textbf{Q.} Analyze the interplay between the environmental conditions and the sequential logic presented. Which two conclusions accurately characterize the intentionality of the design and its formal execution?
\medskip

\textbf{A.} The environmental brightness functions as a psychological cue to suggest a sense of optimism and emotional reprieve for the participant during the final stages of the journey.

\textbf{B.} The implementation of a daylight setting serves as a strategic mechanism to heighten the visibility of hazards.

\textbf{C.} The transition suffers from a structural flaw where the absence of specific visual markers, such as the transport vehicle, breaks narrative continuity.

\textbf{D.} The exclusion of the C-130 transport plane at the start of the sequence is a calculated design choice to emphasize the survivors' isolation.
\medskip

\textbf{Key:} \texttt{""} (unresolved)
\medskip

\rule{\linewidth}{0.2pt}\par
\textbf{Failure modes addressed in Round 2.} abstract academic register; no concrete design choice; generation pipeline failed to commit a unique correct option.
\end{tcolorbox}
\end{minipage}%
\hfill
\begin{minipage}[t]{0.32\textwidth}
\begin{tcolorbox}[colback=yellow!8!white, colframe=orange!70!black, title=\textbf{Round 2: named-game decision focus}, fontupper=\scriptsize, boxrule=0.4pt]
\textbf{Q.} In the ``Blood Harvest'' campaign of \emph{Left 4 Dead}, the storefronts on the building facades carry the logos of the Richardson Atlantic Company and its subsidiaries. What player experience does this design choice most directly aim to produce?
\medskip

\textbf{A.} \textbf{(correct)} A sense that the world is a real place where people lived and worked, rather than a generic zombie shooting gallery.

\textbf{B.} A way to help the player quickly locate the destructible weak points of each building.

\textbf{C.} A cue that collecting company logos is required to unlock map rewards.

\textbf{D.} A direct narrative critique of corporate headquarters as the central theme of the map.
\medskip

\textbf{Key:} A; \texttt{game\_referenced}: Left 4 Dead
\medskip

\rule{\linewidth}{0.2pt}\par
\textbf{Failure mode addressed in Round 3.} the named title plus the named campaign in the stem are sufficient priors for a strong language model to retrieve the answer from training-set memory without watching the clip.
\end{tcolorbox}
\end{minipage}%
\hfill
\begin{minipage}[t]{0.32\textwidth}
\begin{tcolorbox}[colback=green!5!white, colframe=green!45!black, title=\textbf{Round 3: visible-anchor, no proper nouns}, fontupper=\scriptsize, boxrule=0.4pt]
\textbf{Q.} How does the arrangement of interior spaces and the player's route through them shape the experience of this level?
\medskip

\textbf{A.} \textbf{(correct)} A prolonged sense of confinement and unease by delaying escape and increasing exposure to threats.

\textbf{B.} A methodical resource-scarcity loop, where the cramped rooms force players to ration ammo by limiting engagement distances.

\textbf{C.} A tactical chokepoint system that rewards reading enemy patrol paths and timing movement through each corridor.

\textbf{D.} A spatial memory challenge where vertical layering forces the player to mentally reconstruct the building layout to navigate efficiently.
\medskip

\textbf{Key:} A; \texttt{visual\_anchor}: ``narrow corridors, locked-in spaces, vertical multi-level buildings''. The internal title field is withheld from the evaluated model.
\medskip

\rule{\linewidth}{0.2pt}\par
\textbf{Stable form.} the stem identifies the interior layout and the player's route as the focus of analysis without describing their spatial constraints or stating the intended effect.
\end{tcolorbox}
\end{minipage}
\caption{Game Arts question evolution across three prompt revisions. Round~1 produces an abstract academic stem with no committed key; Round~2 enforces a named-game decision focus but introduces a title-recognition shortcut; Round~3 removes proper nouns and anchors the question to the interior layout and player movement, leaving their relationship and expressive effect to be inferred from the clip. Each box lists the failure mode that motivated the next revision.}
\label{fig:prompt_evolution_game}
\end{figure}

\subsection{Failure Taxonomy}
\label{sec:appendix:failure_taxonomy}

We additionally conducted a round-by-round human audit of an early benchmark draft of $5{,}000$ QA pairs. In each round, domain experts inspected the full corpus and, at the end of the round, consolidated the newly observed failures into a tag set that fed the next round of prompt revision. Across rounds, the audit surfaced eight failure tags in four severity tiers, each accompanied by a mitigation rule. Table~\ref{tab:failure_taxonomy} records these construction-stage counts and interventions. Flagged CRITICAL and HIGH cases were replaced from the QA pool; MEDIUM and LOW cases were addressed through prompt refinement and programmatic alignment. The final-set audit in Appendix~\ref{sec:appendix:option_audit} separately measures residual wording and equivalent-option issues.

\input{tables/failure_taxonomy}

Beyond the eight tagged failures in \Cref{tab:failure_taxonomy}, the audit identified seven systemic issues requiring prompt-level refinement rather than item-level replacement: low option discriminability, overly academic or verbose language, stems failing to target artistic intent, imprecise distractors without distinct error strategies, misclassified text-vision alignment, inconsistent enforcement of the visual-evidence requirement, and embedded option lists in stems. The corresponding prompt-level fixes are reflected in the QA-Generation and Distractor prompts of \Cref{sec:appendix:phase_c,sec:appendix:phase_d}, and in the four reviewer-facing dimensions of \Cref{fig:quality_review_matrix}.

\subsection{Human Evaluation Details}
\label{sec:appendix:human_eval_details}

\begin{figure}[!htbp]
\centering
\includegraphics[width=0.6\textwidth]{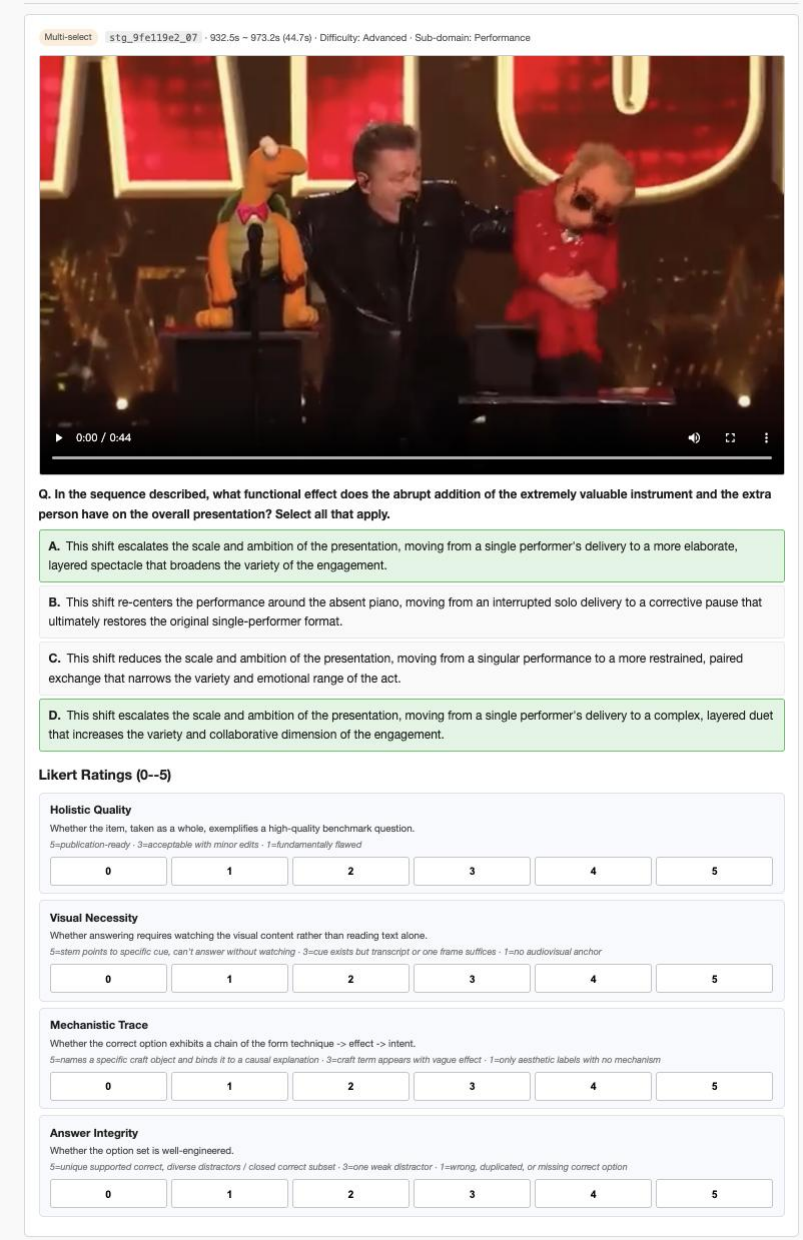}
\caption{Human evaluation interface used by domain experts. The top panel shows the corresponding clip, the question, and the answer options with correct option(s) highlighted in green; the bottom panel shows the 0--5 Likert sliders for the four quality dimensions (Holistic Quality, Visual Necessity, Mechanistic Trace, Answer Integrity), each accompanied by its anchor descriptions.}
\label{fig:eval_ui}
\end{figure}

To support the human evaluation reported in \Cref{sec:hitl_loop}, we built a self-contained web interface (\Cref{fig:eval_ui}) that lets each domain expert play the corresponding clip in place, read the question and options, and rate the item on the four Likert dimensions defined below. 

\paragraph{Rater Pool.}
Four domain experts participated as unpaid collaborators and were assigned to categories matching their formal training. Two raters trained in Drama, Film, and Literature evaluated \textit{Stage Performing Arts} and \textit{Cinematic Arts}; two trained in Computational Media and Arts evaluated \textit{Game Arts} and \textit{Static Visual Arts}. Each category was rated independently by two experts, yielding two scores per item that were averaged to produce the per-category means in \Cref{fig:human_eval_likert}.

\paragraph{Expert performance baseline.}
The expert performance baseline was collected separately from the four-person quality review above. Two domain experts, each with a master's degree in a relevant field, were assigned the 4{,}016-question benchmark through disjoint category assignments: one covered Cinematic Arts and Stage Performing Arts, and the other covered Static Visual Arts and Game Arts. The baseline uses the valid scored responses from these assignments. Applying the same scoring rules used for models to these responses gave 87.18\% Overall ACC, 90.98\% Single CAA, and 78.00\% Multi EM.

\paragraph{Non-expert comparison.}
In an auxiliary study, we randomly sampled 500 questions and asked two participants without formal training in the covered art domains to answer them. The non-expert participants, domain experts, and Claude-4.6-Opus were evaluated on the same subset with the same scoring protocol. \Cref{tab:nonexpert_comparison} reports the individual non-expert scores rather than pooling two participants into a single estimate.

\begin{table}[htbp]
\centering
\caption{Human and model results on the same random 500-question subset (\%).}
\label{tab:nonexpert_comparison}
\footnotesize
\setlength{\tabcolsep}{4pt}
\begin{tabular}{@{}lrrrr@{}}
\toprule
\textbf{Evaluator} & \textbf{Overall ACC} & \textbf{Single CAA} & \textbf{Multi EM} & \textbf{Multi F1} \\
\midrule
Non-expert participant 1 & 54.60 & 65.57 & 36.44 & 74.16 \\
Non-expert participant 2 & 42.80 & 51.74 & 23.89 & 68.32 \\
Domain experts & 85.40 & 90.35 & 78.14 & 93.70 \\
Claude-4.6-Opus & 49.20 & 55.44 & 33.60 & 70.45 \\
\bottomrule
\end{tabular}
\end{table}

\paragraph{Scoring Rubric.}
For every item, raters score the four dimensions on a 0--5 Likert scale, with the one-line definition and anchor descriptions shown beside each slider.
\begin{itemize}[leftmargin=*, parsep=2pt, topsep=2pt]
    \item \textbf{Holistic Quality} -- whether the item, taken as a whole, exemplifies a high-quality benchmark question. Anchors: 5 = publication-ready; 3 = acceptable with minor edits; 1 = fundamentally flawed.
    \item \textbf{Visual Necessity} -- whether answering requires watching the visual content rather than reading text alone. Anchors: 5 = stem points to a specific on-screen cue and cannot be answered without watching; 3 = cue exists but the transcript or a single frame suffices; 1 = no audiovisual anchor.
    \item \textbf{Mechanistic Trace} -- whether the correct option exhibits a chain of the form \emph{technique $\rightarrow$ effect $\rightarrow$ intent}. Anchors: 5 = names a specific craft object and binds it to a causal explanation; 3 = craft term appears with a vague effect; 1 = only aesthetic labels with no mechanism.
    \item \textbf{Answer Integrity} -- whether the option set is well-engineered. Anchors: 5 = unique supported correct option with diverse distractors (or, for multi-select, a closed correct subset); 3 = one weak distractor; 1 = wrong, duplicated, or missing correct option.
\end{itemize}
Raters receive a brief calibration walkthrough on three pre-selected pairs per category before scoring begins, and are instructed to reserve the extreme anchors (0 and 5) for unambiguous cases.

\input{sections/revision_answer_agreement}
\input{sections/revision_text_audit}
\input{sections/revision_option_audit}

\section{Detailed Evaluation Metrics}
\label{sec:appendix:eval_metrics_details}

\begingroup
\paragraph{Human expert scores.}
\label{sec:appendix:human_caa}
Let $h_i$ be the assigned expert's answer, $y_i$ the reference answer, and $\mathcal{S}$ the set of valid scored single-select questions. The two experts' disjoint category assignments supply one response per included question, so the human correctness indicator and percentage-scaled CAA are
\begin{equation}
a_i^{\mathrm{human}}=\mathbf{1}[h_i=y_i],\qquad
\mathrm{CAA}_{\mathrm{human}}=\frac{100}{|\mathcal{S}|}\sum_{i\in\mathcal{S}}\frac{a_i^{\mathrm{human}}-1/K_i}{1-1/K_i}.
\end{equation}
A correct response contributes 1 before percentage scaling; an incorrect response contributes $-1/(K_i-1)$. Category scores use the corresponding valid single-select subset. Overall ACC averages fully correct responses over all valid scored items, and Multi EM averages exact-set correctness over valid multi-select items. This procedure gives 90.98\% Single CAA, 87.18\% Overall ACC, and 78.00\% Multi EM for the expert baseline. The overlapping study in Appendix~\ref{sec:appendix:answer_agreement} separately measures answer agreement.
\endgroup

\subsection{Single-Select Evaluation: Chance-Adjusted Accuracy}
\label{sec:appendix:single_select_metric}

For single-select questions, the number of answer options varies across instances. As a result, raw accuracy is not directly comparable across questions with different random-guess baselines. To account for this, we report \emph{chance-adjusted accuracy} (CAA), which measures performance relative to random guessing.

For the $i$-th single-select question, let $K_i$ be the number of answer options and let $a_i=\mathbf{1}[\hat{y}_i=y_i]$ denote whether the model prediction $\hat{y}_i$ matches the ground-truth answer $y_i$. We define
\begin{equation}
c_i=\frac{1}{K_i},
\qquad
\mathrm{CAA}_i
=
\frac{a_i-c_i}{1-c_i}
=
\frac{\mathbf{1}[\hat{y}_i=y_i]-\frac{1}{K_i}}
{1-\frac{1}{K_i}}.
\end{equation}
Here, $c_i$ is the random-guess accuracy for question $i$. This normalization gives $\mathrm{CAA}_i=1$ for a correct prediction, has expected score $0$ under uniform random guessing, and yields negative scores for worse-than-chance performance.

Given $N_{\mathrm{single}}$ valid single-select questions included in the evaluation, the reported score is
\begin{equation}
\mathrm{CAA}
=
\frac{1}{N_{\mathrm{single}}}
\sum_{i=1}^{N_{\mathrm{single}}}
\mathrm{CAA}_i.
\end{equation}
In practice, this metric makes results more comparable across questions with different option counts, since achieving the same raw accuracy on a question with more options reflects stronger performance.

\subsection{Multi-Select Evaluation: Exact Match, Precision, Recall, and F1}
\label{sec:appendix:multi_select_metric}

The primary multi-select metric is exact-match (EM) accuracy. Here, $N_{\mathrm{multi}}$ counts valid multi-select questions included in the evaluation; category-level EM uses the corresponding category subset. For ground-truth option set $Y_j$ and predicted set $\hat{Y}_j$, it is
\begin{equation}
\mathrm{EM}
= \frac{1}{N_{\mathrm{multi}}}
\sum_{j=1}^{N_{\mathrm{multi}}}
\mathbf{1}[\hat{Y}_j=Y_j].
\label{eq:em}
\end{equation}
Because EM requires the complete correct set, we also report precision, recall, and F1 on the predicted answer sets using macro averages to diagnose partially correct predictions.

For each multi-select question $q_j$, let $Y_j \subseteq \mathcal{O}_j$ denote the set of ground-truth correct options and let $\hat{Y}_j \subseteq \mathcal{O}_j$ denote the model-predicted set. We first define
\begin{equation}
\mathrm{TP}_j = |\hat{Y}_j \cap Y_j|,
\qquad
\mathrm{FP}_j = |\hat{Y}_j \setminus Y_j|,
\qquad
\mathrm{FN}_j = |Y_j \setminus \hat{Y}_j|.
\end{equation}
The per-question precision, recall, and F1 are then
\begin{equation}
P_j =
\frac{\mathrm{TP}_j}{\mathrm{TP}_j+\mathrm{FP}_j},
\qquad
R_j =
\frac{\mathrm{TP}_j}{\mathrm{TP}_j+\mathrm{FN}_j},
\qquad
F1_j =
\frac{2P_jR_j}{P_j+R_j}.
\label{eq:f1}
\end{equation}

For macro averaging, we average the per-question scores:
\begin{equation}
M_{\mathrm{macro}}
=
\frac{1}{N_{\mathrm{multi}}}
\sum_{j=1}^{N_{\mathrm{multi}}}
M_j,
\qquad
M \in \{P,R,F1\}.
\end{equation}

For completeness, micro averaging instead aggregates counts across all multi-select questions,
\begin{equation}
\mathrm{TP} =
\sum_{j=1}^{N_{\mathrm{multi}}}
\mathrm{TP}_j,
\qquad
\mathrm{FP} =
\sum_{j=1}^{N_{\mathrm{multi}}}
\mathrm{FP}_j,
\qquad
\mathrm{FN} =
\sum_{j=1}^{N_{\mathrm{multi}}}
\mathrm{FN}_j,
\end{equation}
and then compute
\begin{equation}
P_{\mathrm{micro}} =
\frac{\mathrm{TP}}{\mathrm{TP}+\mathrm{FP}},
\qquad
R_{\mathrm{micro}} =
\frac{\mathrm{TP}}{\mathrm{TP}+\mathrm{FN}},
\qquad
F1_{\mathrm{micro}} =
\frac{2P_{\mathrm{micro}}R_{\mathrm{micro}}}
{P_{\mathrm{micro}}+R_{\mathrm{micro}}}.
\end{equation}

Compared with exact-match accuracy, these set-based metrics provide a more informative diagnosis of model behavior. In particular, precision captures the tendency to select incorrect distractors, while recall reflects whether the model misses valid analytical perspectives. F1 balances both aspects into a single score.

\subsection{Overall Accuracy}
\label{sec:appendix:overall_accuracy}
Overall ACC is the fraction of fully correct answers among valid scored items across both question formats:
\begin{equation}
\mathrm{ACC}
= \frac{\sum_{i=1}^{N_{\mathrm{single}}} a_i
+ \sum_{j=1}^{N_{\mathrm{multi}}}\mathbf{1}[\hat{Y}_j=Y_j]}
{N_{\mathrm{single}}+N_{\mathrm{multi}}}.
\end{equation}
Each question has equal weight. CAA is reported separately and is not used to compute Overall ACC. All tabulated scores are expressed as percentages.

\section{Complete Experimental Results}
\label{sec:appendix:add_exp}
\input{tables/app_results_details}

\paragraph{Compute resources.}
Local-weight evaluations use an internal cluster equipped with NVIDIA A800 GPUs. Proprietary models are queried through APIs. GPT-5.4, Claude-4.6-Opus, and Gemini-3.1-pro-preview use their official APIs; the OpenRouter and Volcengine Ark routes are detailed in Appendix~\ref{sec:appendix:input_routes}. Evaluations use zero-shot prompts, with sample counts reported alongside the input settings.

\noindent\Cref{tab:app_evaluation_results} reports per-category ACC, CAA, precision, recall, and F1 for the models with these diagnostics. The V+T and V+A+T leaderboards in Tables~\ref{tab:evaluation_results} and~\ref{tab:audio_results} cover all 34 model versions. Native-audio results have the lowest CAA and EM on Game Arts for every model, while the V+T profiles vary by model. Table~\ref{tab:paired_audio_results} gives the category-level input comparisons.

The multi-select diagnostics separate missed answers from excess selections. For example, in Cinematic Arts, GPT-5.4 achieves 77.52\% precision and 71.05\% recall, while Claude-4.6-Opus achieves 79.60\% precision and 70.68\% recall. Video-R1 has higher recall than precision across all four categories, and LongVU achieves 91.65\% recall on Game Arts despite recording only 1.72\% EM. These results show that high partial agreement can coexist with poor recovery of the exact answer set.

\subsection{Language and Option-Set Shortcut Controls}
\label{sec:appendix:shortcut_controls}
We evaluate two input controls for VideoLLaMA2.1-AV and Qwen2.5-Omni-7B: options only, and transcript plus question and options. The first probes answer-option priors; the second tests whether the source narration can support answers without audiovisual input. Decoding settings, item order, option order, and scoring were held fixed across these controls. \Cref{tab:shortcut_controls} reports these conditions as language-only diagnostics. The source-narration input differs from the narrator-removed audio transcripts used in V+T.

\begin{table}[htbp]
\centering
\caption{Language and option-set controls (\%).}
\label{tab:shortcut_controls}
\footnotesize
\setlength{\tabcolsep}{4pt}
\begin{tabular}{@{}llrrr@{}}
\toprule
\textbf{Model} & \textbf{Input} & \textbf{Overall ACC} & \textbf{Single CAA} & \textbf{Multi EM} \\
\midrule
VideoLLaMA2.1-AV & Options only & 11.05 & 0.39 & 0.05 \\
VideoLLaMA2.1-AV & Transcript + question + options & 11.30 & 0.92 & 0.12 \\
Qwen2.5-Omni-7B & Options only & 17.83 & 9.53 & 6.39 \\
Qwen2.5-Omni-7B & Transcript + question + options & 21.47 & 14.51 & 9.73 \\
\bottomrule
\end{tabular}
\end{table}

\input{sections/revision_language_controls}
\input{sections/revision_input_routes}
\input{sections/revision_main_validation}

\section{Broader Impact}
\label{sec:appendix:broader_impact}
\benchname targets a capability that is currently underrepresented in MLLM evaluation, namely intent-level reasoning about audiovisual artistic expression across cinematic, static visual, stage performing, and game arts. By exposing the gap between expert artistic understanding and current models, the benchmark can guide future training corpora and instruction tuning toward richer cultural and stylistic supervision. It can also serve as a standardized probe for arts education tools, accessibility applications such as audio description for visually impaired audiences, and content analysis pipelines used by archivists, educators, and independent critics.

The gold answers represent source-supported interpretations within the provided answer space. The source-language and regional distribution in \Cref{tab:source_distribution} limits which artistic traditions are represented and therefore the scope of the reported model scores.

\clearpage
\section{Additional Examples of \benchname}
\label{sec:appendix:add_examples}

This section showcases additional samples spanning all four categories of \benchname. Each card displays the eight uniformly sampled frames that form the visual prompt, the multiple-choice question, and all answer options, with the correct option(s) highlighted in red. The header bar reports the question type (single- or multi-select), the number of options, and the sub-domain. Across categories, the pairs consistently demand inferential reasoning grounded in the on-screen evidence rather than surface recognition: models must explain \emph{why} a creative choice produces a particular effect, not merely \emph{what} appears in the frame.

\begin{figure}[!htbp]
\centering
\begin{subfigure}[t]{0.48\textwidth}
\centering
\includegraphics[width=\linewidth]{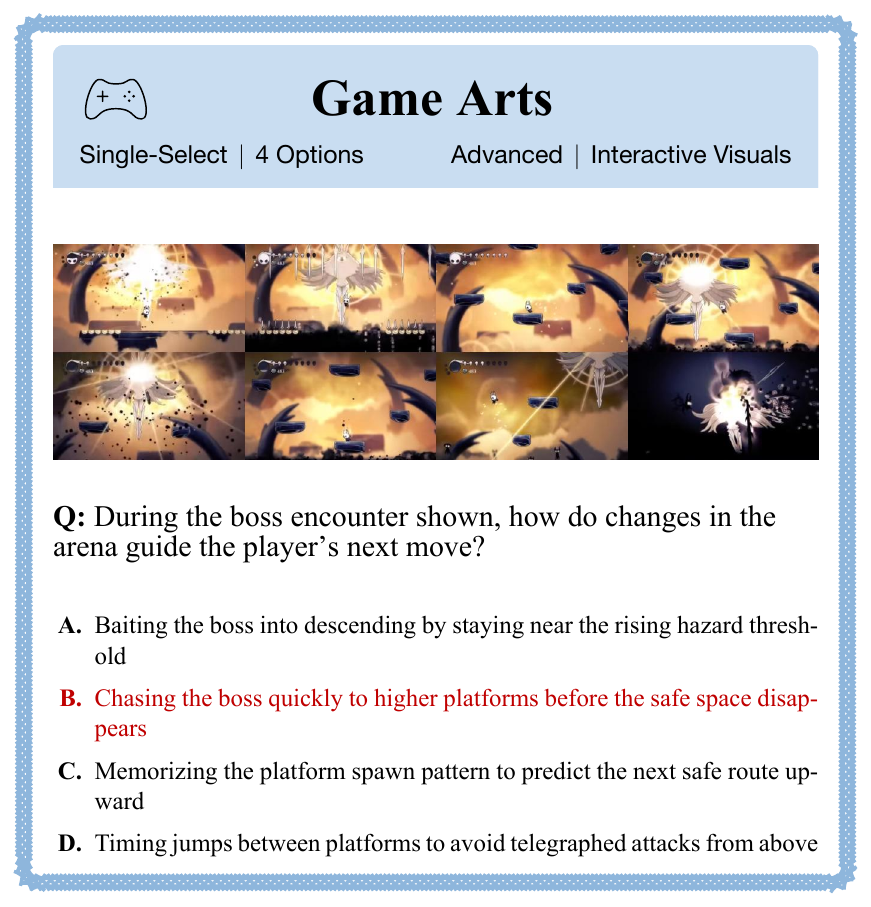}
\caption{\textit{Game Arts} sample 1.}
\label{fig:appendix_example_game_01}
\end{subfigure}
\hfill
\begin{subfigure}[t]{0.48\textwidth}
\centering
\includegraphics[width=\linewidth]{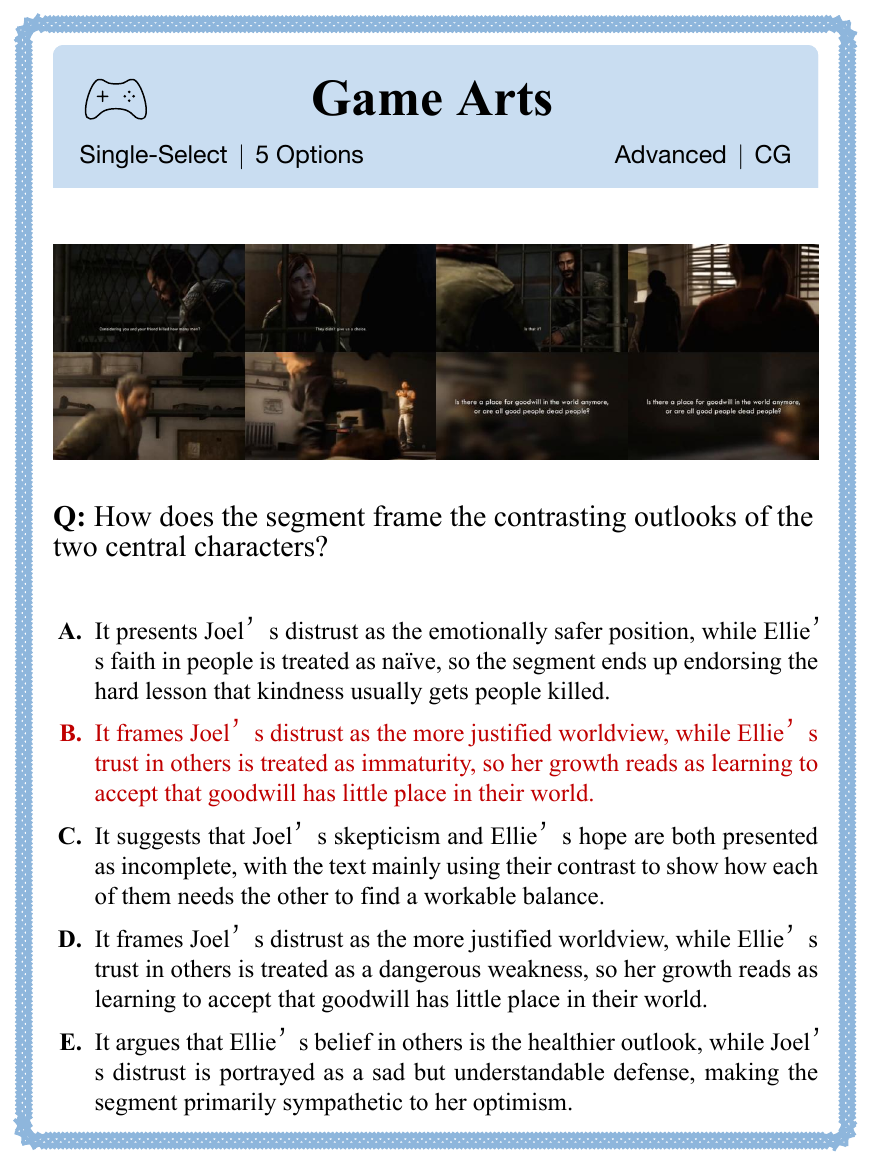}
\caption{\textit{Game Arts} sample 2.}
\label{fig:appendix_example_game_02}
\end{subfigure}

\vspace{0.6em}

\begin{subfigure}[t]{0.48\textwidth}
\centering
\includegraphics[width=\linewidth]{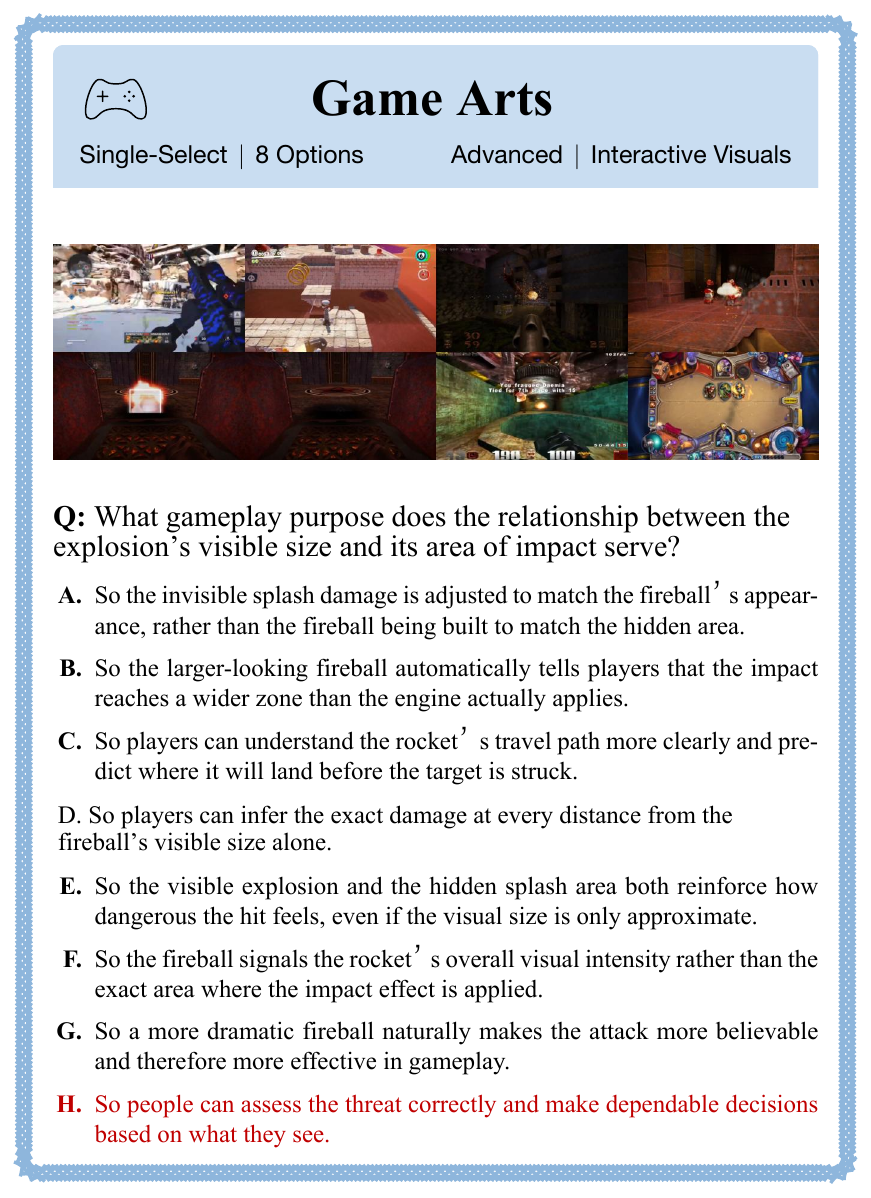}
\caption{\textit{Game Arts} sample 3.}
\label{fig:appendix_example_game_03}
\end{subfigure}
\hfill
\begin{subfigure}[t]{0.48\textwidth}
\centering
\includegraphics[width=\linewidth]{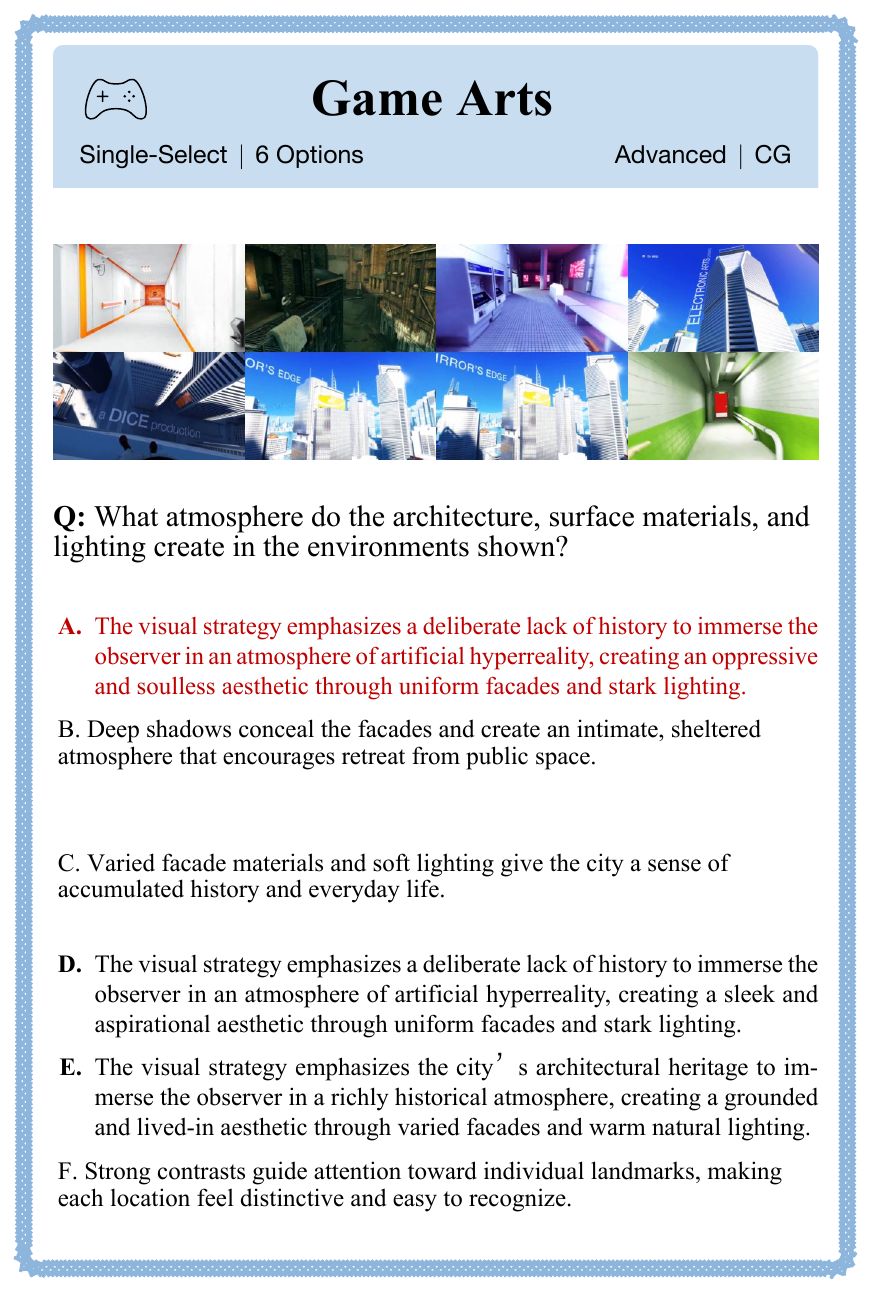}
\caption{\textit{Game Arts} sample 4.}
\label{fig:appendix_example_game_04}
\end{subfigure}
\caption{Additional \textit{Game Arts} samples from \benchname.}
\label{fig:appendix_example_game}
\end{figure}

\paragraph{Game Arts.}
\Cref{fig:appendix_example_game} presents Game Arts pairs that probe analytical reading of boss design, level pacing, narrative cinematography, and audiovisual signaling, requiring models to reason about the design intent behind each visual choice.

\begin{figure}[!htbp]
\centering
\begin{subfigure}[t]{0.48\textwidth}
\centering
\includegraphics[width=\linewidth]{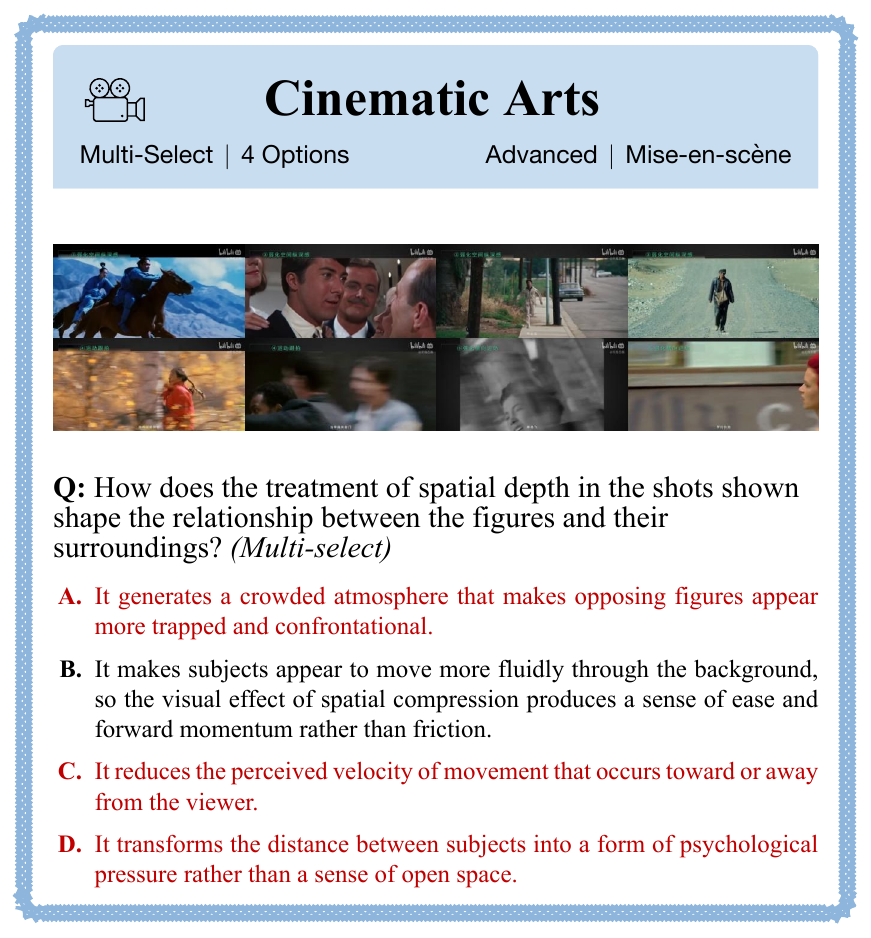}
\caption{\textit{Cinematic Arts} sample 1.}
\label{fig:appendix_example_cin_01}
\end{subfigure}
\hfill
\begin{subfigure}[t]{0.48\textwidth}
\centering
\includegraphics[width=\linewidth]{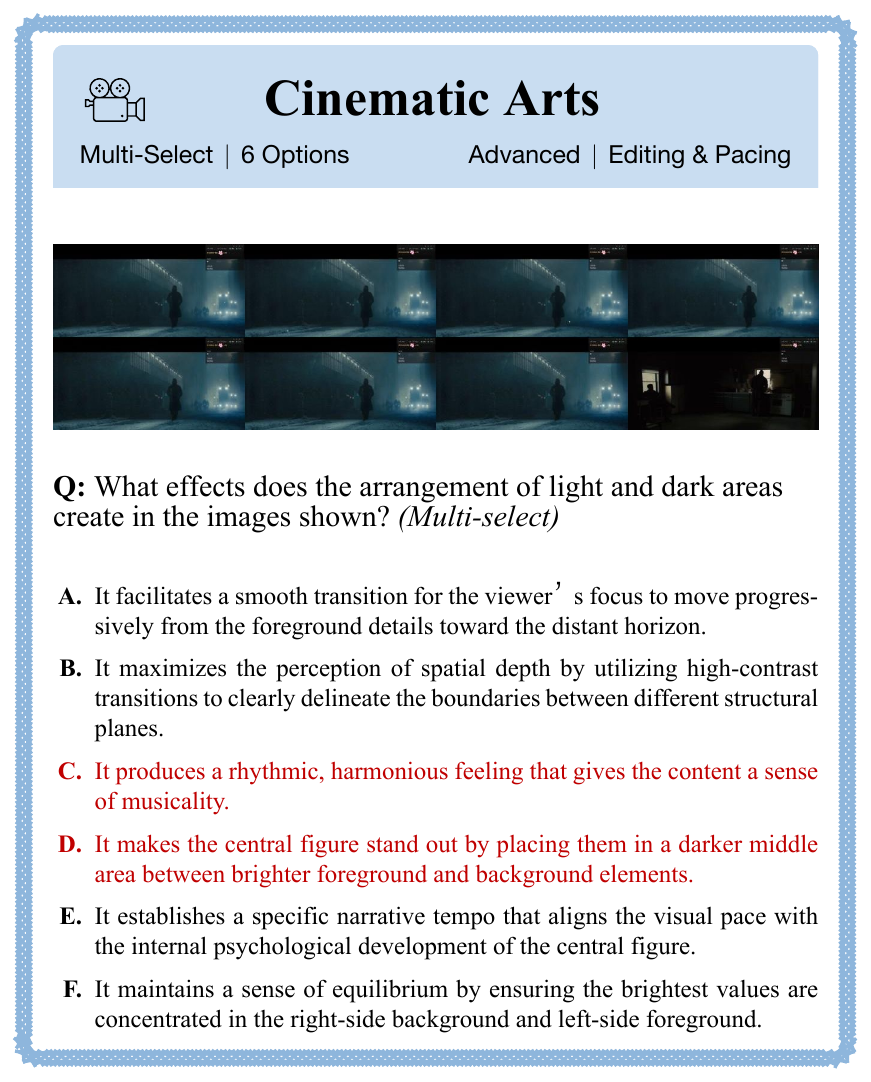}
\caption{\textit{Cinematic Arts} sample 2.}
\label{fig:appendix_example_cin_02}
\end{subfigure}

\vspace{0.6em}

\begin{subfigure}[t]{0.48\textwidth}
\centering
\includegraphics[width=\linewidth]{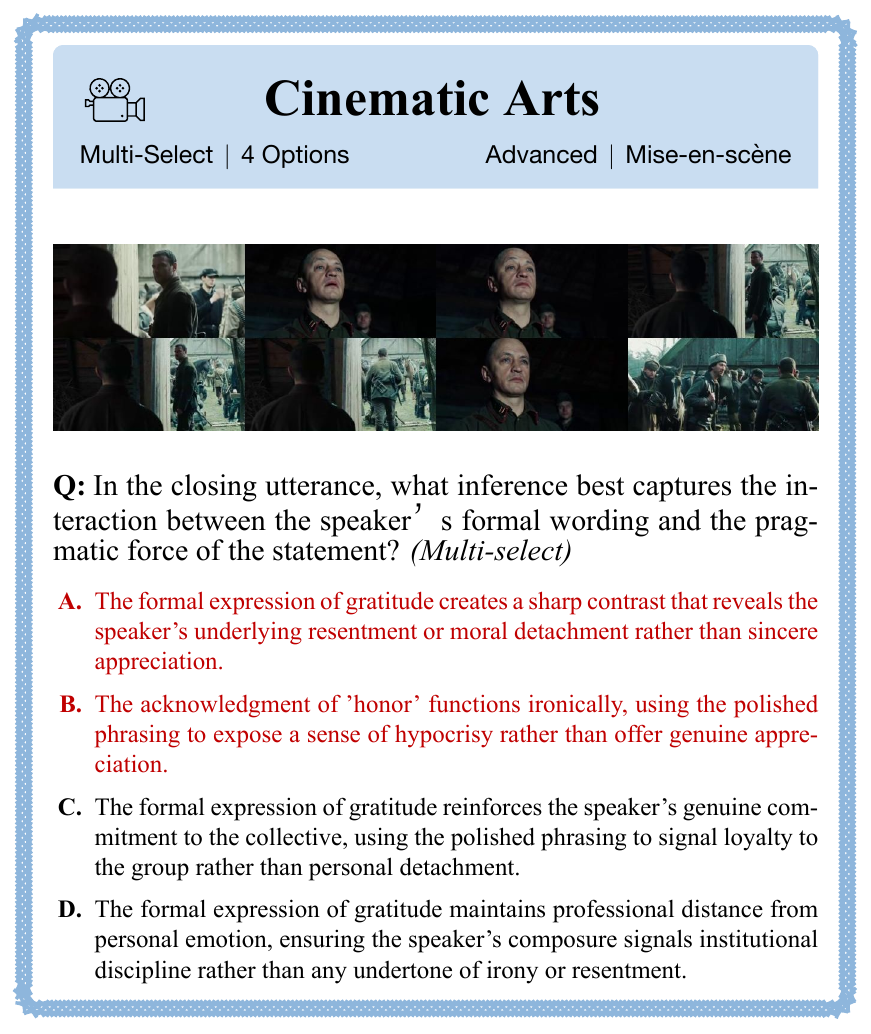}
\caption{\textit{Cinematic Arts} sample 3.}
\label{fig:appendix_example_cin_03}
\end{subfigure}
\hfill
\begin{subfigure}[t]{0.48\textwidth}
\centering
\includegraphics[width=\linewidth]{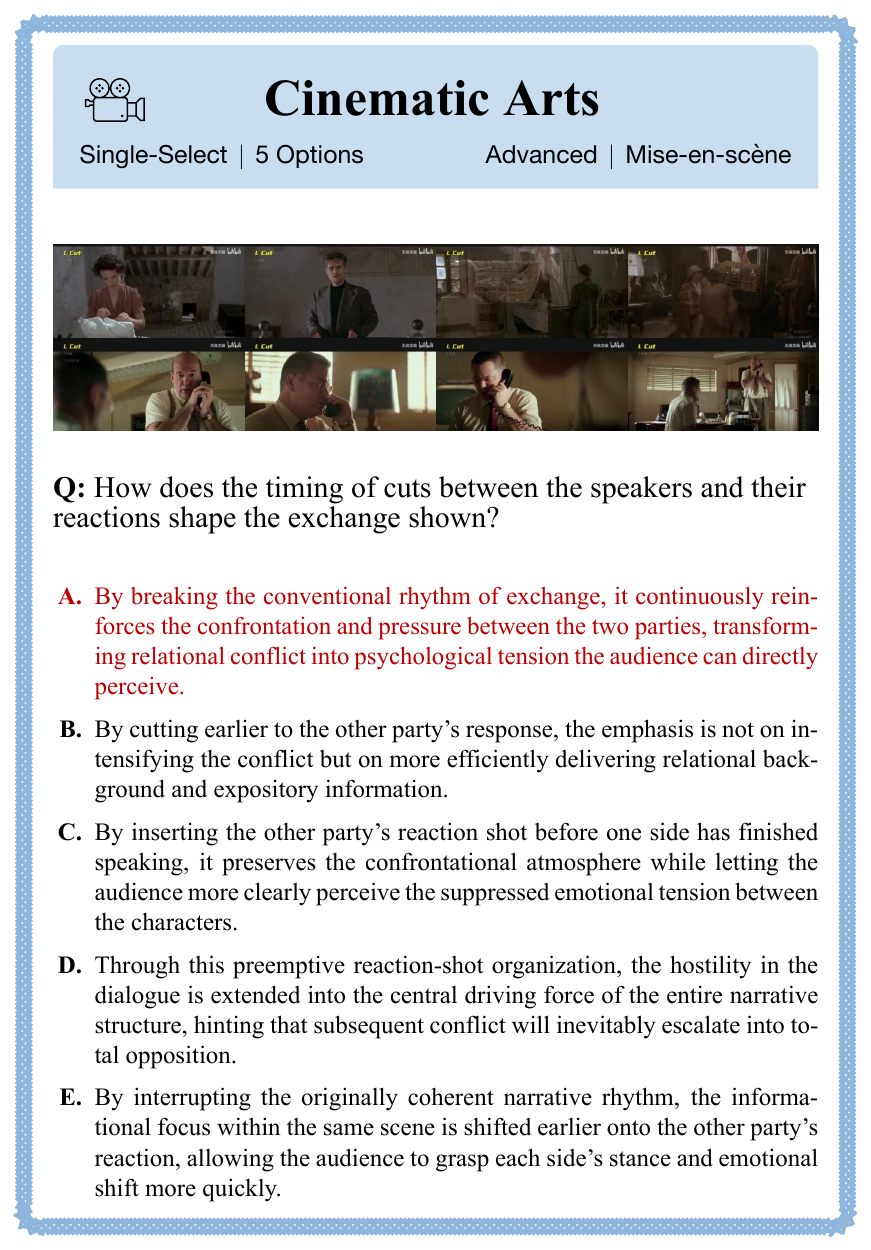}
\caption{\textit{Cinematic Arts} sample 4.}
\label{fig:appendix_example_cin_04}
\end{subfigure}
\caption{Additional \textit{Cinematic Arts} samples from \benchname.}
\label{fig:appendix_example_cin}
\end{figure}

\paragraph{Cinematic Arts.}
\Cref{fig:appendix_example_cin} highlights Cinematic Arts pairs that span shot composition, lighting, editing rhythm, and dialogue pragmatics, emphasizing inferential judgments about how a director's choice yields a particular dramatic or compositional effect.

\begin{figure}[!htbp]
\centering
\begin{subfigure}[t]{0.48\textwidth}
\centering
\includegraphics[width=\linewidth]{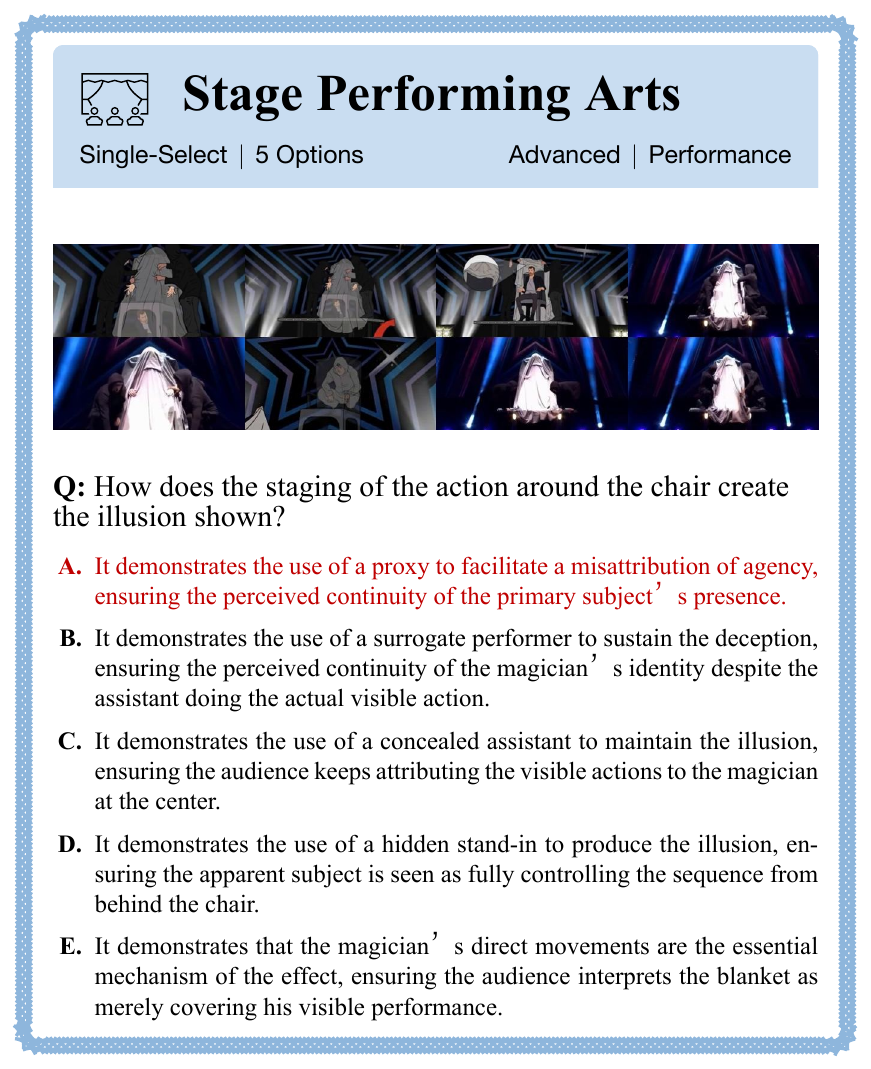}
\caption{\textit{Stage Performing Arts} sample 1.}
\label{fig:appendix_example_stg_01}
\end{subfigure}
\hfill
\begin{subfigure}[t]{0.48\textwidth}
\centering
\includegraphics[width=\linewidth]{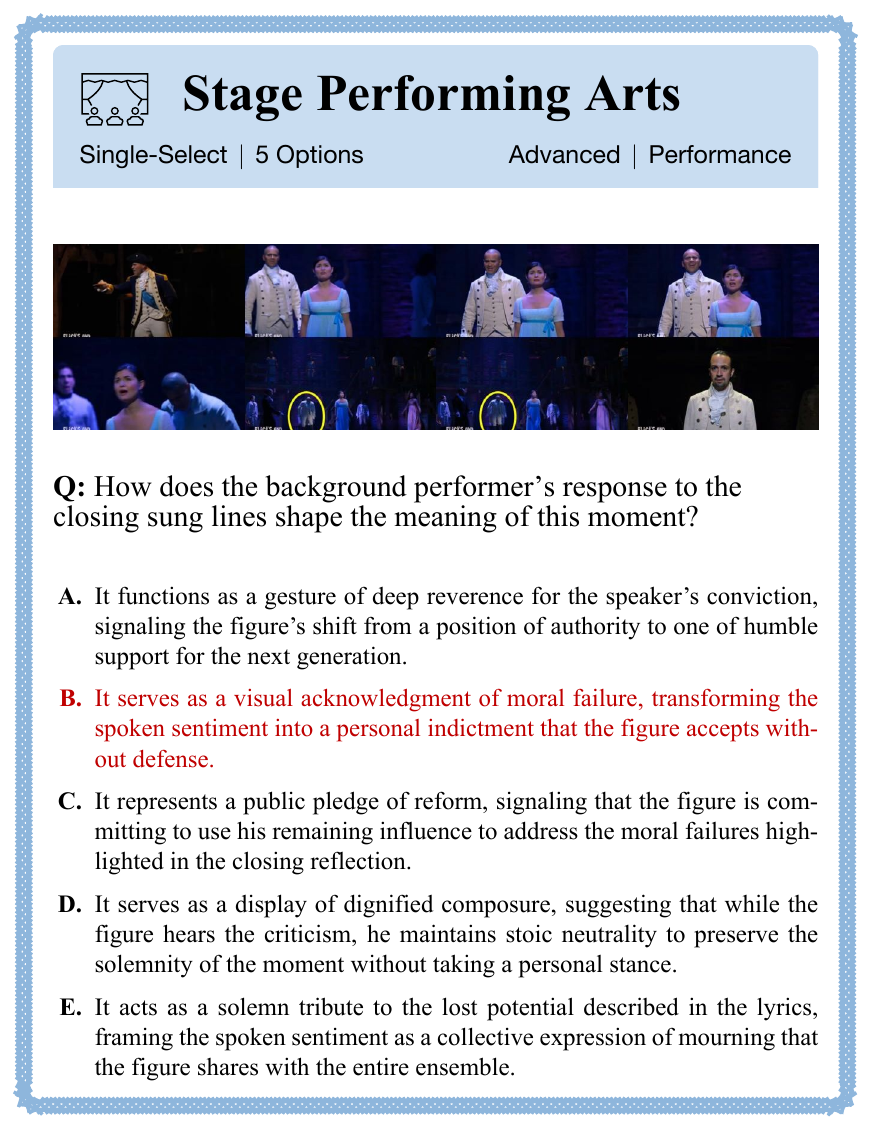}
\caption{\textit{Stage Performing Arts} sample 2.}
\label{fig:appendix_example_stg_02}
\end{subfigure}

\vspace{0.6em}

\begin{subfigure}[t]{0.48\textwidth}
\centering
\includegraphics[width=\linewidth]{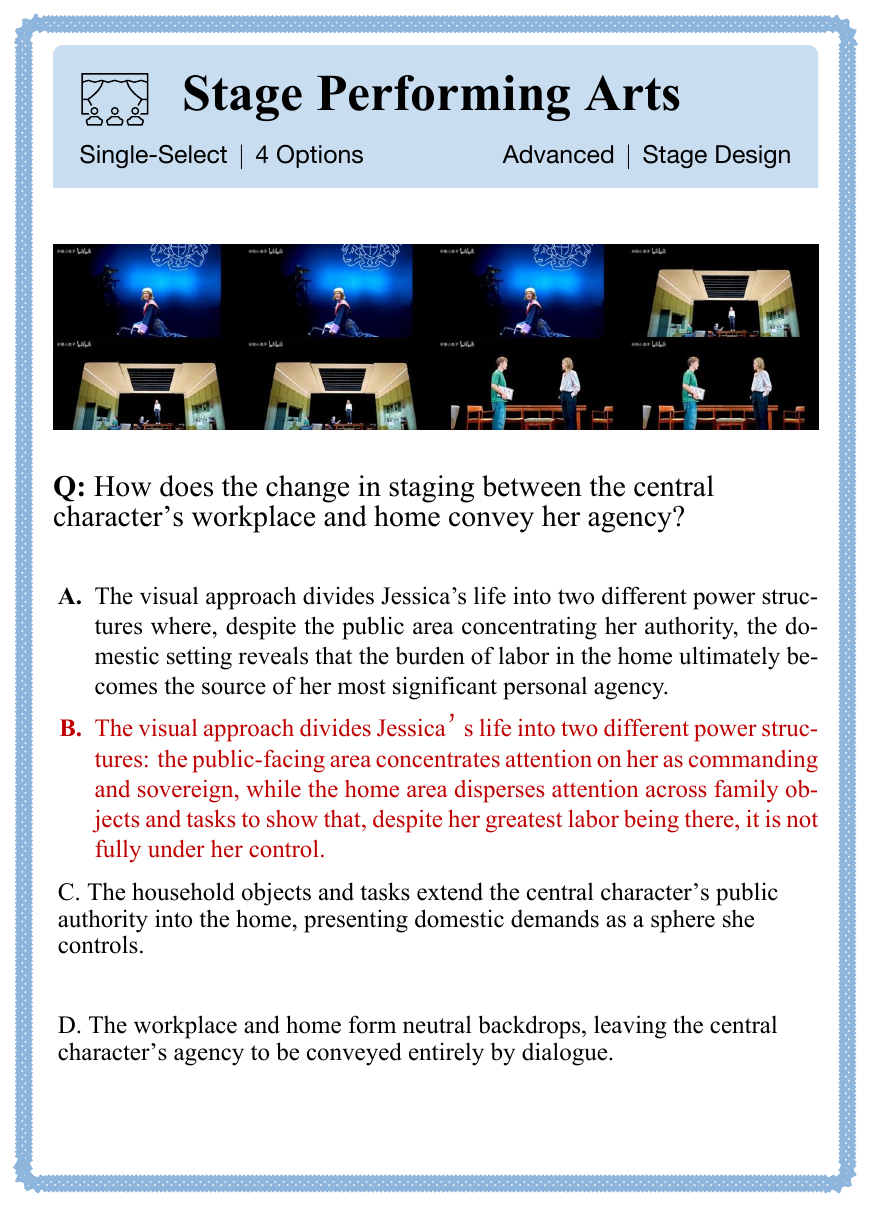}
\caption{\textit{Stage Performing Arts} sample 3.}
\label{fig:appendix_example_stg_03}
\end{subfigure}
\hfill
\begin{subfigure}[t]{0.48\textwidth}
\centering
\includegraphics[width=\linewidth]{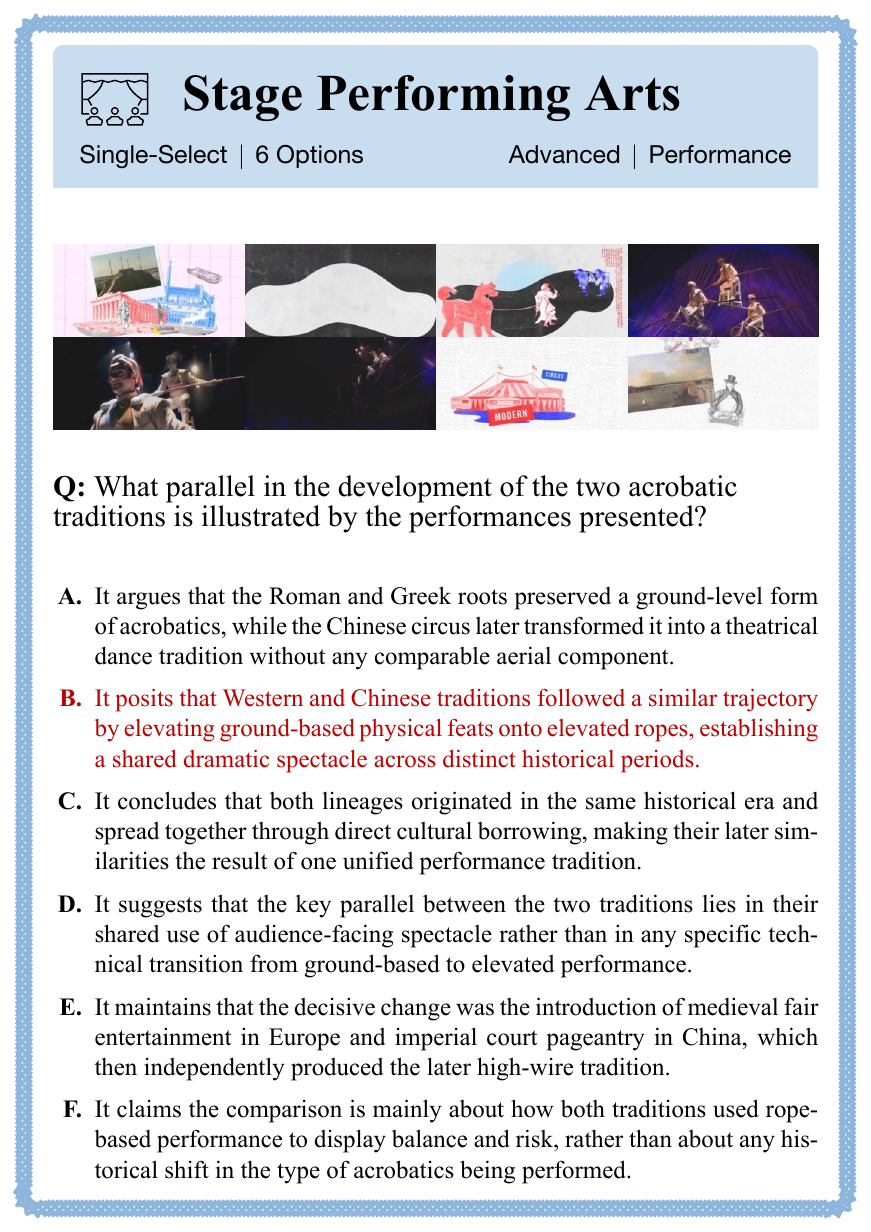}
\caption{\textit{Stage Performing Arts} sample 4.}
\label{fig:appendix_example_stg_04}
\end{subfigure}
\caption{Additional \textit{Stage Performing Arts} samples from \benchname.}
\label{fig:appendix_example_stg}
\end{figure}

\paragraph{Stage Performing Arts.}
\Cref{fig:appendix_example_stg} presents Stage Performing Arts cases that test understanding of staging, choreography, lighting design, and audio-visual coordination in live performance, where the answer hinges on how staged elements jointly construct meaning.

\begin{figure}[!htbp]
\centering
\begin{subfigure}[t]{0.48\textwidth}
\centering
\includegraphics[width=\linewidth]{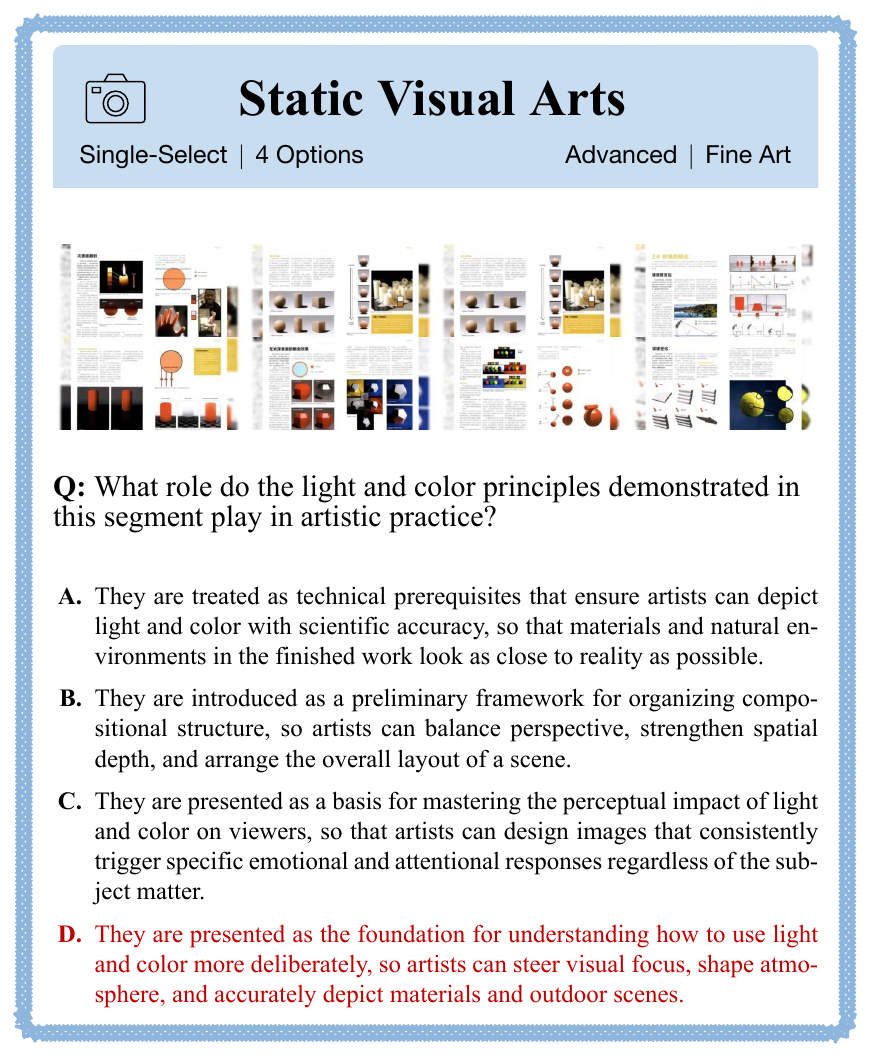}
\caption{\textit{Static Visual Arts} sample 1.}
\label{fig:appendix_example_vis_01}
\end{subfigure}
\hfill
\begin{subfigure}[t]{0.48\textwidth}
\centering
\includegraphics[width=\linewidth]{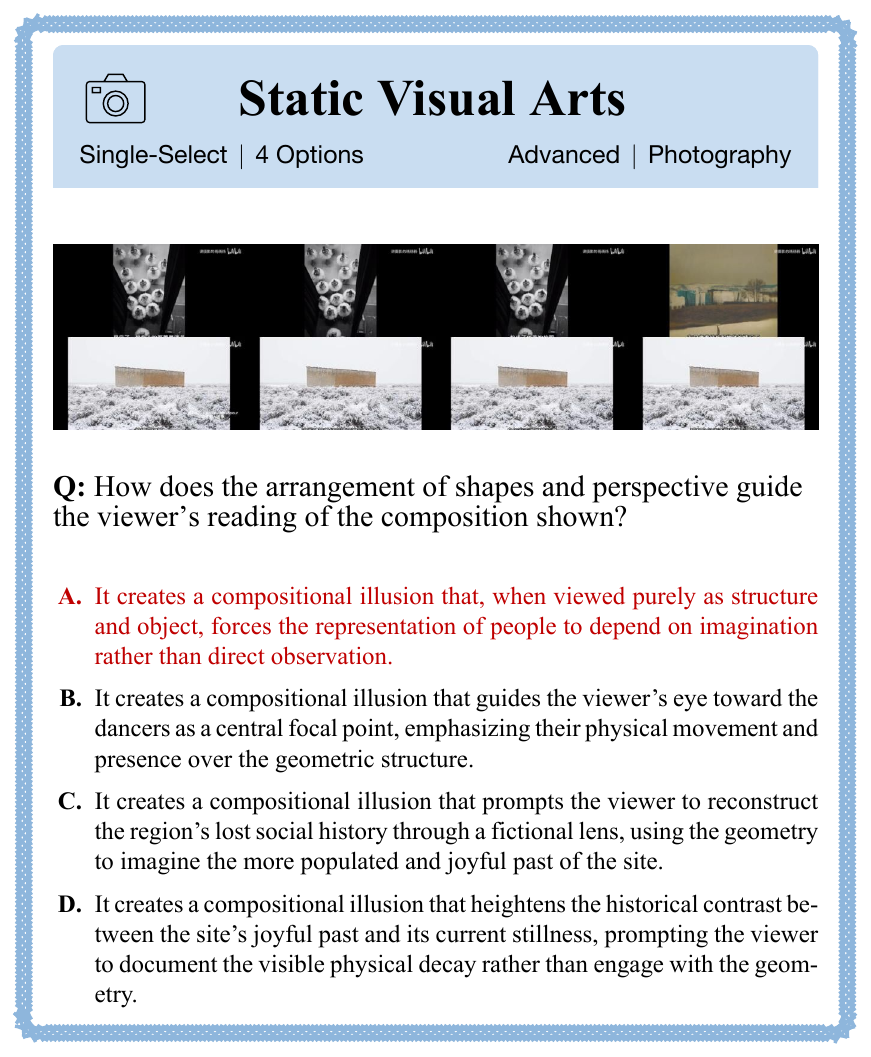}
\caption{\textit{Static Visual Arts} sample 2.}
\label{fig:appendix_example_vis_02}
\end{subfigure}

\vspace{0.6em}

\begin{subfigure}[t]{0.48\textwidth}
\centering
\includegraphics[width=\linewidth]{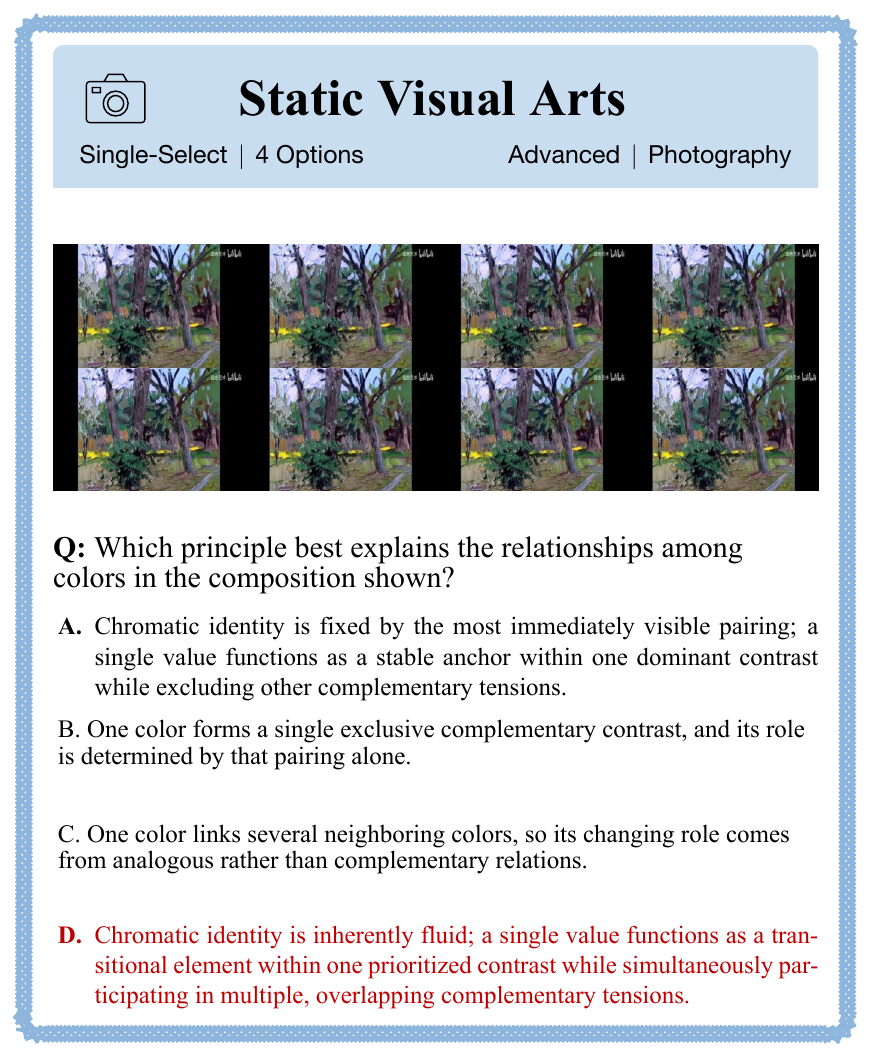}
\caption{\textit{Static Visual Arts} sample 3.}
\label{fig:appendix_example_vis_03}
\end{subfigure}
\hfill
\begin{subfigure}[t]{0.48\textwidth}
\centering
\includegraphics[width=\linewidth]{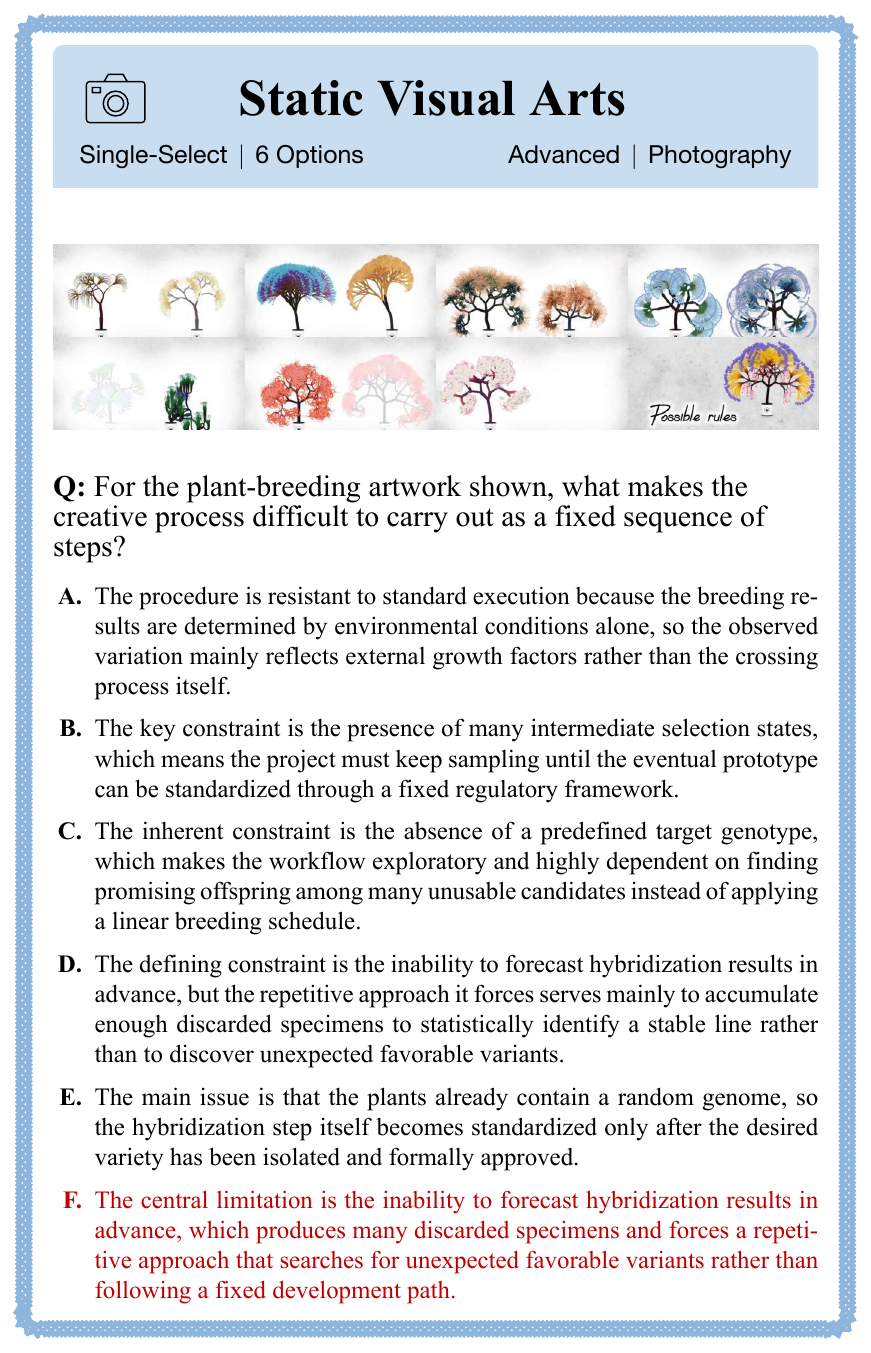}
\caption{\textit{Static Visual Arts} sample 4. Variation among the depicted plant forms connects exploratory material selection to the resulting artwork.}
\label{fig:appendix_example_vis_04}
\end{subfigure}
\caption{Additional \textit{Static Visual Arts} samples from \benchname.}
\label{fig:appendix_example_vis}
\end{figure}

\paragraph{Static Visual Arts.}
\Cref{fig:appendix_example_vis} features Static Visual Arts pairs covering composition, color theory, technique attribution, and aesthetic evaluation across painting, photography, and illustration, requiring fine-grained discrimination among visually similar artistic choices.

\clearpage

\clearpage

\clearpage

\clearpage

%% file: tables/comparison_table.tex
\begin{table*}[!htbp]
    \centering
    \small
    \setlength{\tabcolsep}{3pt}
    \resizebox{\textwidth}{!}{%
    \begin{tabular}{lrrrcrccccc}
        \toprule \textbf{Benchmarks} & \textbf{\#Clips} & \textbf{Avg Len.} & \textbf{\#QA} & \textbf{Anno.} & \textbf{QA Tok.} & \textbf{Sub.} & \textbf{Open} & \textbf{Aud.} & \textbf{Domain} & \textbf{Visual} \\
         & & \textbf{(s)} & \textbf{Pairs} & & & \textbf{Tok.} & \textbf{Domain} & & \textbf{Expert} & \textbf{Dep.} \\
        \midrule
        MSRVTT-QA~\citep{xu2017video} & 2,990 & 15.2 & 72,821 & A & 8.4 & \xmark & \cmark & \xmark & \xmark & \xmark \\
        MSVD-QA~\citep{chowdhury2018hierarchical} & 504 & 9.8 & 13,157 & A & 7.6 & \xmark & \cmark & \xmark & \xmark & \xmark \\
        TGIF-QA~\citep{jang2017tgif} & 9,575 & 3.0 & 8,506 & A\&M & 20.5 & \xmark & \cmark & \xmark & \xmark & \xmark \\
        ActivityNet-QA~\citep{yu2019activitynet} & 800 & 111.4 & 8,000 & M & 10.2 & \xmark & \xmark & \xmark & \xmark & \xmark \\
        STAR~\citep{wu2024star} & 7,098 & 11.9 & 7,098 & A & 19.5 & \xmark & \cmark & \xmark & \xmark & \xmark \\
        NExT-QA~\citep{xiao2021next} & 1,000 & 39.5 & 8,564 & A & 25.3 & \xmark & \cmark & \xmark & \xmark & \xmark \\
        MVBench~\citep{li2024mvbench} & 3,641 & 16.0 & 4,016 & A & 27.3 & \xmark & \cmark & \xmark & \xmark & \xmark \\
        Video-Bench~\citep{han2025video} & 5,917 & 56.0 & 17,036 & A\&M & 21.3 & \xmark & \cmark & \xmark & \xmark & \xmark \\
        EgoSchema~\citep{mangalam2023egoschema} & 5,063 & 180.0 & 5,063 & A\&M & 126.8 & \xmark & \xmark & \xmark & \xmark & \xmark \\
        AutoEval-Video~\citep{chen2024autoeval} & 327 & 14.6 & 327 & M & 11.9 & \xmark & \cmark & \xmark & \xmark & \xmark \\
        TempCompass~\citep{liu2024tempcompass} & 500 & 11.4 & 7,540 & A\&M & 49.2 & \xmark & \cmark & \xmark & \xmark & \xmark \\
        Video-MMMU~\citep{Hu2025VideoMMMUEK} & 900 & 588.4 & 900 & M & -- & \cmark & \xmark & \xmark & \cmark & \xmark \\
        \midrule
        \rowcolor{myblue} \benchname (Ours) & 4,016 & 50.26 & 4,016 & A & 168 & \cmark & \cmark & \cmark & \cmark & \cmark \\
        \bottomrule
    \end{tabular}%
    }

    \caption{\textbf{Comparison of \benchname with existing video understanding benchmarks.} ``Domain Expert'' indicates whether the benchmark requires domain-specific expertise. ``Visual Dep.'' indicates whether questions are explicitly designed to require visual evidence beyond text transcripts. \benchname is a comprehensive benchmark that simultaneously targets domain expertise in audiovisual arts and enforces visual dependency.}

    \label{tab:comparison}
\end{table*}

%% file: sections/revision_comparison.tex
\begingroup
\paragraph{Comparison with ShotBench.}
ShotBench~\citep{liu2025shotbench} evaluates expert cinematic language, while \benchname examines how creative choices produce expressive or functional effects across four artistic domains. Table~\ref{tab:shotbench_comparison} summarizes their complementary scope and design.
\begin{table}[htbp]
\centering
\caption{Comparison with ShotBench.}
\label{tab:shotbench_comparison}
\small
\begin{tabularx}{\linewidth}{@{}p{0.16\linewidth}XX@{}}
\toprule
Aspect & ShotBench & \benchname \\
\midrule
Coverage & Cinematography across eight dimensions & Four art categories and 11 sub-domains \\
Question focus & Cinematic attributes and visual grammar & Relations between creative choices, effects, and functions \\
Evidence & Film images and video clips & Narrator-removed clips selected from expert video essays \\
Answer format & Four-option single-choice; an option may contain multiple attributes & Four to eight options; single-select and multi-select \\
Construction & Trained annotation and professional cinematography resources with expert verification & Expert-commentary-informed generation, iterative review, and manual verification \\
Contribution & ShotQA training data and the ShotVL model & Cross-domain evaluation and set-valued interpretation \\
\bottomrule
\end{tabularx}
\end{table}

For example, recognizing a low-angle shot identifies a cinematic attribute, whereas explaining how that angle establishes a character's authority connects the technique to its function. Similarly, identifying a centrally placed performer describes staging, but explaining how a shift from a concentrated workplace composition to dispersed domestic objects alters perceived agency requires interpreting that staging (Figure~\ref{fig:appendix_example_stg_03}). These pairs distinguish attribute recognition from the technique-to-function reasoning emphasized by \benchname. Multi-select items further capture multiple valid dimensions within one answer space. This distinction impacts model ranking: InternVL3-78B exceeds InternVL3-8B in Overall ACC (37.80\% versus 33.07\%) but achieves lower Multi EM (13.53\% versus 20.30\%).
\endgroup

%% file: figures/source_keyword_prompt.tex
\begin{figure}[!htbp]
\centering
\begin{tcolorbox}[colback=blue!4!white, colframe=blue!50!black, title=\textbf{Keyword generation prompt (Cinematic Arts)}, fontupper=\scriptsize, boxrule=0.4pt, width=\linewidth]
\textbf{System.}\\
You are a video search expert. Given a user topic, generate high-quality search queries that target long-form video essays.\\
Requirements. (1) Focus on film breakdowns, film essays, audiovisual language, and director shot technique. (2) Use cinematic terminology such as montage, shot language, mise-en-sc\`ene, long take, color composition. (3) Match the way analytical viewers actually phrase searches. (4) Diversify across film genres and directorial styles. (5) Each query must be short, analytically loaded, and avoid pure title-only searches.\\
Output. A single JSON object \texttt{\{"keywords":[...]\}} and nothing else.
\medskip

\textbf{User.}\\
Topic: \texttt{\{topic\}}. Generate \texttt{\{num\_keywords\}} effective search queries that cover film breakdowns, film essays, and cinematic technique. Required focal points (combine freely):\\
1. Cinematography. Shot size (long shot, extreme close-up), camera angle (Dutch angle, bird's eye), movement (dolly zoom, tracking), lighting (high key, low key), composition (rule of thirds, leading lines).\\
2. Editing. Montage, long take, jump cut, parallel cut, flashback, transitions (match cut, smash cut), Kuleshov effect.\\
3. Mise-en-sc\`ene. Set and prop symbolism, blocking, costume.\\
4. Sound design. Sound-to-image relation, diegetic vs.\ non-diegetic, sound bridge, ambient sound, silence.\\
Examples may anchor on a director or auteur case.
\end{tcolorbox}
\caption{Keyword generation prompt for Cinematic Arts. The Static Visual, Stage Performing, and Game Arts variants share the same envelope, with the four numbered focal points replaced by the corresponding controlled vocabulary of the target category.}
\label{fig:source_keyword_prompt}
\end{figure}

%% file: figures/source_relevance_prompt.tex
\begin{figure}[!htbp]
\centering
\begin{tcolorbox}[colback=green!4!white, colframe=green!45!black, title=\textbf{Relevance judgment prompt (Stage Performing Arts, broad form)}, fontupper=\scriptsize, boxrule=0.4pt, width=\linewidth]
\textbf{System.}\\
You are a content analyst for stage and live performance video essays. Decide whether the given video belongs to analysis, commentary, or in-depth discussion of any stage or live performance form.\\
Scope (broad). Theater, opera, musical theater (stage design, lighting, directorial style); stand-up comedy specials; sketch comedy; long-form and short-form improv; ballet, modern dance, dance theater; cabaret and drag; circus and magic; mime, physical comedy, clowning; spoken word; talk-show performance; traditional opera forms such as Peking opera, kabuki, and noh.\\
Positive. Any of the above forms with explanation, analysis, ranking, or in-depth discussion. Comedy special reviews, comedian technique breakdowns, musical theater reviews.\\
Negative. Pure performance recording without analytical commentary, music videos, film analysis, gaming or anime content, news segments, or popular-science content.\\
Output. A single JSON object \texttt{\{"is\_relevant": bool, "reason": str, "confidence": float in [0,1]\}}.
\medskip

\textbf{User.}\\
Active keywords: \texttt{\{keywords\}}.\\
Candidate metadata.\\
Title: \texttt{\{title\}}\\
Description: \texttt{\{description\}}\\
Channel: \texttt{\{channel\}}\\
View count: \texttt{\{view\_count\}}\\
Decide whether this candidate focuses on analysis, commentary, or discussion of stage or live performance, and provide a short reason.
\end{tcolorbox}
\caption{Relevance judgment prompt for Stage Performing Arts. The Cinematic, Static Visual, and Game variants share the envelope and additionally hard-exclude DIY tutorials, software walkthroughs, and gameplay-only content respectively.}
\label{fig:source_relevance_prompt}
\end{figure}

%% file: figures/source_variant_prompt.tex
\begin{figure}[!htbp]
\centering
\begin{tcolorbox}[colback=orange!5!white, colframe=orange!70!black, title=\textbf{Variant expansion prompt (category-agnostic envelope)}, fontupper=\scriptsize, boxrule=0.4pt, width=\linewidth]
\textbf{System.}\\
You are an advanced search expert. Given a list of existing keywords, produce novel non-duplicate variants that cover additional relevant videos. Stay strictly inside the focal vocabulary supplied by the user. Do not return queries that are paraphrases of an existing one or that target a single named work.\\
Output. A single JSON object \texttt{\{"keywords":[...]\}} and nothing else.
\medskip

\textbf{User.}\\
Existing keywords (most recent ten):\\
\texttt{- \{kw\_1\}}\\
\texttt{- \{kw\_2\}}\\
\texttt{- ...}\\
\texttt{- \{kw\_10\}}\\
Generate \texttt{\{num\_variants\}} novel non-duplicate keywords that focus on \texttt{\{variant\_focus\}}.
\end{tcolorbox}
\caption{Variant expansion prompt. \texttt{variant\_focus} is the category-specific focal string; for example, \emph{cinematography, editing, mise-en-sc\`ene and sound design} for Cinematic Arts and \emph{stage performance, musical theater, stand-up comedy, dance and live performance art analysis} for Stage Performing Arts.}
\label{fig:source_variant_prompt}
\end{figure}

%% file: figures/source_human_vet_prompt.tex
\begin{figure}[!htbp]
\centering
\begin{tcolorbox}[colback=purple!4!white, colframe=purple!60!black, title=\textbf{Human-vetting prompt (final source-list cut)}, fontupper=\scriptsize, boxrule=0.4pt, width=\linewidth]
\textbf{System.}\\
You are a final-stage source curator. The candidate has already passed the automated relevance judgment. Your job is to reject the residual three failure modes that metadata filtering cannot detect. Return a single decision JSON.\\
Reject (i) \emph{narrator-only}. The on-screen visuals are a static talking head, generic stock footage, or a slide deck rather than the analyzed artifact itself.\\
Reject (ii) \emph{tutorial drift}. The video is structured as a how-to guide that walks through a workflow rather than as an analytical essay about an artifact, even when the keywords match.\\
Reject (iii) \emph{promotional cut-down}. The video is a trailer, showreel, advertisement, or fan compilation rather than commentary, even when long.\\
Output. A single JSON object\\
\texttt{\{"keep": bool, "reject\_reason": "none | narrator\_only | tutorial\_drift | promotional"\}}.
\medskip

\textbf{User.}\\
Title: \texttt{\{title\}}\\
Description: \texttt{\{description\}}\\
Visual sample (eight uniformly sampled frames): \texttt{\{frames\}}\\
Audio sample (first 30 seconds transcript): \texttt{\{transcript\}}\\
Decide whether the candidate should enter the construction pool.
\end{tcolorbox}
\caption{Human-vetting prompt applied as a final cut over admitted candidates.}
\label{fig:source_human_vet}
\end{figure}

%% file: tables/source_seeds_summary.tex
\begin{table}[!htbp]
\centering
\caption{Summary of the deterministic-fallback source-discovery keyword lists, with per-block totals and representative examples (3--5 per block, full list omitted).}
\label{tab:source_seeds_summary}
\small
\setlength{\tabcolsep}{3pt}
\begin{tabular}{@{}lrp{0.62\textwidth}@{}}
\toprule
\textbf{Block} & \textbf{\#KW} & \textbf{Representative keywords} \\
\midrule
\multicolumn{3}{@{}l}{\textbf{Cinematic Arts}} \\
\addlinespace[1pt]
Cinematography and lighting     & 4  & cinematography shot size breakdown; dutch angle vs bird's eye; dolly zoom vertigo effect; high/low key lighting study \\
Composition and mise-en-sc\`ene & 2  & rule of thirds leading lines composition; mise en scene blocking symbolism \\
Editing                         & 2  & montage cross-cutting film editing; Kuleshov effect explanation \\
Sound                           & 2  & diegetic vs non-diegetic sound design; sound bridge ambient noise film \\
\midrule
\multicolumn{3}{@{}l}{\textbf{Static Visual Arts}} \\
\addlinespace[1pt]
Fine art and art history        & 5  & painting composition analysis; color theory visual art breakdown; chiaroscuro oil painting; impressionism / post-impressionism; art history renaissance and baroque \\
Photography                     & 3  & photography visual language analysis; rule of thirds photography composition; fine art photography essay depth of field \\
Concept and digital art         & 2  & concept art design principles essay; digital art critique aesthetic analysis \\
\midrule
\multicolumn{3}{@{}l}{\textbf{Game Arts}} \\
\addlinespace[1pt]
Controlled-vocabulary core      & 10 & game visual storytelling environmental design; ray tracing real-time rendering aesthetic; cel shading hand-drawn game art; game CG cinematic animation; level design visual guidance \\
Case-anchored studies           & 30 & Elden Ring art direction dark fantasy; Hollow Knight hand-drawn art; Red Dead Redemption~2 lighting; Cuphead 1930s cartoon animation; Disco Elysium painting style \\
Studio / director vocabulary    & 11 & Naughty Dog cinematic game design; FromSoftware visual design philosophy; Nintendo art direction philosophy; Silent Hill visual symbolism \\
General game-art vocabulary     & 19 & game UI/UX design aesthetic; game environment art world building; Unreal Engine~5 Nanite showcase; game lighting mood atmosphere; horror game visual atmosphere \\
\midrule
\multicolumn{3}{@{}l}{\textbf{Stage Performing Arts}} \\
\addlinespace[1pt]
Stand-up comedy and specials    & 41 & George Carlin comedy genius; Bo Burnham Inside special; John Mulaney comedy style; stand-up structure and callback craft \\
Sketch comedy and improv        & 16 & SNL sketch comedy analysis; Key and Peele sketch genius; Monty Python comedy; UCB / Second City improv training \\
Musical theater                 & 41 & Sondheim musical genius; Hamilton cultural impact; Hadestown mythology; Lion King staging puppetry; Sweeney Todd analysis \\
Opera                           & 11 & Wagner Ring Cycle; Verdi dramatic analysis; Puccini La Boh\`eme; Carmen opera; opera vs musical theater \\
Drama and theater               & 12 & Shakespeare staging; Arthur Miller \emph{Death of a Salesman}; Tennessee Williams \emph{Streetcar}; immersive theater Sleep No More \\
Dance and ballet                & 8  & Swan Lake interpretation; Pina Bausch dance theater; contemporary dance analysis; Nutcracker staging \\
Cabaret, drag, variety          & 9  & RuPaul Drag Race performance; drag queen lip-sync; burlesque history; America's Got Talent best acts \\
Circus and magic                & 8  & Cirque du Soleil show analysis; Penn and Teller magic explained; Derren Brown mentalism; acrobatics performance art \\
Physical comedy and clown       & 7  & Charlie Chaplin comedy genius; Buster Keaton physical comedy; Marcel Marceau mime; Jacques Tati comedy style \\
Spoken word and poetry          & 5  & spoken word poetry performance; poetry slam competition; TED talk performance technique; oral storytelling tradition \\
Roast and panel comedy          & 9  & comedy roast best moments; late-night monologue analysis; Conan O'Brien comedy style; Graham Norton best moments \\
World theater traditions        & 10 & kabuki theater visual analysis; Beijing opera performance; Noh theater mask symbolism; commedia dell'arte; Bharatanatyam \\
General performance analysis    & 10 & stage presence technique; concert staging visual design; Super Bowl halftime analysis; Eurovision performance \\
\midrule
\textbf{Total}                  & \textbf{273} & \\
\bottomrule
\end{tabular}
\end{table}

%% file: figures/transcript_json_example.tex
\begin{figure}[!htbp]
\centering
\begin{tcolorbox}[colback=gray!4!white, colframe=gray!55!black, title=\textbf{Per-video transcription record (illustrative)}, fontupper=\scriptsize, boxrule=0.4pt, width=\linewidth]
\begin{verbatim}
{
  "video_id": "kuleshov_effect_explained_eisenstein_breakdown",
  "source": "<provider>",
  "category": "cinematic_arts",
  "video_path": "data/videos/cinematic_arts/<video_file>.mp4",
  "transcription": {
    "model": "Whisper-Large-v3",
    "language": "en",
    "text": "In this segment we examine how Eisenstein layers two...",
    "segments": [
      { "id": 0, "start":  0.00, "end":  6.84,
        "text": "In this segment we examine how Eisenstein layers two
                 unrelated shots so the cut itself produces meaning." },
      { "id": 1, "start":  6.84, "end": 13.21,
        "text": "Notice that the close-up of the face does not change,
                 yet the read of the emotion shifts with the cutaway." },
      { "id": 2, "start": 13.21, "end": 19.05,
        "text": "This is the Kuleshov effect, the foundation of analytic
                 montage we will trace through the rest of the essay." }
    ],
    "created_at": "2026-04-08T11:42:18+00:00"
  },
  "metadata": {
    "duration_seconds": 612.5,
    "frame_rate":       29.97,
    "resolution":       "1280x720",
    "audio_channels":   2,
    "audio_sample_rate": 48000
  }
}
\end{verbatim}
\end{tcolorbox}
\caption{Schema of the per-video transcription record produced by the preprocessing stage. Each video yields one such JSON file. The top-level fields anchor the record to a category and a source file; the \texttt{transcription} block contains the full free-text transcript together with timestamped sentence-level segments used downstream by the clip-level captioning and QA stages; \texttt{metadata} stores container properties used by the segmenter to bound clip duration and resolution.}
\label{fig:transcript_json_example}
\end{figure}

%% file: figures/clip_description_prompt.tex
\begin{figure}[!htbp]
\centering
\begin{tcolorbox}[colback=cyan!4!white, colframe=cyan!50!black, title=\textbf{Clip description prompt (Phase B)}, fontupper=\scriptsize, boxrule=0.4pt, width=\linewidth]
\textbf{System.}\\
You are an audiovisual arts analyst. Your task is to extract professional, fine-grained visual information from consecutive frames of a video essay clip.\\
General requirements.\\
1. Analyze the frame sequence chronologically and interpret it as one coherent visual segment.\\
2. Adjacent frames may look similar because they are continuous frames; do not misinterpret this as visual effects.\\
3. If on-screen text appears, quote the original text and provide an English translation when needed, then explain its contextual meaning.\\
4. Distinguish different people using concrete cues (clothing, position, posture, etc.).\\
5. Provide rich visual detail covering color, shape, texture, motion traits, and scene background.\\
Language. The \texttt{clip\_description} value must be written in natural English only.\\
Output. A single JSON object \texttt{\{"clip\_description": "The clip starts with..., develops through..., and ends with..."\}}.\\
At runtime the system prompt is concatenated with a category-specific guidance block (for Cinematic Arts: shot scale, camera angle, camera movement, light and color style, composition; the Static Visual, Stage Performing, and Game variants substitute their own controlled vocabulary).
\medskip

\textbf{User (first clip).}\\
Transcript segment for the current clip (narration or commentary): \texttt{\{transcript\_segment\}}\\
Please describe this first clip of the video.
\medskip

\textbf{User (subsequent clips, with context).}\\
Previous clip descriptions (chronological): \texttt{\{previous\_descriptions\}}\\
Transcript segment for the current clip (narration or commentary): \texttt{\{transcript\_segment\}}\\
Please describe the current clip, using previous clips as context when relevant.
\end{tcolorbox}
\caption{Clip description prompt used by Phase B of the construction pipeline (\Cref{sec:collection_annotation}). The first-clip user template omits the previous-clip context; subsequent clips receive the running chronological narrative so that descriptions remain locally coherent. The system prompt is concatenated at runtime with one of four category-specific guidance blocks.}
\label{fig:clip_description_prompt}
\end{figure}

%% file: figures/qa_generation_prompt.tex
\begin{figure}[!htbp]
\centering
\begin{tcolorbox}[colback=violet!4!white, colframe=violet!50!black, title=\textbf{QA generation prompt (Phase C)}, fontupper=\scriptsize, boxrule=0.4pt, width=\linewidth]
\textbf{System.}\\
You are an exam designer for an audiovisual arts benchmark. Based on clip-level visual descriptions and transcript text, generate high-quality QA items for the target art domain.\\
Core requirements.\\
1. Select evidence first. For each item, identify a single contiguous \texttt{relevant\_clips} range from the chronological clip descriptions and aligned transcript before writing the question. The item must be answerable from the narrator-removed evidence in that range.\\
2. Generate the question and correct answer from the selected evidence before generating distractors. Use a concrete visual or auditory anchor to locate the question without stating the relationship or effect that the solver must infer. Provide an accurate, professional, and well-reasoned correct answer.\\
3. Difficulty labels. Use \texttt{basic}, \texttt{intermediate}, or \texttt{advanced}.\\
4. Sub-domain labels. Assign a sub-domain from the category's controlled list.\\
5. Question type. Target a \texttt{multi\_select} share of 44.67\% across the generated corpus (2 to 4 independent correct answer points returned in \texttt{correct\_answers}); the rest are \texttt{single\_select} (one \texttt{correct\_answer}).\\
Quantity. Generate 3 to 5 questions per video.\\
Language. All textual fields must be in English only.\\
Output. A single JSON object \texttt{\{"questions":[\,...\,]\}} with each entry listing \texttt{relevant\_clips}, \texttt{question}, \texttt{question\_type}, \texttt{correct\_answer} or \texttt{correct\_answers}, \texttt{difficulty}, and \texttt{sub\_domain}.\\
At runtime the system prompt is concatenated with a category-specific question-design guidance block.
\medskip

\textbf{User.}\\
Video metadata. Category: \texttt{\{category\}}; Video ID: \texttt{\{video\_id\}}; Total clips: \texttt{\{num\_clips\}}.\\
Clip visual descriptions (chronological): \texttt{\{clip\_descriptions\}}.\\
Full transcript: \texttt{\{full\_transcript\}}.\\
First select a contiguous evidence range for each item, then generate 3 to 5 questions and their correct answers from the selected ranges, targeting a \texttt{multi\_select} share of 44.67\% across the generated corpus. Each question must test visual understanding and should not be answerable from transcript text alone.
\end{tcolorbox}
\caption{QA generation prompt used by Phase C. The full transcript and chronological clip descriptions support evidence selection first, followed by question and correct-answer generation from the selected range. Each item must be answerable from the narrator-removed evaluation clip. The system prompt is augmented with one of four category-specific question-design guidance blocks at runtime.}
\label{fig:qa_generation_prompt}
\end{figure}

%% file: figures/distractor_generation_prompt.tex
\begin{figure}[!htbp]
\centering
\begin{tcolorbox}[colback=red!4!white, colframe=red!50!black, title=\textbf{Distractor generation prompt (Phase D)}, fontupper=\scriptsize, boxrule=0.4pt, width=\linewidth]
\textbf{System.}\\
You are a benchmark item reviewer specializing in plausible distractors for audiovisual-arts multiple-choice questions.\\
Distractor strategies (use a different strategy for each distractor).\\
1. \emph{Technical misread}. Uses valid terminology but draws a wrong analysis.\\
2. \emph{Over-simplification}. Sounds partially correct but misses the core insight.\\
3. \emph{Concept swap}. Confuses closely related but distinct concepts.\\
4. \emph{Cause-effect inversion}. Reverses the causal relation between technique and effect.\\
5. \emph{Scope error}. Applies a correct technique analysis to the wrong scope, scene, or time range.\\
6. \emph{Temporal confusion}. Misattributes timing, sequence, or duration of techniques.\\
7. \emph{Partial truth}. Correct about one aspect but fundamentally wrong about the key insight.\\
Requirements. Each distractor uses a different strategy. Each distractor must look plausible on the surface and require domain knowledge to reject. Keep all distractors tightly grounded in domain-specific concepts and avoid semantic repetition. All distractor text must be in English.\\
Output. A single JSON object \texttt{\{"distractors":[\{"text":...,"misconception\_type":...\},\,...\,]\}} containing exactly \texttt{\{num\_distractors\}} entries.
\medskip

\textbf{User.}\\
Category: \texttt{\{category\}}; Sub-domain: \texttt{\{sub\_domain\}}; Question type: \texttt{\{question\_type\}}.\\
Question: \texttt{\{question\}}.\\
Correct answer(s): \texttt{\{correct\_answer\}}.\\
Scope for this category: \texttt{\{scope\_description\}}.\\
Generate exactly \texttt{\{num\_distractors\}} plausible distractors using a different strategy for each. \texttt{\{multi\_select\_guidance\}}.
\end{tcolorbox}
\caption{Distractor generation prompt used by Phase D. Each distractor uses a different strategy from the seven listed above. The category-specific \texttt{scope\_description} field uses the vocabulary of the source-curation prompts in \Cref{sec:appendix:source_curation}.}
\label{fig:distractor_generation_prompt}
\end{figure}

%% file: figures/quality_review_matrix.tex
\begin{figure}[!htbp]
\centering

\begin{tcolorbox}[colback=red!3!white, colframe=red!50!black, title=\textbf{Failure 1: narrator-dependent answerability}, fontupper=\scriptsize, boxrule=0.4pt, width=\linewidth]
\textbf{Observed bad case (stem).} ``Why does the narrator argue that the dolly zoom in this scene is more emotionally effective than a simple zoom?''\\
\textbf{Why it fails.} The answer paraphrases a verbal claim from the narrator transcript. Once the narrator track is removed for evaluation, the clip itself contains no spoken argument, so a model that has watched the visual evidence cannot recover the answer.\\
\textbf{Rule added (R3, \emph{not transcript-only}).} Every stem must be answerable from visual or audible evidence inside the narrator-removed clip. Questions that paraphrase narrator claims, summarize narrator opinions, or otherwise treat the transcript as ground truth are forbidden.\\
\textbf{Regenerated stem.} ``In the shot centered on the protagonist, how does the changing spatial relationship between the figure and the room shape the emotional effect?''
\end{tcolorbox}

\medskip

\begin{tcolorbox}[colback=orange!3!white, colframe=orange!70!black, title=\textbf{Failure 2: ambiguous stems}, fontupper=\scriptsize, boxrule=0.4pt, width=\linewidth]
\textbf{Observed bad case (stem).} ``What is interesting about the design of the still life shown?''\\
\textbf{Why it fails.} The stem names no observable element of the artifact; many mutually inconsistent answers are equally valid. Source clips whose narration only restates generic art theory frequently produce such stems.\\
\textbf{Rules added (R2, \emph{grounded in AV evidence}; F4, \emph{under-specified}).} R2 requires each stem to identify a concrete visible or audible element that locates the question, without supplying the relationship or effect that the solver must infer. F4 filters out source clips whose narration is too generic to anchor any concrete observable, so the question generator never receives them.\\
\textbf{Regenerated stem.} ``How does the placement of the porcelain bowl within the frame guide the viewer's attention in the still life shown?''
\end{tcolorbox}

\medskip

\begin{tcolorbox}[colback=yellow!6!white, colframe=olive!60!black, title=\textbf{Failure 3: weak or factually incorrect distractors}, fontupper=\scriptsize, boxrule=0.4pt, width=\linewidth]
\textbf{Observed bad case (options).} A. ``creates rhythm through editing tempo''; B. ``creates rhythm through editing pacing''; C. ``creates rhythm through editing speed''; D. ``creates rhythm through cut frequency''. Four lexical paraphrases of the same idea; a reader cannot use domain knowledge to choose between them.\\
\textbf{Rule added (Distractor prompt).} Each distractor must use a distinct strategy from the seven-strategy taxonomy (\emph{technical misread}, \emph{over-simplification}, \emph{concept swap}, \emph{cause-effect inversion}, \emph{scope error}, \emph{temporal confusion}, \emph{partial truth}); no two options may share a fifty-character prefix; option texts must be unique.\\
\textbf{Regenerated options.} A. (correct) ``locks the audience into the protagonist's heartbeat by aligning each edit point with a percussive accent''; B. (\emph{technical misread}) ``shortens the average shot length to compensate for an underlit interior''; C. (\emph{concept swap}) ``cross-cuts between two unrelated scenes to build narrative parallelism''; D. (\emph{cause-effect inversion}) ``slows the apparent action by inserting establishing shots after each cut''.
\end{tcolorbox}

\medskip

\begin{tcolorbox}[colback=blue!3!white, colframe=blue!55!black, title=\textbf{Failure 4: misaligned clip references}, fontupper=\scriptsize, boxrule=0.4pt, width=\linewidth]
\textbf{Observed bad case (\texttt{relevant\_clips}).} \texttt{[3, 8, 12]}. The matcher attached three disjoint clips even though the relevant evidence sits inside clip~8 alone; the evaluator who plays the concatenation of these three windows cannot reconstruct the original visual context.\\
\textbf{Rule added (Clip Match prompt).} The \texttt{relevant\_clips} field must form a single contiguous range. When the evidence spans disjoint regions of the video, the evidence-selection step first chooses the most informative contiguous range; the question generator then writes the stem and correct answer using only evidence within that range.\\
\textbf{Regenerated reference.} \texttt{relevant\_clips = [7, 8, 9]}, selected before generating a question about the visible setup that unfolds across those three consecutive ten-second windows.
\end{tcolorbox}

\caption{Four representative failure modes uncovered during the quality review loop (\Cref{sec:hitl_loop}), each shown with the bad case observed during pilot generation, the prompt-level rule added in response, and the regenerated form of the same item after the rule fired. The four rows correspond exactly to the four failure dimensions named in the main text. R/F-tagged rules are content-semantic red lines on the stem and options; the remaining two rules are schema-level constraints written into the Distractor and Clip Match prompts of the construction pipeline.}
\label{fig:quality_review_matrix}
\end{figure}

%% file: tables/failure_taxonomy.tex
\begin{table}[!htbp]
\centering
\caption{Construction-stage failure taxonomy from the early 5,000-item draft. Each row gives a review tag, severity, affected-item count, and mitigation. Counts describe the construction review; the final-set audit in Appendix~\ref{sec:appendix:option_audit} separately reports residual wording and equivalent-option issues.}
\label{tab:failure_taxonomy}
\small
\setlength{\tabcolsep}{3pt}
\renewcommand{\arraystretch}{1.15}
\begin{tabular}{@{}>{\raggedright\arraybackslash}p{0.11\textwidth}>{\raggedright\arraybackslash}p{0.20\textwidth}>{\centering\arraybackslash}p{0.07\textwidth}>{\raggedright\arraybackslash}p{0.27\textwidth}>{\raggedright\arraybackslash}p{0.27\textwidth}@{}}
\toprule
\textbf{Severity} & \textbf{review tag} & \textbf{Count} & \textbf{What the failure looks like} & \textbf{Mitigating rule added} \\
\midrule
CRITICAL & \texttt{INVALID\_\allowbreak{}LABEL}        & 175 & The \texttt{correct\_\allowbreak{}label} points to an option index that does not exist in the options array. & Distractor-prompt schema check rejects items whose label set does not match the option set; residual items replaced from the QA pool. \\
CRITICAL & \texttt{EMBEDDED\_\allowbreak{}MISMATCH}    &  60 & The stem contains an inline option list whose labels disagree with the canonical options array. & QA-Generation rule forbids inline option lists in stems. \\
CRITICAL & \texttt{MULTI\_\allowbreak{}0\_\allowbreak{}ANSWER}      &  37 & A multi-select item carries an empty \texttt{correct\_\allowbreak{}answers} list. & QA-Generation rule requires at least two correct answers for any \texttt{multi\_\allowbreak{}select} item. \\
\midrule
HIGH     & \texttt{ALL\_\allowbreak{}SAME\_\allowbreak{}PREFIX}     &  99 & All options share a fifty-character prefix and become indistinguishable on a first read. & Distractor-prompt rule: no two options may share a fifty-character prefix; strategy diversity enforced across options. \\
HIGH     & \texttt{DUPLICATE\_\allowbreak{}OPTS}       &  85 & Two or more options have identical text. & Distractor-prompt rule: option texts must be unique. \\
\midrule
MEDIUM   & \texttt{MULTI\_\allowbreak{}1\_\allowbreak{}ANSWER}      & 785 & A multi-select item carries only one correct answer and degenerates to single-select. & QA-Generation rule strengthened: \texttt{multi\_\allowbreak{}select} items must list 2 to 4 independent correct points. \\
MEDIUM   & \texttt{MANY\_\allowbreak{}OPTS\_\allowbreak{}SIMILAR}   &  83 & More than half of the option pairs share a fifty-character prefix. & Distractor-prompt rule: pairwise prefix-diversity threshold lowered to thirty percent of option pairs. \\
\midrule
LOW      & \texttt{ANSWER\_\allowbreak{}TEXT\_\allowbreak{}MISMATCH}& 639 & The \texttt{correct\_\allowbreak{}answer} field is a paraphrase rather than the exact option text. & Programmatic post-hoc alignment of \texttt{correct\_\allowbreak{}answer} to the exact option string. \\
\bottomrule
\end{tabular}
\end{table}

%% file: sections/revision_answer_agreement.tex
\begingroup
\subsubsection{Independent Answer Agreement and Adjudication}
\label{sec:appendix:answer_agreement}
We additionally evaluate 320 questions, with 40 single-select and 40 multi-select items from each of the four art categories. Two domain-matched experts independently answer every item using the evaluation clips and options, with the gold labels, source commentary, and other experts' answers hidden. Agreement is measured before adjudication. A third expert then reviews disagreements against the audiovisual evidence and records the rationale for the accepted answers, incorporating additional supported options where appropriate.

\begin{table}[htbp]
\centering
\caption{Independent expert answer agreement before adjudication. Each category contains 40 single-select and 40 multi-select questions, with two experts per item.}
\label{tab:answer_agreement}
\small
\begin{tabular}{@{}lrrrr@{}}
\toprule
Category & Single agreement (\%) & Cohen's $\kappa$ & Multi exact set (\%) & Jaccard \\
\midrule
Cinematic Arts & 92.50 & 0.89 & 75.00 & 0.86 \\
Static Visual Arts & 90.00 & 0.84 & 70.00 & 0.84 \\
Stage Performing Arts & 85.00 & 0.77 & 67.50 & 0.81 \\
Game Arts & 80.00 & 0.70 & 62.50 & 0.77 \\
\midrule
Overall & 86.88 & 0.79 & 68.75 & 0.82 \\
\bottomrule
\end{tabular}
\end{table}

The two independent experts achieve 86.6\% and 84.4\% ACC against the benchmark key on this subset. Adjudication confirms 303 of the 320 reference answer sets (94.69\%) and updates 17 (5.31\%); 13 questions or option sets receive wording revisions. Table~\ref{tab:answer_validation_scores} reports model scores on the same audited items. Label-only updates use stored predictions, while revised questions or options receive new evaluations. Claude-4.6-Opus gains 1.25 ACC points and Qwen-3.5-Plus gains 0.94, with their relative ordering preserved.

\begin{table}[htbp]
\centering
\caption{Score sensitivity to answer validation on 320 audited questions (160 single-select and 160 multi-select). Scores are percentages.}
\label{tab:answer_validation_scores}
\small
\begin{tabular}{@{}lrrrr@{}}
\toprule
& \multicolumn{2}{c}{Benchmark key} & \multicolumn{2}{c}{Adjudicated key} \\
Model & ACC & Multi EM & ACC & Multi EM \\
\midrule
Claude-4.6-Opus & 49.06 & 30.00 & 50.31 & 31.25 \\
Qwen-3.5-Plus & 47.19 & 23.75 & 48.13 & 24.38 \\
\bottomrule
\end{tabular}
\end{table}

This overlapping study measures answer-level agreement. The 90-item quality-rating exercise measures agreement on the four Likert dimensions, and the full-benchmark expert baseline combines one domain-matched answer per item. These three evaluations address item quality, consistency of interpretations, and expert performance, respectively. Reference answers represent source-supported interpretations within the provided answer space; alternative interpretations are assessed against the clip evidence during adjudication.
\endgroup

%% file: sections/revision_text_audit.tex
\begingroup
\subsection{On-Screen Text Review and Paired Controls}
\label{sec:appendix:text_audit}
All 4,016 clips selected for final evaluation were manually inspected, with OCR assistance, for embedded text. Hardcoded narration subtitles and explanatory or answer-revealing annotations were removed or masked, and the processed clips were manually checked. Dialogue subtitles, game interfaces, and text intrinsic to an artwork were retained as part of the work. Generated clip captions remain construction resources and are withheld from evaluated models.

\begin{table}[htbp]
\centering
\caption{Embedded text identified before processing during the full 4,016-clip review. Categories overlap; 1,449 items (36.08\%) contain at least one type.}
\label{tab:text_audit}
\small
\begin{tabularx}{\linewidth}{@{}Xrrp{0.29\linewidth}@{}}
\toprule
Text type & Items & \% & Final treatment \\
\midrule
Narration subtitles & 212 & 5.28 & Blurred \\
Explanatory or answer-revealing overlays & 187 & 4.66 & Blurred \\
Dialogue subtitles, game UI, artwork-intrinsic text & 1,146 & 28.54 & Retained \\
\bottomrule
\end{tabularx}
\end{table}

We compare explanatory-text masking on the same 187 affected questions (94 single-select and 93 multi-select), fixing the audio transcript, question, options, and evaluation settings. A text-only condition supplies the extracted text together with the question and options (Q+O). Table~\ref{tab:text_masking} reports the paired results. Masking decreases ACC by 2.67 points for Claude-4.6-Opus (95\% CI: $[-5.94,0.58]$) and 2.14 for Qwen-3.5-Plus ($[-5.31,1.03]$). Input with masked video and the audio transcript outperforms extracted text plus Q+O by 16.58 points for both models. A comparable-region blur control decreases ACC by 0.53 points for each model. These comparisons quantify the contribution of explanatory text and the value of the remaining evaluation inputs.

\begin{table}[htbp]
\centering
\caption{Paired text controls on the 187 affected questions (\%).}
\label{tab:text_masking}
\small
\begin{tabular}{@{}llrrr@{}}
\toprule
Model & Condition & ACC & Single CAA & Multi EM \\
\midrule
Claude-4.6-Opus & Before text masking & 52.41 & 68.44 & 30.11 \\
& Text-masked clips & 49.73 & 63.16 & 29.03 \\
& Extracted text + Q+O & 33.16 & 44.76 & 10.75 \\
\midrule
Qwen-3.5-Plus & Before text masking & 50.27 & 67.12 & 26.88 \\
& Text-masked clips & 48.13 & 63.16 & 25.81 \\
& Extracted text + Q+O & 31.55 & 42.12 & 9.68 \\
\bottomrule
\end{tabular}
\end{table}
\endgroup

%% file: sections/revision_option_audit.tex
\begingroup
\subsection{Final-Set Option Audit}
\label{sec:appendix:option_audit}
We distinguish shared wording from semantic equivalence. Shared prefixes can hold the subject and terminology constant while alternatives test different interpretations of the clip; equivalent options make the answer set ambiguous. We conduct a stratified random audit of 400 questions, with 200 single-select and 200 multi-select items. Of these, 386 (96.50\%) require no changes, 11 (2.75\%; Wilson 95\% CI: 1.54--4.86\%) require clearer wording, and three (0.75\%; 95\% CI: 0.26--2.18\%) contain equivalent options requiring answer-set correction. These 14 revisions are evaluated on the same audit sample in Table~\ref{tab:option_audit_scores}.

A separate full-benchmark screen identifies 617 questions with at least one option pair sharing a prefix of 50 or more characters. Expert review finds clearly distinct claims in 543 questions (88.0\%). The remaining 74 (12.0\%) show distinguishable interpretations with minor semantic overlap. This targeted screen identifies repeated phrasing; the separate 400-item random audit estimates the prevalence of wording and equivalence issues.

\begin{table}[htbp]
\centering
\caption{Option-audit score sensitivity on 400 questions (200 single-select and 200 multi-select). Scores are percentages.}
\label{tab:option_audit_scores}
\small
\begin{tabular}{@{}lrrrr@{}}
\toprule
& \multicolumn{2}{c}{Benchmark items} & \multicolumn{2}{c}{Audited items} \\
Model & ACC & Multi EM & ACC & Multi EM \\
\midrule
Claude-4.6-Opus & 48.75 & 29.50 & 50.25 & 30.50 \\
Qwen-3.5-Plus & 47.25 & 23.50 & 48.25 & 24.50 \\
\bottomrule
\end{tabular}
\end{table}

Claude's correct-answer count changes from 195 to 201 (+1.50 ACC points), and Qwen's from 189 to 193 (+1.00 point). Each model answers two additional multi-select questions exactly, increasing EM by 1.00 point. Their relative ordering is preserved on this audit sample. Label-only updates use stored predictions, while rewritten items receive new evaluations.

\paragraph{Option distinctions.}
In Figure~\ref{fig:appendix_example_vis_03}, D describes a color participating in several complementary relations. B states a single exclusive complementary contrast, and C specifies analogous rather than complementary relations. In Figure~\ref{fig:appendix_example_game_04}, A expresses the reference interpretation of uniform facades and stark lighting as artificial and oppressive; B, C, and F propose distinct sheltered, historical, and landmark-oriented readings. In Figure~\ref{fig:appendix_example_stg_03}, B expresses the contrast between public authority and domestic demands beyond the character's control; C expresses domestic authority, and D treats both settings as neutral backdrops. In Figure~\ref{fig:appendix_example_game_03}, H explains that visible extent helps players assess danger, while D asserts that size alone reveals exact damage at every distance. These alternatives test distinct interpretations of the visible evidence.
\endgroup

%% file: tables/app_results_details.tex
\begin{table*}[t]
\centering
\scriptsize
\setlength{\tabcolsep}{3.2pt}
\caption{Per-category accuracy and multi-select diagnostics (\%). Gemini-3.1-pro-preview, Qwen2.5-Omni-7B, MiniCPM-o-4.5, Gemma-4-E4B, and VideoLLaMA2.1-AV use V+A+T; other rows use V+T. Metrics use each evaluation's valid items within the corresponding category and question format. ACC measures fully correct answers, CAA adjusts single-select accuracy for chance, and P/R/F1 describe multi-select partial agreement.}
\label{tab:app_evaluation_results}
\resizebox{\textwidth}{!}{%
\begin{tabular}{@{}l ccccc ccccc ccccc ccccc c@{}}
\toprule
\multirow{3}{*}{\textbf{Model}}
& \multicolumn{5}{c}{\textbf{Cine.}}
& \multicolumn{5}{c}{\textbf{Static}}
& \multicolumn{5}{c}{\textbf{Stage}}
& \multicolumn{5}{c}{\textbf{Game}}
& \multirow{2}{*}{\textbf{Overall}} \\
\cmidrule(lr){2-6} \cmidrule(lr){7-11} \cmidrule(lr){12-16} \cmidrule(lr){17-21}
& \textbf{ACC} & \textbf{CAA} & \textbf{P} & \textbf{R} & \textbf{F1}
& \textbf{ACC} & \textbf{CAA} & \textbf{P} & \textbf{R} & \textbf{F1}
& \textbf{ACC} & \textbf{CAA} & \textbf{P} & \textbf{R} & \textbf{F1}
& \textbf{ACC} & \textbf{CAA} & \textbf{P} & \textbf{R} & \textbf{F1}
& \textbf{ACC} \\
\rowcolor{myblue} \multicolumn{22}{l}{\textit{Proprietary MLLMs}} \\
\midrule
GPT-5.4~\citep{singh2025openai}                  & 47.58 & 56.50 & 77.52 & 71.05 & 74.14 & 49.49 & 54.24 & 70.03 & 68.71 & 69.36 & 51.68 & 56.43 & 76.32 & 68.60 & 72.25 & 30.11 & 32.00 & 59.43 & 53.57 & 56.35 & 44.57 \\
Claude-4.6-Opus~\citep{anthropic2026opus46}              & 50.20 & 63.26 & 79.60 & 70.68 & 74.87 & 52.93 & 58.51 & 73.36 & 72.60 & 72.98 & 57.77 & 62.65 & 79.69 & 74.87 & 77.21 & 32.84 & 34.07 & 63.95 & 55.78 & 59.59 & 48.28 \\
Gemini-3.1-pro-preview~\citep{team2023gemini} & 34.34 & 43.16 & 41.48 & 43.29 & 42.37 & 39.70 & 42.70 & 42.63 & 39.07 & 40.78 & 43.86 & 49.50 & 47.51 & 46.01 & 46.75 & 29.91 & 38.72 & 44.05 & 45.43 & 44.73 & 36.90 \\
Grok-4.1~\citep{grok_4}               & 18.18 & 14.19 & 29.14 & 32.21 & 30.60 & 25.86 & 19.70 & 31.15 & 37.09 & 33.86 & 25.28 & 15.90 & 32.38 & 36.49 & 34.31 & 13.10 & 3.20 & 23.55 & 29.53 & 26.20 & 20.54 \\
Qwen-3.5-Plus~\citep{bai2025qwen3}                 & 51.11 & 68.88 & 65.89 & 58.32 & 61.87 & 51.62 & 64.36 & 49.28 & 48.06 & 48.67 & 55.23 & 60.69 & 57.43 & 55.48 & 56.44 & 31.67 & 38.80 & 44.53 & 45.06 & 44.79 & 47.26 \\
doubao-seed-1-8-251228~\citep{seed2025seed18}                          & 48.38 & 62.10 & 78.07 & 69.01 & 73.26 & 54.75 & 65.63 & 70.88 & 66.22 & 68.47 & 50.56 & 56.84 & 74.28 & 65.35 & 69.62 & 31.28 & 32.86 & 60.99 & 54.99 & 57.84 & 46.12 \\
GLM-4.5v~\citep{hong2025glm}                             & 23.03 & 16.17 & 33.52 & 35.69 & 34.57 & 13.94 & -2.97 & 29.70 & 35.74 & 32.44 & 22.64 & 13.60 & 33.04 & 34.17 & 33.60 & 9.19 & -4.34 & 21.26 & 27.90 & 24.13 & 17.13 \\
Kimi-K2.5~\citep{team2026kimi}                & 19.49 & 23.06 & 7.54 & 10.71 & 8.85 & 25.15 & 24.05 & 7.07 & 10.90 & 8.58 & 24.87 & 23.62 & 7.98 & 12.37 & 9.70 & 10.46 & 0.35 & 5.80 & 10.56 & 7.49 & 19.92 \\
\midrule
\rowcolor{myblue} \multicolumn{22}{l}{\textit{Open Source General Purpose MLLMs}} \\
\midrule
Qwen3.5-397B-A17B~\citep{bai2025qwen3}        & 50.91 & 62.65 & 56.88 & 55.30 & 56.08 & 49.60 & 57.45 & 47.30 & 50.83 & 49.00 & 49.44 & 56.60 & 42.30 & 45.13 & 43.67 & 29.62 & 35.79 & 30.33 & 35.95 & 32.90 & 44.77 \\
Qwen2.5-Omni-7B~\citep{xu2025qwen25omni}      & 34.55 & 36.47 & 65.45 & 63.97 & 64.70 & 39.19 & 40.76 & 58.47 & 62.30 & 60.33 & 38.38 & 34.43 & 62.81 & 60.44 & 61.60 & 19.16 & 8.31 & 51.79 & 50.92 & 51.36 & 32.69 \\
InternVL3-78B~\citep{chen2024expanding}       & 35.25 & 47.98 & 62.05 & 55.54 & 58.62 & 43.43 & 48.79 & 60.39 & 57.10 & 58.70 & 47.01 & 57.25 & 62.96 & 58.03 & 60.39 & 26.00 & 31.59 & 52.23 & 48.80 & 50.46 & 37.80 \\
InternVL3-8B~\citep{chen2024expanding}        & 40.71 & 43.41 & 67.64 & 72.12 & 69.81 & 40.71 & 36.23 & 64.54 & 70.64 & 67.45 & 33.60 & 30.48 & 59.08 & 66.23 & 62.45 & 17.79 & 6.66 & 59.40 & 45.06 & 51.25 & 33.07 \\
LLaVA-OneVision-7B~\citep{li2024llava}        & 17.68 & 22.01 & 40.32 & 29.79 & 34.27 & 25.66 & 25.77 & 40.99 & 32.98 & 36.55 & 24.26 & 25.09 & 41.89 & 31.62 & 36.04 & 14.27 & 10.24 & 44.20 & 32.33 & 37.34 & 20.42 \\
MiniCPM-o-4.5~\citep{yu2026minicpm}               & 35.15 & 35.72 & 58.04 & 56.24 & 57.12 & 35.96 & 32.92 & 52.02 & 51.47 & 51.74 & 36.04 & 31.49 & 53.72 & 53.11 & 53.41 & 18.67 & 6.14 & 42.12 & 43.34 & 42.72 & 31.35 \\
Gemma-4-E4B~\citep{team2024gemma}             & 30.71 & 39.40 & 60.79 & 47.25 & 53.17 & 33.33 & 32.55 & 56.03 & 47.69 & 51.52 & 30.76 & 31.44 & 60.13 & 47.02 & 52.77 & 16.03 & 10.30 & 45.12 & 39.85 & 42.32 & 27.61 \\
\midrule
\rowcolor{myblue} \multicolumn{22}{l}{\textit{Open Source Video-Specific MLLMs}} \\
\midrule
VideoLLaMA2.1-AV~\citep{cheng2024videollama}                          & 22.42 & 31.35 & 39.80 & 32.86 & 36.00 & 21.82 & 18.07 & 35.35 & 30.02 & 32.47 & 24.87 & 25.34 & 48.60 & 37.02 & 42.02 & 12.51 & 5.40 & 35.71 & 28.27 & 31.55 & 20.34 \\
VideoLLaMA3~\citep{zhang2025videollama}                          & 27.88 & 34.37 & 56.67 & 55.28 & 55.97 & 28.79 & 24.55 & 48.82 & 51.07 & 49.92 & 32.28 & 33.76 & 53.29 & 50.83 & 52.03 & 20.04 & 14.02 & 46.65 & 47.61 & 47.12 & 27.19 \\
Video-R1~\citep{feng2025video}                             & 23.74 & 30.50 & 55.80 & 61.21 & 58.38 & 30.10 & 28.70 & 53.53 & 65.63 & 58.97 & 34.92 & 38.49 & 56.04 & 62.37 & 59.03 & 18.48 & 13.87 & 47.81 & 51.12 & 49.41 & 26.72 \\
LongVU~\citep{shen2024longvu}                 & 14.85 & 14.40 & 54.52 & 87.69 & 67.24 & 15.66 & 6.50 & 52.39 & 86.35 & 65.21 & 14.82 & 5.46 & 53.03 & 89.52 & 66.60 & 14.17 & 7.75 & 53.60 & 91.65 & 67.64 & 14.87 \\
VideoRFT~\citep{wang2025videorft}             & 23.74 & 26.50 & 54.44 & 50.10 & 52.18 & 31.21 & 30.14 & 51.28 & 54.47 & 52.83 & 33.10 & 36.04 & 54.47 & 51.23 & 52.80 & 16.81 & 9.01 & 46.46 & 46.32 & 46.39 & 26.12 \\
VideoChat-R1~\citep{li2025videochat}                         & 25.86 & 35.87 & 53.18 & 47.66 & 50.27 & 31.11 & 29.05 & 55.69 & 52.83 & 54.22 & 30.56 & 33.04 & 53.46 & 42.02 & 47.06 & 17.11 & 6.46 & 44.80 & 42.78 & 43.77 & 26.07 \\
VideoChat2~\citep{li2024mvbench}                           & 15.96 & 17.20 & 44.59 & 31.77 & 37.11 & 19.29 & 14.35 & 34.36 & 32.44 & 33.38 & 21.42 & 18.67 & 41.18 & 32.98 & 36.63 & 14.57 & 10.45 & 38.05 & 28.76 & 32.76 & 17.78 \\
Video-XL-2~\citep{Shu2024VideoXLEV}                           & 20.30 & 28.63 & 42.59 & 32.13 & 36.63 & 27.07 & 29.29 & 42.47 & 35.12 & 38.45 & 31.88 & 40.42 & 43.73 & 33.33 & 37.83 & 17.69 & 19.17 & 41.75 & 31.42 & 35.86 & 24.18 \\
AKS~\citep{tang2025adaptive}                       & 16.06 & 17.61 & 34.57 & 24.95 & 28.98 & 22.83 & 21.46 & 33.56 & 30.01 & 31.68 & 25.79 & 28.01 & 45.22 & 32.85 & 38.06 & 12.81 & 6.34 & 34.54 & 28.74 & 31.38 & 19.30 \\
Q-Frame~\citep{zhang2025q}                    & 19.49 & 13.81 & 57.08 & 62.78 & 59.80 & 22.73 & 13.83 & 53.80 & 62.64 & 57.89 & 17.06 & 3.30 & 55.35 & 72.98 & 62.96 & 15.84 & 8.19 & 54.96 & 60.80 & 57.73 & 18.75 \\
LongVT~\citep{yang2025longvt}                                 & 17.58 & 17.99 & 32.34 & 30.70 & 31.50 & 26.26 & 21.61 & 46.06 & 41.90 & 43.88 & 25.18 & 21.92 & 49.74 & 40.09 & 44.39 & 13.29 & 5.02 & 25.93 & 25.30 & 25.61 & 20.52 \\
Video-CCAM~\citep{fei2024video}               & 18.18 & 23.38 & 38.76 & 28.78 & 33.04 & 21.21 & 18.31 & 37.63 & 32.25 & 34.73 & 20.30 & 16.92 & 37.76 & 29.30 & 33.00 & 10.65 & 1.05 & 38.78 & 29.94 & 33.79 & 17.53 \\
TimeChat~\citep{ren2024timechat}              & 14.14 & 12.27 & 49.04 & 63.50 & 55.34 & 16.87 & 9.66 & 42.52 & 57.33 & 48.83 & 14.31 & 4.61 & 43.50 & 52.76 & 47.69 & 12.41 & 5.04 & 45.60 & 49.31 & 47.38 & 14.42 \\
\bottomrule
\end{tabular}%
}
\end{table*}

%% file: sections/revision_language_controls.tex
\begingroup
\subsubsection{Strong-Model Question-and-Options Controls}
\label{sec:appendix:strong_shortcut_controls}
We evaluate options-only, question-plus-options (Q+O), and full-input conditions on all 4,016 questions, with the same items, option order, decoding settings, and scoring for each model. Q+O includes the task instruction, question, and options. Full input adds sampled video frames and a Whisper-Large-v3 transcript of the narrator-removed audio. Table~\ref{tab:strong_shortcut_controls} reports Claude-4.6-Opus and Qwen-3.5-Plus.

\begin{table}[htbp]
\centering
\caption{Strong-model language controls on all 4,016 questions (\%). Full input is video + audio transcript + question + options.}
\label{tab:strong_shortcut_controls}
\small
\begin{tabular}{@{}llrrr@{}}
\toprule
Model & Input & ACC & Single CAA & Multi EM \\
\midrule
Claude-4.6-Opus & Options only & 21.34 & 18.67 & 4.12 \\
& Q+O & 29.18 & 31.89 & 8.64 \\
& Full input & 48.28 & 55.13 & 28.91 \\
\midrule
Qwen-3.5-Plus & Options only & 22.41 & 20.36 & 4.85 \\
& Q+O & 30.35 & 33.93 & 9.25 \\
& Full input & 47.26 & 58.52 & 23.21 \\
\bottomrule
\end{tabular}
\end{table}

Adding video and the audio transcript to Q+O improves ACC by 19.10 for Claude-4.6-Opus, and 16.91 for Qwen-3.5-Plus. Single CAA and Multi EM improve across both systems, showing gains from the added evidence across both answer formats. Single CAA uses each question's option count to account for differences in guessing probability.
\endgroup

%% file: sections/revision_input_routes.tex
\begingroup
\subsection{Evaluation Input Routes}
\label{sec:appendix:input_routes}
All models receive the benchmark questions, options, and scoring instructions. Visual inputs are sampled at 1\,fps within each model's frame limit. V denotes video, A native audio, and T text. In V+T, T includes an audio transcript in addition to the instruction, question, and options. In V+A+T, native audio replaces the transcript. Expert narration and construction captions are withheld from the main evaluation. Table~\ref{tab:input_routes} summarizes model access and audio-derived inputs for the modality and language controls.

\begin{table}[htbp]
\centering
\caption{Input routes for the modality and language-control comparisons. All rows use 1\,fps visual sampling within the model's frame limit, plus the task instruction, question, and options.}
\label{tab:input_routes}
\small
\begin{tabularx}{\linewidth}{@{}lp{0.20\linewidth}X@{}}
\toprule
Model & Access & Audio-derived input \\
\midrule
GPT-5.4 & Official API & Whisper-Large-v3 transcript \\
Claude-4.6-Opus & Official API & Whisper-Large-v3 transcript \\
Qwen-3.5-Plus & Official API & Whisper-Large-v3 transcript \\
Grok-4.1 & API & Audio transcript \\
doubao-seed-1-8-251228 & API & Audio transcript \\
VideoLLaMA3 & Local weights & Audio transcript \\
Qwen2.5-Omni-7B & Local weights & Native audio waveform \\
\bottomrule
\end{tabularx}
\end{table}

The models in Table~\ref{tab:paired_audio_results} are accessed through OpenRouter and Volcengine Ark. V+T supplies video and audio transcripts. For V+A+T, Doubao, Gemini, MiMo, and Qwen receive MP4 files with audio tracks parsed by the server. Muse-Ember receives video and an MP3 audio file extracted with FFmpeg. The text input in V+A+T consists of the task instruction, question, and options.

\input{tables/paired_audio_results}

\begin{table}[!htbp]
\centering
\caption{Input-setting comparison on \benchname (\%): transcript-based V+T versus native-audio V+A+T.}
\label{tab:modality_ablation}
\footnotesize
\setlength{\tabcolsep}{3pt}
\begin{tabular}{@{}llrrr@{}}
\toprule
\textbf{Model} & \textbf{Input} & \textbf{ACC} & \textbf{CAA} & \textbf{EM} \\
\midrule
MiniCPM-o-4.5 & V+T & 23.83 & 19.28 & 11.72 \\
MiniCPM-o-4.5 & V+A+T & \textbf{31.35} & \textbf{27.05} & \textbf{18.90} \\
Gemma-4-E4B & V+T & 21.19 & 20.56 & 6.27 \\
Gemma-4-E4B & V+A+T & \textbf{27.61} & \textbf{28.67} & \textbf{9.06} \\
\bottomrule
    \end{tabular}
\end{table}

\begin{figure}[p]
\centering
\includegraphics[width=\textwidth,height=0.78\textheight,keepaspectratio]{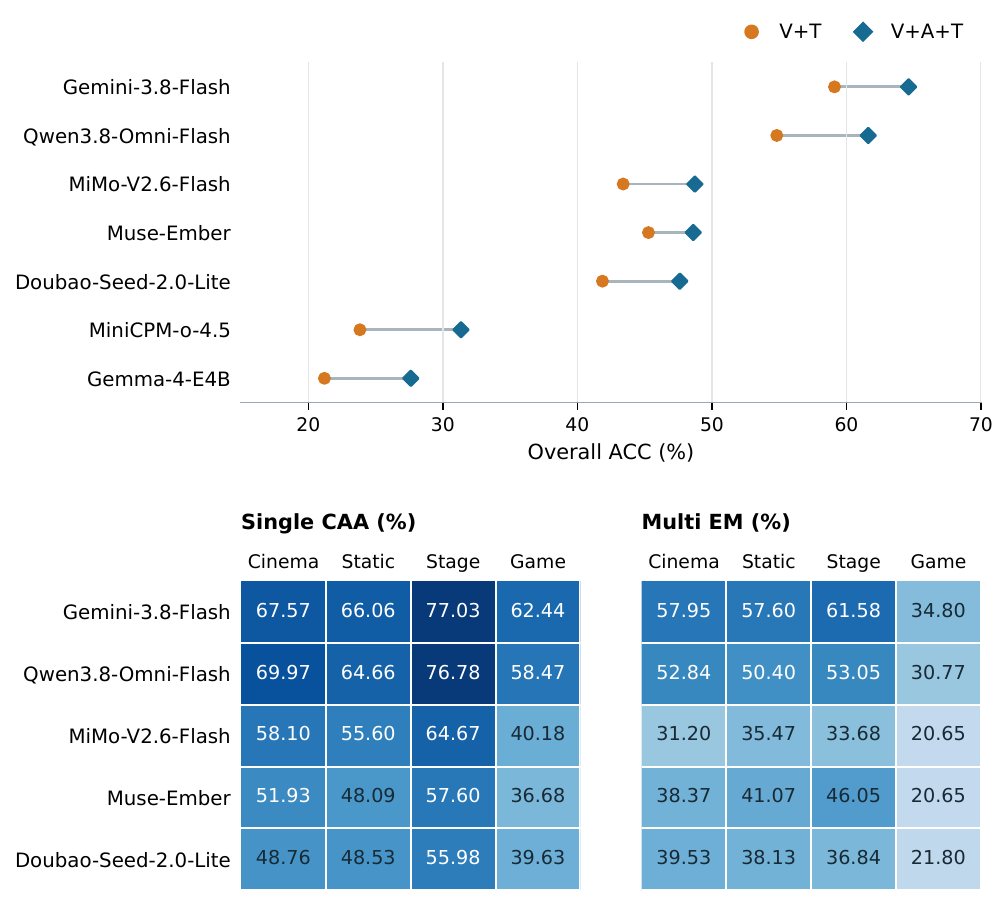}
\caption{Performance by input setting and art category. Top: Overall ACC for seven models evaluated with both V+T and V+A+T (Tables~\ref{tab:paired_audio_results} and~\ref{tab:modality_ablation}). Bottom: per-category Single CAA and Multi EM with native audio for the five models in Table~\ref{tab:paired_audio_results}. V+T includes audio transcripts; V+A+T replaces transcripts with native audio. Cinema, Static, Stage, and Game denote the four art categories.}
\label{fig:input_setting_comparison}
\end{figure}

Figure~\ref{fig:input_setting_comparison} summarizes the input-setting comparisons and native-audio category profiles. Gemini-3.8-Flash leads Multi EM in all four categories, while Qwen3.8-Omni-Flash leads Cinematic Single CAA. With transcript-based V+T, Gemini-3.8-Flash, MiMo-V2.6-Flash, and Muse-Ember achieve their highest category CAA on Game Arts (Table~\ref{tab:evaluation_results}); with native audio, Game Arts has their lowest CAA and EM.

Doubao and Gemini report dedicated audio-token counts; other providers report total input tokens.

Understanding the soundtrack contributes to artistic-intent understanding: rhythm, timbre, vocal delivery, and sound--image relationships can convey emotion, atmosphere, and narrative emphasis. Transcripts preserve spoken content but omit many acoustic cues. We compare transcript-based V+T with native-audio V+A+T. Overall ACC is higher with native audio by 5.75 points for Doubao, 5.51 for Gemini, 5.33 for MiMo, 3.33 for Muse-Ember, and 6.80 for Qwen. Single CAA increases by 2.28--8.78 points and Multi EM by 1.67--5.62 points. Muse-Ember has 4,016 versus 4,011 evaluated items, and Qwen has 3,994 versus 3,995, in V+A+T versus V+T; their differences are aggregate comparisons over the reported samples rather than matched-item estimates.

The benefit varies across artistic domains. All five models have higher CAA and EM in Cinematic Arts, Static Visual Arts, and Stage Performing Arts with native audio. In Game Arts, both metrics increase for Doubao but decrease for the other four models. These comparisons measure the performance difference when native audio replaces transcripts, with gains that depend on the model and domain.

The modality grouping describes the evidence available through each model's supported interface. The within-model comparisons in Table~\ref{tab:modality_ablation} compare the two input settings for each model, using valid scored items in each setting: V+A+T exceeds transcript-based V+T in ACC by 7.52 points for MiniCPM-o-4.5 and 6.42 for Gemma-4-E4B, with gains in both question formats. These results complement the domain-specific capability profiles with within-model input comparisons.
\endgroup

%% file: tables/paired_audio_results.tex
\begin{table*}[!htbp]
\centering

\caption{Input-setting comparisons on \benchname (\%). Single-select uses chance-adjusted accuracy (CAA); multi-select uses exact match (EM). Overall ACC is the percentage of questions answered exactly. Five models are evaluated with both V+T and V+A+T; Grok-4.3 uses V+T only. V+T includes audio transcripts without native audio; V+A+T includes native audio without transcripts. Each setting has 4,016 evaluated items except Muse-Ember V+T (4,011), Qwen V+T (3,995), and Qwen V+A+T (3,994). Comparisons for these two models use the reported samples rather than a matched-item subset.
}
\label{tab:paired_audio_results}
\vspace{-0.1in}
\resizebox{\textwidth}{!}{%
\begin{tabular}{@{}l|c|c|c>{\columncolor{mymultiem}}c|c>{\columncolor{mymultiem}}c|c>{\columncolor{mymultiem}}c|c>{\columncolor{mymultiem}}c|c>{\columncolor{mymultiem}[\tabcolsep][0pt]}c@{}}
\toprule
\multirow{3}{*}{\textbf{Model}}
  & \multirow{3}{*}{\textbf{Modality}}
  & \multirow{3}{*}{\textbf{Overall}}
  & \multicolumn{2}{c|}{\textbf{Overall}}
  & \multicolumn{2}{c|}{\textbf{Cinematic Arts}}
  & \multicolumn{2}{c|}{\textbf{Static Visual Arts}}
  & \multicolumn{2}{c|}{\textbf{Stage Perf. Arts}}
  & \multicolumn{2}{c}{\textbf{Game Arts}} \\
\cmidrule(lr){4-5} \cmidrule(lr){6-7} \cmidrule(lr){8-9} \cmidrule(lr){10-11} \cmidrule(lr){12-13}
  &
  &
  & \textbf{Single} & \textbf{Multi}
  & \textbf{Single} & \textbf{Multi}
  & \textbf{Single} & \textbf{Multi}
  & \textbf{Single} & \textbf{Multi}
  & \textbf{Single} & \textbf{Multi} \\
  &
  & \textbf{ACC}
  & \textbf{CAA} & \textbf{EM}
  & \textbf{CAA} & \textbf{EM}
  & \textbf{CAA} & \textbf{EM}
  & \textbf{CAA} & \textbf{EM}
  & \textbf{CAA} & \textbf{EM} \\
\midrule
doubao-seed-2-0-lite-260428~\citep{seed2026seed2} & V+A+T & 47.61 & 48.60 & 33.50 & 48.76 & 39.53 & 48.53 & 38.13 & 55.98 & 36.84 & 39.63 & 21.80 \\
doubao-seed-2-0-lite-260428~\citep{seed2026seed2} & V+T & 41.86 & 39.82 & 29.26 & 40.88 & 30.62 & 38.67 & 36.53 & 45.36 & 33.42 & 33.55 & 19.69 \\
gemini-3.8-flash~\citep{team2023gemini} & V+A+T & 64.62 & 68.57 & 51.90 & 67.57 & 57.95 & 66.06 & 57.60 & 77.03 & 61.58 & 62.44 & 34.80 \\
gemini-3.8-flash~\citep{team2023gemini} & V+T & 59.11 & 63.50 & 46.93 & 60.12 & 48.64 & 62.14 & 46.93 & 65.57 & 43.42 & 65.94 & 47.80 \\
mimo-v2.6-flash~\citep{qiao2026mimo} & V+A+T & 48.73 & 55.12 & 29.54 & 58.10 & 31.20 & 55.60 & 35.47 & 64.67 & 33.68 & 40.18 & 20.65 \\
mimo-v2.6-flash~\citep{qiao2026mimo} & V+T & 43.40 & 47.39 & 27.87 & 42.93 & 28.29 & 41.33 & 27.20 & 49.95 & 27.37 & 56.10 & 28.30 \\
muse-ember~\citep{meta2026museember} & V+A+T & 48.61 & 48.94 & 35.40 & 51.93 & 38.37 & 48.09 & 41.07 & 57.60 & 46.05 & 36.68 & 20.65 \\
muse-ember~\citep{meta2026museember} & V+T & 45.28 & 46.66 & 32.76 & 42.15 & 34.56 & 41.53 & 30.13 & 47.48 & 34.21 & 56.35 & 31.79 \\
qwen3.8-omni-flash~\citep{team2026qwen3} & V+A+T & 61.62 & 67.71 & 45.93 & 69.97 & 52.84 & 64.66 & 50.40 & 76.78 & 53.05 & 58.47 & 30.77 \\
qwen3.8-omni-flash~\citep{team2026qwen3} & V+T & 54.82 & 60.54 & 40.31 & 58.20 & 41.10 & 57.78 & 38.07 & 61.55 & 44.03 & 64.99 & 38.46 \\
grok-4.3~\citep{xai2026grok43} & V+T & 45.99 & 51.18 & 27.26 & 55.31 & 28.29 & 53.31 & 34.13 & 62.23 & 32.11 & 31.29 & 17.78 \\
\bottomrule
\end{tabular}%
}
\vspace{-0.1in}
\end{table*}

%% file: sections/revision_main_validation.tex
\begingroup
\makeatletter
\renewcommand\paragraph{\@startsection{paragraph}{4}{\z@}{0pt}{-1em}{\normalfont\normalsize\bfseries}}
\makeatother
\subsection{Evidence Controls and Answer Validation}
\label{sec:evidence_validation}

\input{tables/evidence_controls}
\paragraph{Language controls.}
Adding video and the Whisper-Large-v3 transcript to question-plus-options input improves ACC by 16.91--19.10 points and Multi EM by 13.96--20.27 points across two models (Table~\ref{tab:main_language_controls}). Appendix~\ref{sec:appendix:strong_shortcut_controls} details the matched protocol, options-only controls, and Single CAA.

\paragraph{On-screen text.}
The final clips have narration subtitles and answer-revealing overlays removed or masked. On 187 affected questions, masking explanatory text changes ACC by $-2.67$ points for Claude-4.6-Opus and $-2.14$ for Qwen-3.5-Plus. Input with masked video and the audio transcript outperforms extracted text plus Q+O by 16.58 points for both models. Appendix~\ref{sec:appendix:text_audit} details the matched inputs, paired intervals, and comparable-region blur control.

\paragraph{Answer validation.}
On 320 questions independently answered by two blinded domain experts, single-answer agreement is 86.88\% ($\kappa=0.79$) and multi-answer exact-set agreement is 68.75\% (Jaccard $=0.82$). Adjudication confirms 303 answer sets (94.69\%) and updates 17. Across this study and the separate 400-question option audit, ACC changes by 0.94--1.50 points for Claude-4.6-Opus and Qwen-3.5-Plus, preserving their ordering on each sample (Appendices~\ref{sec:appendix:answer_agreement} and~\ref{sec:appendix:option_audit}). The sensitivity scores are computed on the audited subsets; the leaderboards report scores against the benchmark reference answers.
\par
\endgroup

%% file: tables/evidence_controls.tex
\begin{wraptable}{r}{0.53\textwidth}
\centering
\vspace{-0.8em}

\caption{Language controls on all 4,016 questions (\%). Q+O is the task instruction, question, and options. Full input adds video and the audio transcript.}
\label{tab:main_language_controls}
\footnotesize
\setlength{\tabcolsep}{2pt}
\begin{tabular}{@{}lrrrrr@{}}
\toprule
& \multicolumn{3}{c}{Overall ACC} & \multicolumn{2}{c}{Multi EM} \\
Model & Q+O & Full & Gain (pp) & Q+O & Full \\
\midrule
Claude-4.6-Opus & 29.18 & 48.28 & 19.10 & 8.64 & 28.91 \\
Qwen-3.5-Plus & 30.35 & 47.26 & 16.91 & 9.25 & 23.21 \\
\bottomrule
\end{tabular}
\end{wraptable}